\documentclass[11pt]{article}

\usepackage[preprint]{acl}

\usepackage{times}
\usepackage{latexsym}
\usepackage[T1]{fontenc}
\usepackage[utf8]{inputenc}
\usepackage{microtype}
\usepackage{inconsolata}
\usepackage{graphicx}
\usepackage{amsmath,amssymb}
\usepackage{booktabs}
\usepackage[table]{xcolor}
\usepackage{caption}
\usepackage{subcaption}
\usepackage{multirow}
\usepackage{makecell}
\usepackage{pifont}
\usepackage{enumitem}
\usepackage{tikz}
\usetikzlibrary{arrows.meta,positioning,calc,shapes.geometric}
\usepackage{pgfplots}
\usepackage{placeins}
\pgfplotsset{compat=1.18}
\usepgfplotslibrary{groupplots}

\definecolor{cJev}{HTML}{2F5DA8}   %
\definecolor{cDLKT}{HTML}{D9762B}  %
\definecolor{cKNN}{HTML}{8C8C8C}   %
\definecolor{cAnch}{HTML}{3A9A7A}  %
\definecolor{focusbg}{RGB}{235,245,255} %
\pgfplotscreateplotcyclelist{dsets}{%
  {color=cJev!55, mark=o, mark size=1.2pt, line width=0.6pt},
  {color=cDLKT!70, mark=square, mark size=1.1pt, line width=0.6pt},
  {color=cAnch!70, mark=triangle, mark size=1.3pt, line width=0.6pt},
  {color=violet!60, mark=diamond, mark size=1.3pt, line width=0.6pt},
  {color=red!60, mark=pentagon, mark size=1.2pt, line width=0.6pt},
  {color=teal!60, mark=x, mark size=1.4pt, line width=0.6pt},
  {color=brown!70, mark=star, mark size=1.4pt, line width=0.6pt}}

\tikzset{mock/.style={font=\Huge\bfseries, text=red!25, rotate=20, opacity=0.8}} %
\newcommand{\JevZ}{Jev\textsubscript{0-shot}}

\newcommand{\cmark}{\ding{51}}
\newcommand{\xmark}{\ding{55}}

\newcommand{\csDeltaEight}{+.033}
\newcommand{\csWinsEight}{7/7}
\newcommand{\csDeltaSixteen}{+.023}
\newcommand{\csWinsSixteen}{7/7}
\newcommand{\csDeltaThirtyTwo}{+.016}
\newcommand{\csWinsThirtyTwo}{7/7}
\newcommand{\csDeltaSixtyFour}{+.004}
\newcommand{\csWinsSixtyFour}{4/7}
\newcommand{\csDeltaOneTwentyEight}{$-$.002}

\newcommand{\csDeltaTwoFiftySix}{$-$.008}

\newcommand{\csDeltaFull}{$-$.028}

\newcommand{\nLead}{64}
\newcommand{\nCross}{128}
\newcommand{\avgPriorZero}{.706}
\newcommand{\avgEvidenceEight}{.595}
\newcommand{\avgBothEight}{.722}
\newcommand{\avgDlktEight}{.689}
\newcommand{\loktEight}{.672}
\newcommand{\loktFull}{.672}
\newcommand{\zeroWinsEight}{5/7}
\newcommand{\sigTestEight}{7/7}

\newcommand{\sigTestSixteen}{6/7}

\newcommand{\sigTestThirtyTwo}{3/7}

\newcommand{\sigTestSixtyFour}{2/7}

\newcommand{\sigTestTwoFiftySix}{1/7}

\newcommand{\tunedDeltaEight}{+.070}
\newcommand{\tunedSigEight}{7/7}
\newcommand{\tunedDeltaSixteen}{+.042}
\newcommand{\tunedSigSixteen}{7/7}
\newcommand{\tunedDeltaThirtyTwo}{+.030}
\newcommand{\tunedSigThirtyTwo}{5/7}
\newcommand{\tunedDeltaSixtyFour}{+.010}
\newcommand{\tunedSigSixtyFour}{4/7}
\newcommand{\nSig}{16}
\newcommand{\genGptZero}{.680}
\newcommand{\genGemZero}{.670}
\newcommand{\genDsZero}{.626}

\newcommand{\genNineBZero}{.687}
\newcommand{\genNineBEight}{.699}
\newcommand{\genLayaZero}{.536}

\newcommand{\gainJev}{+.016}
\newcommand{\gainDs}{+.010}
\newcommand{\gainJevSig}{2/7}
\newcommand{\gainDsSig}{1/7}
\newcommand{\gainDiffSig}{0/7}
\newcommand{\gapZeroMatched}{.077}
\newcommand{\ptwoWins}{33/35}
\newcommand{\ptwoZeroWins}{25/35}
\newcommand{\ptwoJevEarly}{.714}
\newcommand{\ptwoDlktEarly}{.691}
\newcommand{\pthreeJev}{.713}
\newcommand{\pthreeZero}{.704}
\newcommand{\pthreeLokt}{.664}
\newcommand{\pthreeDlkt}{.729}
\newcommand{\pthreeWins}{1/7}
\newcommand{\abZero}{.706}

\newcommand{\abKnn}{.729}
\newcommand{\abKCsixteen}{.720}
\newcommand{\abKCsixtyfour}{.724}
\newcommand{\abFinal}{.730}
\newcommand{\clstWins}{12/12}
\newcommand{\clstBelowDlkt}{11/12}
\newcommand{\eceTkt}{.253}

\newcommand{\eceLokt}{.209}

\newcommand{\eceZero}{.084}

\newcommand{\eceDlkt}{.076}
\newcommand{\brierDlkt}{.192}
\newcommand{\eceJev}{.070}
\newcommand{\brierJev}{.182}
\newcommand{\errItemSeen}{+.036}
\newcommand{\errItemUnseen}{+.028}
\newcommand{\errAccLow}{+.050}

\newcommand{\errAccHigh}{+.063}
\newcommand{\nDLKT}{28}
\newcommand{\nDLKTadded}{17}
\newcommand{\DLKTaddedList}{AT-DKT~\citep{liu_2023_atdkt}, ATKT-fix~\citep{guo_2021_atkt}, Deep-IRT~\citep{yeung_2019_deepirt}, DIMKT~\citep{shen_2022_dimkt}, DKVMN~\citep{zhang_2017_dkvmn}, DTransformer~\citep{yin_2023_dtransformer}, IEKT~\citep{long_2021_iekt}, KQN~\citep{lee_2019_kqn}, lefoKT-AKT~\citep{bai_2025_lefokt}, MoC-KT~\citep{hou_2026_mockt}, qDKT~\citep{sonkar_2020_qdkt}, QIKT~\citep{chen_2023_qikt}, reKT~\citep{shen_2024_rekt}, SAINT~\citep{choi_2020_saint}, SAINT+~\citep{shin_2021_saintplus}, SAKT~\citep{pandey_2019_sakt}, UKT~\citep{cheng_2025_ukt}}
\newcommand{\ps}[1]{\textsuperscript{\tiny(#1)}} %

\title{Can a System-One LLM Perform Knowledge Tracing\\
When Few or No Learners Are Logged?}

\author{
\textbf{Unggi Lee}$^{1,\dagger}$ \quad \textbf{Haeun Park}$^{2}$ \\
$^{1}$Korea University Sejong Campus \quad $^{2}$Korea University \\
\texttt{codingchild@korea.ac.kr} \quad \texttt{hxxnpark@gmail.com} \\
$^{\dagger}$Corresponding author
}

\begin{document}
\maketitle

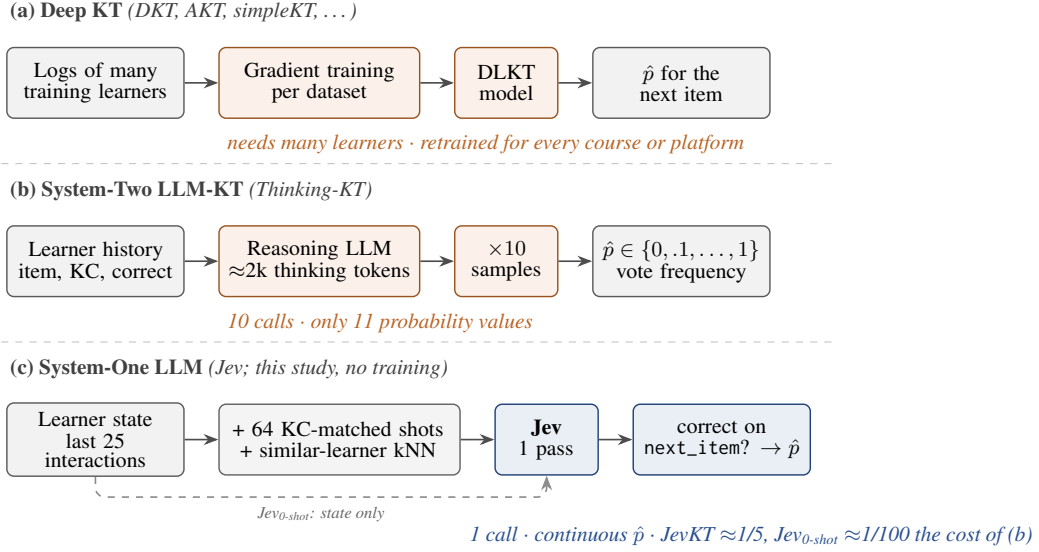
\begin{figure*}[t]
\centering
\resizebox{0.86\textwidth}{!}{%
\begin{tikzpicture}[
  font=\small,
  box/.style={draw=#1!70!black, fill=#1!10, rounded corners=3pt, align=center,
              minimum height=1.05cm, inner sep=4pt, line width=0.6pt},
  arr/.style={-{Stealth[length=2.2mm]}, line width=0.7pt, draw=black!70},
  lab/.style={font=\small\bfseries, text=black!75},
  note/.style={font=\footnotesize\itshape, align=left},
  node distance=0.5cm and 0.5cm]

\node[lab, anchor=west] at (-0.2,1.6) {(a) Deep KT \textmd{\textit{(DKT, AKT, simpleKT, \dots)}}};
\node[box=gray, text width=2.35cm] (h3) at (1.2,0.55) {Logs of many\\[-1pt]\footnotesize training learners};
\node[box=cDLKT, text width=2.7cm, right=of h3] (r3) {Gradient training\\[-1pt]\footnotesize per dataset};
\node[box=cDLKT, text width=1.25cm, right=of r3] (v3) {DLKT\\[-1pt]\footnotesize model};
\node[box=gray, text width=2.35cm, right=of v3] (p3) {$\hat p$ for the\\[-1pt]\footnotesize next item};
\draw[arr] (h3) -- (r3); \draw[arr] (r3) -- (v3); \draw[arr] (v3) -- (p3);
\node[note, text=cDLKT!85!black, below=0.1cm of r3.south west, anchor=north west]
  {needs many learners $\cdot$ retrained for every course or platform};

\draw[black!25, dashed] (-0.2,-0.65) -- (12.2,-0.65);

\node[lab, anchor=west] at (-0.2,-1.05) {(b) System-Two LLM-KT \textmd{\textit{(Thinking-KT)}}};
\node[box=gray, text width=2.35cm] (h2) at (1.2,-2.1) {Learner history\\[-1pt]\footnotesize item, KC, correct};
\node[box=cDLKT, text width=2.7cm, right=of h2] (r2) {Reasoning LLM\\[-1pt]\footnotesize $\approx$2k thinking tokens};
\node[box=cDLKT, text width=1.25cm, right=of r2] (v2) {$\times 10$\\[-1pt]\footnotesize samples};
\node[box=gray, text width=2.35cm, right=of v2] (p2) {$\hat p \in \{0, .1, \dots, 1\}$\\[-1pt]\footnotesize vote frequency};
\draw[arr] (h2) -- (r2); \draw[arr] (r2) -- (v2); \draw[arr] (v2) -- (p2);
\node[note, text=cDLKT!85!black, below=0.1cm of r2.south west, anchor=north west]
  {10 calls $\cdot$ only 11 probability values};

\draw[black!25, dashed] (-0.2,-3.3) -- (12.2,-3.3);

\node[lab, anchor=west] at (-0.2,-3.7) {(c) System-One LLM \textmd{\textit{(Jev; this study, no training)}}};
\node[box=gray, text width=2.35cm] (h1) at (1.2,-4.75) {Learner state\\[-1pt]\footnotesize last 25\\[-2pt]\footnotesize interactions};
\node[box=cKNN, text width=3.3cm, right=of h1] (e1) {\footnotesize + 64 KC-matched shots\\[-1pt]\footnotesize + similar-learner kNN};
\node[box=cJev, text width=1.25cm, right=of e1] (j1) {\textbf{Jev}\\[-1pt]\footnotesize 1 pass};
\node[box=cJev, text width=2.35cm, right=of j1] (p1) {correct on\\[-1pt]\texttt{next\_item}? $\to \hat p$};
\draw[arr] (h1) -- (e1); \draw[arr] (e1) -- (j1); \draw[arr] (j1) -- (p1);
\draw[arr, black!45, dashed, rounded corners=4pt] (h1.south) -- ++(0,-0.3) -| (j1.south)
  node[pos=0.25, below, font=\scriptsize\itshape, text=black!60] {\JevZ: state only};
\node[note, text=cJev!85!black, below=0.62cm of e1.south east, anchor=north west]
  {1 call $\cdot$ continuous $\hat p$ $\cdot$ JevKT $\approx$1/5, \JevZ{} $\approx$1/100 the cost of (b)};
\end{tikzpicture}}
\caption{Three ways to predict whether a learner answers the next item correctly. (a) Deep KT
is trained on the logs of many learners. (b) System-Two LLM-KT samples ten reasoning traces
and reads a probability from vote frequencies. (c) This study poses the prediction to an
off-the-shelf System-One model (Jev) as one typed question about a serialised learner state,
either alone (\JevZ) or with 64 KC-matched examples and a similar-learner statistic (JevKT = Jev + 64-shot + kNN);
the probability is read from a single pass, without gradient training on the target
dataset.}
\label{fig:pipeline}
\end{figure*}

\begin{abstract}
Knowledge tracing (KT) models need many logged learners, so a new course or platform starts
without a usable model. In LLM-based KT the LLM generates the answer, which we call System-Two;
it is either fine-tuned on the target data or reasons and votes over ten samples, which is slow and
gives coarse probabilities. We ask whether an off-the-shelf System-One LLM, which returns a
probability for a typed question directly in a single pass, can perform KT when few or no
learners are logged. On seven datasets, Jev without any data from the target platform reaches
a mean AUC of \avgPriorZero, above the best of \nDLKT{} deep KT models trained on 8 learners
(\avgDlktEight) and above System-Two Thinking-KT on all seven datasets (.650) at about
1/100 of its API cost. Adding examples and a similar-learner statistic from the logged
learners (JevKT) raises this to \avgBothEight{}; JevKT stays significantly ahead of deep KT up to
\nSig{} learners and ahead on average up to \nLead, and supervised KT catches up between
\nLead{} and \nCross{} learners. Among the readers we tested, the gain is specific to Jev, since
three other LLMs queried with the byte-identical typed request through the official System-One
adapter fall below it on all seven datasets, and reader swaps and contamination checks find no
evidence that the input format or memorised data explain the gain.
For new learners the advantage holds from their first interactions, whereas on unseen items
with all learners logged, deep KT remains ahead.
\end{abstract}

\section{Introduction}
\label{sec:intro}

Knowledge tracing (KT) predicts whether a learner will answer the next item correctly from
the learner's interaction history, but new courses and platforms begin with few logged learners. Deep KT models (DLKT;
\citealp{piech_2015_dkt,ghosh_2020_akt,liu_2023_simplekt}) learn item and skill representations
from many learner sequences, and with little data they can lose to simpler
models~\citep{gervet_2020_deep_kt_when}. LLM-based KT brings prior knowledge, but the LLM generates the answer, either after being
fine-tuned on the target data~\citep{jung_2025_clst_jedm} or after reasoning,
with the probability read from ten sampled traces~\citep{lee_2026_thinkingkt}, which costs ten
calls per prediction and allows only eleven probability values. We call such methods, in which
the LLM generates the answer by reasoning, voting or decoding a label, \emph{System-Two} LLM-KT,
after the slow, deliberate mode of thinking of \citet{kahneman_2011_thinking}.

We ask whether an off-the-shelf System-One LLM can already perform KT when only a
few learners are logged. KT asks a typed question, whether this learner will answer the next
item correctly, and a \emph{System-One} model such as Jev returns the probability of the answer
to a typed question about a state directly, in a single pass, without generating text,
sampling or reasoning traces. We study whether such a model can predict with no
learner logged (RQ1), how long it stays ahead as learners are logged (RQ2), and whether the
advantage holds for new learners and new items (RQ3).

We pose KT to Jev as one typed question (Figure~\ref{fig:pipeline}) and vary only what it contains, namely the learner's recent
history alone (\JevZ{}), which uses no data from the target platform, or the history plus
evidence from the $N$ logged learners (JevKT, with 64 solved examples on the same knowledge
components and the accuracy of similar learners on the target item), the input selected by an
exploration whose selection rule was fixed before test was evaluated. On seven public datasets we compare both with the best of
\nDLKT{} deep KT models and with LLM-based KT (LOKT, Thinking-KT, CLST) in three cold-start settings
(few training learners, new learners, new items), always on identical targets.

With no learners logged, \JevZ{} reaches a mean AUC of \avgPriorZero, above the best deep KT
model trained on 8 learners (\avgDlktEight) and above Thinking-KT on all seven datasets. With
evidence from the logged learners, JevKT stays significantly ahead of the best deep KT model up
to \nSig{} learners (\csDeltaEight{} AUC at $N{=}8$, \csWinsEight{} datasets) and on average up to
\nLead; supervised KT catches up between \nLead{} and \nCross{} learners. Among the readers we
tested, the gain comes from the model, as other LLMs reading the identical input, directly or
through the vendor's System-One adapter, are worse on every dataset, and contamination checks
find no sign of memorised data. For the Qwen3.5-9B backbone of Thinking-KT, reasoning does not
help, as it is never significantly better than the same model without reasoning.

\paragraph{Contributions.}
\begin{itemize}[leftmargin=*,nosep]
\item \textbf{Finding.} With no learner logs, an off-the-shelf System-One LLM (Jev) outperforms deep KT trained on 8 learners on average and prompted System-Two LLM-KT on all seven datasets (RQ1).
\item \textbf{Where it holds.} A cold-start evaluation on identical targets (7 datasets; few training learners, new learners, new items) shows how long the lead lasts as learners are logged (RQ2, RQ3); a training-free input recipe (JevKT), whose gain comes mostly from one similar-learner statistic, keeps it until supervised KT takes over.
\item \textbf{Why.} Controlled analyses (reader swap, other LLMs behind the same interface, contamination checks) find that the gain comes from the model rather than from the typed format or memorised data.\footnote{Code and result summaries are available at \url{https://anonymous.4open.science/r/jevkt-coldstart}.}
\end{itemize}

\section{Related Work}
\label{sec:related}

\subsection{Cold-Start KT}
\label{sec:rw-coldstart}
Deep KT (DLKT) models~\citep{piech_2015_dkt,ghosh_2020_akt,liu_2023_simplekt,liu_2022_pykt}
learn learner and item representations from large interaction logs. When such logs are scarce,
they lose this advantage, and with little data even logistic regression outperforms
DKT~\citep{gervet_2020_deep_kt_when}. Cold-start KT has therefore been studied in three forms.
Some methods train with only a few learners from the target platform~\citep{jung_2025_clst_jedm,kim_2024_lokt},
others predict for new students who have answered only a few
items~\citep{bhattacharjee_2026_mamlkt,lbath_2026_livekt_tabpfn}, and others transfer a model
across platforms~\citep{xie_2024_dgkt,duan_2026_ragkt}. We follow the few-learner protocol of
CLST~\citep{jung_2025_clst_jedm} and, for every number of training learners $N$, compare against the
best of \nDLKT{} DLKT models.

\subsection{System-Two and System-One LLMs for KT}
\label{sec:rw-system}
LLMs can bring general knowledge about the items when learner logs are missing. Given only a
raw interaction history, however, a prompted LLM predicts close to
chance~\citep{neshaei_2024_llm_learner_performance,worden_2026_foundationalassist}. Prompts become
competitive once they also carry statistics about other learners~\citep{kim_2024_lokt,
duan_2025_hisekt,li_2026_merit}. Thinking-KT goes a step further, as the model reasons before it
predicts, and the probability is read as the vote frequency over ten sampled reasoning
traces~\citep{lee_2026_thinkingkt}, as in self-consistency decoding~\citep{wang_2023_selfconsistency}.
All of these are System-Two LLM-KT in our sense, since the LLM generates the answer. They are either prompted, like LOKT and Thinking-KT,
or trained on the target logs, like CLST~\citep{jung_2025_clst_jedm} and
LLM-KT~\citep{wang_2026_llmkt_acl_findings}, which fine-tune an LLM on learner interactions. The
gains from reasoning are mixed at larger scale~\citep{worden_2026_foundationalassist}, and
calibration is rarely reported in LLM-KT, although LLM confidence and proper scoring rules are
well studied~\citep{kadavath_2022_lms_know_what_they_know,gneiting_2007_proper_scoring_rules,
damani_2025_rlcr}. A System-One model takes a different route and returns the probability of each answer to a
typed question about a learner's state directly, in a single pass. We study Jev\footnote{\url{https://typesafe.ai/blog/introducing-system-one-models-and-jev}}, a commercial model of this kind, on KT.

\section{Posing KT to a System-One LLM}
\label{sec:method}

We neither train nor modify Jev. We pose KT as one typed question and vary only what the
question contains, namely the learner state alone (\JevZ{}), which probes what the model knows
without any data from the target platform, or the state plus evidence from the logged
learners (JevKT), the input recipe selected by the exploration described below
(Appendix~\ref{app:explore}).

\paragraph{Task.} A learner's history is $X_{<t}=((q_1,r_1),\dots,(q_{t-1},r_{t-1}))$ of items
$q_i$ with responses $r_i\in\{0,1\}$, and $\mathcal{K}(q)$ is the KC set of item $q$. KT predicts
$p_t=P(r_t{=}1\mid X_{<t},q_t)$. In cold start only a pool $\mathcal{U}$ of $N$ training learners
is logged.

\paragraph{\JevZ{}.} A frozen System-One model $f$ (Jev) returns the probability of ``yes''
to a typed question about the learner's last $H{=}25$ interactions $X_{t-H:t}$,
\begin{equation}
p^{0}_t = f\big(X_{t-H:t},\ q_t\big).
\end{equation}

\paragraph{JevKT.} JevKT asks the same question with evidence from the training pool $\mathcal{U}$,
\begin{equation}
p_t = f\big(\mathcal{S}_t,\ X_{t-H:t},\ s_t,\ q_t\big),
\end{equation}
where $\mathcal{S}_t$ holds $k{=}64$ solved
examples from training learners whose item shares a KC with $q_t$,
\begin{equation}
\mathcal{S}_t \subset \{(q^{v}_j,r^{v}_j) : v\in\mathcal{U},\, \mathcal{K}(q^{v}_j)\cap\mathcal{K}(q_t)\neq\emptyset\},
\end{equation}
each shown with $v$'s last 10 interactions before $j$,
and $s_t$ is the first-attempt accuracy on $q_t$ of the learners most similar to $u$,
\begin{equation}
s_t=\frac{1}{n}\sum_{v\in\mathcal{N}_n(u)} r^{v}_{q_t},
\end{equation}
where $\mathcal{N}_n(u)$ are the $n{=}20$ training learners with the smallest weighted
$\ell_1$ distance $\lVert\mathbf{m}_u-\mathbf{m}_v\rVert_{1,w}$ between smoothed per-KC accuracy
vectors ($\mathbf{m}_u$ from $X_{<t}$ only; $w$ = how often $u$ practised each KC). $s_t$ enters the prompt as one
sentence; if fewer than five neighbours answered $q_t$, the KCs of $q_t$ are used instead.
All examples and statistics come from $\mathcal{U}$
and the target's past; a leak audit (including a test that the input is unchanged when $r_t$
is flipped) passes in every setting.

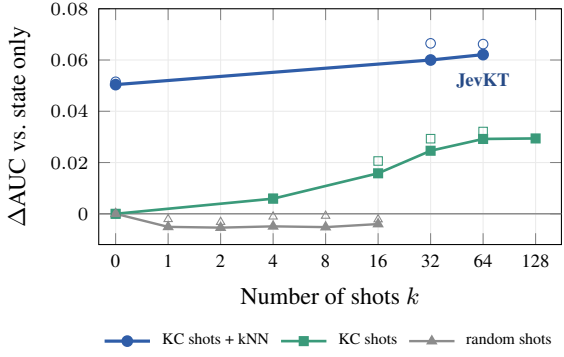
\begin{figure}[t]
\centering
\begin{tikzpicture}
\begin{axis}[
  width=\columnwidth, height=4.7cm, scaled y ticks=false,
  xmode=log, log basis x=2, xmin=0.4, xmax=180,
  xtick={0.5,1,2,4,8,16,32,64,128}, xticklabels={0,1,2,4,8,16,32,64,128},
  ymin=-0.012, ymax=0.08, ytick distance=0.02,
  yticklabel style={/pgf/number format/fixed, /pgf/number format/precision=2},
  tick label style={font=\scriptsize}, label style={font=\footnotesize},
  xlabel={Number of shots $k$}, ylabel={$\Delta$AUC vs.\ state only},
  grid=major, grid style={black!8},
  legend style={font=\tiny, draw=none, at={(0.5,-0.33)}, anchor=north, legend columns=3, column sep=2pt},
]
\draw[black!50, line width=0.5pt] (axis cs:0.4,0) -- (axis cs:180,0);
\addplot[cJev, mark=*, mark size=1.6pt, line width=1.2pt] coordinates {(0.5,0.0504) (32,0.0600) (64,0.0621)};
\addlegendentry{KC shots + kNN}
\addplot[cAnch, mark=square*, mark size=1.4pt, line width=1pt] coordinates {(0.5,0.0000) (4,0.0059) (16,0.0158) (32,0.0246) (64,0.0292) (128,0.0294)};
\addlegendentry{KC shots}
\addplot[cKNN, mark=triangle*, mark size=1.6pt, line width=0.9pt] coordinates {(0.5,0.0000) (1,-0.0051) (2,-0.0054) (4,-0.0049) (8,-0.0052) (16,-0.0040)};
\addlegendentry{random shots}
\addplot[cJev, only marks, mark=o, mark size=1.8pt, forget plot] coordinates {(0.5,0.0515) (32,0.0665) (64,0.0662)};
\addplot[cAnch, only marks, mark=square, mark size=1.6pt, forget plot] coordinates {(0.5,0.0000) (4,0.0060) (16,0.0206) (32,0.0293) (64,0.0321)};
\addplot[cKNN, only marks, mark=triangle, mark size=1.8pt, forget plot] coordinates {(0.5,0.0000) (1,-0.0021) (2,-0.0032) (4,-0.0011) (8,-0.0008) (16,-0.0022)};
\node[font=\scriptsize, text=cJev!80!black, anchor=north, yshift=-3pt] at (axis cs:64,0.0621) {\bfseries JevKT};
\end{axis}
\end{tikzpicture}
\caption{How JevKT was selected: mean $\Delta$AUC over the state-only input (7 datasets) vs.\
number of shots $k$. Filled: development split (used for selection); hollow: test (not used for selection).
Other inputs: Appendix~\ref{app:explore}.}
\label{fig:shots}
\end{figure}

\paragraph{How JevKT was selected.}\label{sec:selection} We explored history length, random vs.\ KC-matched shots
and their number, population statistics vs.\ the similar-learner statistic, repeated calls,
state formats, paraphrased questions and LOKT option weights (Appendix~\ref{app:explore}),
and fixed the selection rule (best mean $\Delta$AUC on the development split) before evaluating test. Figure~\ref{fig:shots} shows the decisive
axis. KC-matched shots help and saturate at $k{\approx}64$, random shots
do not, and the similar-learner statistic adds on top; the ranking on the development split transfers to test.

\section{Experiments}
\label{sec:experiments}
\begin{table*}[t]
\centering
\small
\setlength{\tabcolsep}{4pt}
\caption{Cold-start AUC (P1) of Jev against the best deep KT and the best LLM-based KT baseline with $N$
training learners (mean over 5 seeds). Without training data, \JevZ{} is above LLM-based KT on all seven
datasets and above deep KT trained on 8 learners on average; JevKT stays ahead of deep KT up to
$N{=}\nLead$ on average. \emph{Wins}: datasets on which Jev beats every baseline in its block.
$^\dagger$Best of \nDLKT{} pyKT models, selected on test. $^\ddagger$LOKT, the strongest LLM-based KT
baseline. $^\circ$$N{=}0$ value repeated (no answer options). $^\S$Reported by LLM-KT~\citep{wang_2026_llmkt_acl_findings}
on its own split; not ranked. All models: Appendix~\ref{app:fulltables}. Best in \textbf{bold}, second best \underline{underlined}.}
\label{tab:main-cs}
\resizebox{\textwidth}{!}{%
\begin{tabular}{cl ccccccc cc}
\toprule
$N$ & \textbf{Method} & \textbf{ASSIST09} & \textbf{NIPS34} & \textbf{XES3G5M} & \textbf{DBE-KT22}
& \textbf{Alg05} & \textbf{Junyi} & \textbf{ASSIST17} & \textbf{Avg} & \textbf{Wins} \\
\midrule
 & Best deep KT$^\dagger$ & -- & -- & -- & -- & -- & -- & -- & -- &  \\
 & Best LLM-based KT$^\ddagger$ & \underline{.715} & \underline{.662} & \underline{.699} & \underline{.688} & \underline{.656} & \underline{.671} & \underline{.608} & \underline{.671} &  \\
\rowcolor{focusbg}
\multirow{-3}{*}{0} & \JevZ{} & \textbf{.738} & \textbf{.691} & \textbf{.725} & \textbf{.742} & \textbf{.698} & \textbf{.700} & \textbf{.648} & \textbf{.706} & 7/7 \\
\midrule
 & Best deep KT$^\dagger$ & .700 & .653 & .689 & \underline{.793} & \underline{.711} & .652 & \underline{.624} & \underline{.689} &  \\
 & Best LLM-based KT$^\ddagger$ & \underline{.707} & \underline{.662} & \underline{.699}$^\circ$ & .699 & .656$^\circ$ & \underline{.671}$^\circ$ & .608$^\circ$ & .672 &  \\
\rowcolor{focusbg}
\multirow{-3}{*}{8} & JevKT & \textbf{.735} & \textbf{.707} & \textbf{.720} & \textbf{.821} & \textbf{.734} & \textbf{.701} & \textbf{.638} & \textbf{.722} & 7/7 \\
\midrule
 & Best deep KT$^\dagger$ & .706 & \underline{.696} & \underline{.706} & \underline{.808} & \underline{.718} & \underline{.691} & \underline{.626} & \underline{.707} &  \\
 & Best LLM-based KT$^\ddagger$ & \underline{.707} & .663 & .699$^\circ$ & .701 & .656$^\circ$ & .671$^\circ$ & .608$^\circ$ & .672 &  \\
\rowcolor{focusbg}
\multirow{-3}{*}{16} & JevKT & \textbf{.737} & \textbf{.734} & \textbf{.720} & \textbf{.834} & \textbf{.738} & \textbf{.704} & \textbf{.646} & \textbf{.730} & 7/7 \\
\midrule
 & Best deep KT$^\dagger$ & \underline{.724} & \underline{.732} & \underline{.708} & \underline{.816} & \underline{.744} & \underline{.697} & \underline{.639} & \underline{.723} &  \\
 & Best LLM-based KT$^\ddagger$ & .707 & .660 & .699$^\circ$ & .701 & .656$^\circ$ & .671$^\circ$ & .608$^\circ$ & .672 &  \\
\rowcolor{focusbg}
\multirow{-3}{*}{32} & JevKT & \textbf{.737} & \textbf{.752} & \textbf{.728} & \textbf{.838} & \textbf{.751} & \textbf{.709} & \textbf{.656} & \textbf{.739} & 7/7 \\
\midrule
 & Best deep KT$^\dagger$ & \underline{.734} & \underline{.750} & \underline{.720} & \textbf{.846} & \textbf{.777} & \underline{.704} & \textbf{.668} & \underline{.743} &  \\
 & Best LLM-based KT$^\ddagger$ & .703 & .660 & .699$^\circ$ & .699 & .656$^\circ$ & .671$^\circ$ & .608$^\circ$ & .671 &  \\
\rowcolor{focusbg}
\multirow{-3}{*}{64} & JevKT & \textbf{.745} & \textbf{.762} & \textbf{.739} & \underline{.838} & \underline{.762} & \textbf{.715} & \underline{.667} & \textbf{.747} & 4/7 \\
\midrule
 & Best deep KT$^\dagger$ & \underline{.739} & \textbf{.768} & \underline{.739} & \textbf{.852} & \textbf{.792} & \underline{.708} & \textbf{.692} & \textbf{.756} &  \\
 & Best LLM-based KT$^\ddagger$ & .704 & .658 & .699$^\circ$ & .697 & .656$^\circ$ & .671$^\circ$ & .608$^\circ$ & .670 &  \\
\rowcolor{focusbg}
\multirow{-3}{*}{128} & JevKT & \textbf{.745} & \underline{.764} & \textbf{.747} & \underline{.837} & \underline{.774} & \textbf{.727} & \underline{.680} & \underline{.753} & 3/7 \\
\midrule
 & Best deep KT$^\dagger$ & \textbf{.746} & \textbf{.774} & \textbf{.757} & \textbf{.852} & \textbf{.808} & \underline{.723} & \textbf{.705} & \textbf{.766} &  \\
 & Best LLM-based KT$^\ddagger$ & \underline{.706} & .659 & .699$^\circ$ & .692 & .656$^\circ$ & .671$^\circ$ & .608$^\circ$ & .670 &  \\
\rowcolor{focusbg}
\multirow{-3}{*}{256} & JevKT & \textbf{.746} & \underline{.769} & \underline{.752} & \underline{.838} & \underline{.779} & \textbf{.738} & \underline{.687} & \underline{.758} & 1/7 \\
\midrule
 & Best deep KT$^\dagger$ & \textbf{.794} & \textbf{.789} & \textbf{.791} & \textbf{.862} & \textbf{.814} & \textbf{.794} & \textbf{.743} & \textbf{.798} &  \\
 & Best LLM-based KT$^\ddagger$ & .705 & .666 & .699$^\circ$ & .700 & .656$^\circ$ & .671$^\circ$ & .608$^\circ$ & .672 &  \\
\rowcolor{focusbg}
\multirow{-3}{*}{all} & JevKT & \underline{.780} & \underline{.774} & \underline{.758} & \underline{.829} & \underline{.780} & \underline{.771} & \underline{.701} & \underline{.770} & 0/7 \\
 & LLM-KT (reported) & \textit{.887}$^\S$ & -- & -- & -- & -- & -- & -- & -- &  \\
\bottomrule
\end{tabular}}
\end{table*}

\subsection{Experimental Setup}
\label{sec:setup}

\paragraph{Datasets.} We use seven public KT datasets that cover different subjects, item
granularities and base rates, namely ASSIST2009~\citep{feng_2009_assistments}, NIPS34~\citep{wang_2020_eedi}, XES3G5M~\citep{liu_2023_xes3g5m},
DBE-KT22~\citep{abdelrahman_2022_dbekt22}, Algebra2005 from the KDD Cup 2010~\citep{stamper_2016_kddcup},
Junyi\footnote{Junyi Academy 2018--19 release, \url{https://www.kaggle.com/datasets/junyiacademy/learning-activity-public-dataset-by-junyi-academy}} and ASSIST2017~\citep{patikorn_2020_assist2017}. Learners are split
80/20 at the learner level; statistics and preprocessing are in Appendix~\ref{app:data}.

\paragraph{Cold-start settings.} Following the cold-start KT literature
\citep{jung_2025_clst_jedm,bhattacharjee_2026_mamlkt}, we evaluate three settings on identical
targets for all methods. In \textbf{P1 (few training learners)}, only
$N\in\{8,16,32,64,128,256,\text{all}\}$ training learners are available (nested subsets,
5 seeds), and we score the first 50 interactions of held-out learners (2{,}000 targets per
dataset); $N{=}0$ means no data from the target platform. In \textbf{P2 (new learners)}, the
same predictions with $N{=}8$ are scored by the position of the target in the new learner's
sequence. In \textbf{P3 (new items)}, 20\% of the items are removed from training and from
all examples and statistics, and only targets on these items are scored (3 seeds).

\paragraph{Baselines.} We compare against (i) supervised deep KT, with \nDLKT{} models of the pyKT
library~\citep{liu_2022_pykt}, eleven core models (including DKT~\citep{piech_2015_dkt},
AKT~\citep{ghosh_2020_akt} and simpleKT~\citep{liu_2023_simplekt}) plus \nDLKTadded{} further ones (full
list in Appendix~\ref{app:impl}), reporting for every cell the best model chosen
\emph{on test}, which favours DLKT; (ii) System-Two LLM-KT, either prompted without training,
LOKT~\citep{kim_2024_lokt} and Thinking-KT~\citep{lee_2026_thinkingkt} with Qwen3.5-9B, or trained on
the target data, namely CLST~\citep{jung_2025_clst_jedm}, which we re-implement on the three datasets of the original
paper, and LLM-KT~\citep{wang_2026_llmkt_acl_findings}, for which we cite the reported ASSIST09 value; and
(iii) the classical KT models BKT~\citep{corbett_1995_bkt} and PFA~\citep{pavlik_2009_pfa}. Details are in Appendix~\ref{app:impl}.

\paragraph{Metrics.} We report AUC as the main metric (ACC and F1 in
Appendix~\ref{app:coldstart}), ECE and Brier for calibration, and latency and cost per
million predictions (GPU time priced at a public rental rate). Differences are tested with a one-sided paired learner-cluster bootstrap with
Holm correction over datasets (Appendix~\ref{app:holm}). JevKT uses
\texttt{jev-1.13-20260917} via OpenRouter with $H{=}25$, $k{=}64$ and $n{=}20$
(Appendix~\ref{app:impl}).

\subsection{Main Results}
\label{sec:main-results}
\begin{figure*}[t]
\centering
\resizebox{\textwidth}{!}{%
\begin{tikzpicture}
\begin{groupplot}[
  group style={group size=2 by 1, horizontal sep=1.4cm},
  width=7.2cm, height=4.4cm,
  yticklabel style={/pgf/number format/fixed, /pgf/number format/precision=2},
  tick label style={font=\scriptsize}, label style={font=\footnotesize},
  title style={font=\footnotesize\bfseries, yshift=-1mm},
  grid=major, grid style={black!8},
]
\nextgroupplot[title={(a) P2: new learners, $N{=}8$}, xmin=0.6, xmax=5.4, ymin=0.56, ymax=0.76,
  xtick={1,2,3,4,5}, xticklabels={2--5,6--10,11--15,16--30,31--49},
  xlabel={Position of the target in the learner's sequence}, ylabel={Mean AUC (7 datasets)}]
\addplot[cDLKT, mark=square*, mark size=1.3pt, line width=1.1pt] coordinates {(1,0.6908) (2,0.6950) (3,0.6990) (4,0.6779) (5,0.6988)};
\addplot[cJev, mark=*, mark size=1.6pt, line width=1.4pt] coordinates {(1,0.7136) (2,0.7260) (3,0.7228) (4,0.7246) (5,0.7276)};
\addplot[cJev, dashed, line width=1.1pt] coordinates {(1,0.6873) (2,0.7122) (3,0.7095) (4,0.7038) (5,0.7172)};
\nextgroupplot[title={(b) P3: new items}, ybar, bar width=12pt, ymin=0.6, ymax=0.76,
  symbolic x coords={LOKT,DLKT,Jev0,JevKT}, xtick={LOKT,DLKT,Jev0,JevKT},
  xticklabels={{LOKT},{Best\\DLKT$^\dagger$},{Jev\\(0-shot)},{JevKT}},
  xticklabel style={align=center, font=\scriptsize}, enlarge x limits=0.25,
  xlabel={Held-out 20\% of items (all training learners)},
  nodes near coords, nodes near coords style={font=\tiny}, point meta=y,
  every node near coord/.append style={/pgf/number format/fixed, /pgf/number format/precision=3}]
\addplot[fill=cKNN!35, draw=cKNN, bar shift=0pt] coordinates {(LOKT,0.6640)};
\addplot[fill=cDLKT!45, draw=cDLKT, bar shift=0pt] coordinates {(DLKT,0.7290)};
\addplot[fill=cJev!25, draw=cJev, bar shift=0pt] coordinates {(Jev0,0.7040)};
\addplot[fill=cJev!55, draw=cJev, bar shift=0pt] coordinates {(JevKT,0.7130)};

\end{groupplot}
\node[anchor=south, font=\footnotesize] at ($(group c1r1.north west)!0.5!(group c2r1.north east) + (0,0.55)$) {%
  \tikz\draw[cJev, line width=1.4pt] (0,0)--(0.35,0) node[midway,fill=cJev,circle,inner sep=1.2pt]{}; JevKT (Jev + 64-shot + kNN) \quad
  \tikz\draw[cJev, dashed, line width=1.1pt] (0,0)--(0.35,0); \JevZ{} ($N{=}0$) \quad
  \tikz\draw[cDLKT, line width=1.1pt] (0,0)--(0.35,0) node[midway,fill=cDLKT,inner sep=1.4pt]{}; Best DLKT$^\dagger$};
\end{tikzpicture}}
\caption{Cold start for new learners and new items (mean AUC over 7 datasets; few training
learners, P1, is in Table~\ref{tab:main-cs}). (a) P2: AUC by the position of the target in a new
learner's sequence with $N{=}8$ training learners; \JevZ{} uses no training learners. (b) P3:
targets on items held out of training. All methods are scored on identical targets; per-dataset
results in Appendix~\ref{app:coldstart}. $^\dagger$Best of \nDLKT{} pyKT models selected on
test per cell, which favours DLKT.}
\label{fig:main}
\end{figure*}
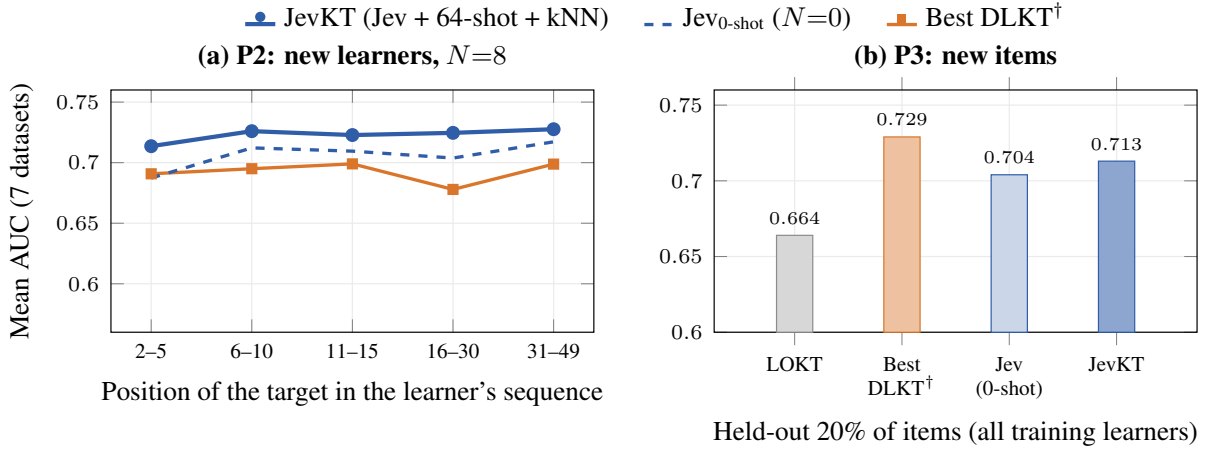

\paragraph{No learners logged, $N{=}0$ (RQ1).}
Without any data from the target platform, \JevZ{} reaches \avgPriorZero{} mean AUC (first block of
Table~\ref{tab:main-cs}). This is above the strongest
training-free LLM-KT baseline, LOKT (.671), and above the best deep KT model trained on 8
learners (\avgDlktEight; higher on \zeroWinsEight{} datasets). It also beats System-Two Thinking-KT
(.650) on all seven datasets with one call instead of ten, at about 1/100 of its
API cost (Section~\ref{sec:cost}). Thinking-KT, CLST and every $N$ are listed in
Tables~\ref{tab:p1n0}--\ref{tab:p1nall} (Appendix~\ref{app:coldstart}).

\paragraph{P1, few training learners (RQ2).}
Table~\ref{tab:main-cs} follows the comparison as learners are logged (curves in Appendix~\ref{app:coldstart}).
JevKT is ahead of the best deep KT model by \csDeltaEight, \csDeltaSixteen{} and
\csDeltaThirtyTwo{} mean AUC at $N{=}8$, 16 and 32 (on \csWinsEight, \csWinsSixteen{} and
\csWinsThirtyTwo{} datasets); the mean difference is \csDeltaSixtyFour{} at $N{=}64$,
\csDeltaOneTwentyEight{} at $N{=}128$, \csDeltaTwoFiftySix{} at $N{=}256$ and \csDeltaFull{} with
the full training data. LOKT, whose only use of the logged learners is answer-option weights,
barely moves with $N$ (\loktEight{} at $N{=}8$ and \loktFull{} with all learners, against .671
at $N{=}0$) and stays below both. The crossover therefore lies between \nLead{} and \nCross{}
learners on average and varies by dataset (Figure~\ref{fig:crossover}, Appendix~\ref{app:coldstart}). The lead is Holm-significant on \sigTestEight{} datasets at $N{=}8$ and
\sigTestSixteen{} at $N{=}16$, \sigTestThirtyTwo{} at $N{=}32$ and \sigTestSixtyFour{} at $N{=}64$
(Table~\ref{tab:holm}). Because the best deep KT model is chosen on
test in every cell, these margins are conservative.

\paragraph{P2, new learners (RQ3).}
Figure~\ref{fig:main}a scores the same $N{=}8$ predictions by the position of the target in a
new learner's sequence. JevKT is above the best deep KT model at every position, already for
the second to fifth interactions (\ptwoJevEarly{} vs.\ \ptwoDlktEarly), when the new learner has shown
little evidence, and in \ptwoWins{} dataset--position cells (Table~\ref{tab:p2}).

\paragraph{P3, new items (RQ3).}
Figure~\ref{fig:main}b restricts the targets to items held out of training and of all examples
and statistics. Here every training learner is available, so this setting isolates new items
rather than few learners. JevKT (\pthreeJev) and \JevZ{} (\pthreeZero) stay well above LOKT
(\pthreeLokt), but the best deep KT model, trained on all learners and chosen on test, is ahead
(\pthreeDlkt; JevKT is better on \pthreeWins{} datasets; Table~\ref{tab:p3}). Unseen items alone
therefore do not reverse the full-data result, and the advantage of the System-One model comes from
having few learners, not from unseen items.

\section{Analysis}
\label{sec:analysis}

\subsection{Is it the model or the format?}
\label{sec:model-format}
To separate the model from the format, we first hold the Qwen3.5-9B weights fixed
(Figure~\ref{fig:modelformat}; Table~\ref{tab:modelformat}). A typed one-pass head without
training (B0) is worse than reading the label probability directly (A), .630 vs.\
.688 AUC, and .701 vs.\ .752 with the similar-learner statistic. Distilling
Jev into these weights (C, JEV-9B) does not beat A (.679 and .741), and the same
holds at $N{=}8$ (A .700, B0 .659, C .699). Conversely, when the identical
serialised input is read by Jev instead of Qwen3.5-9B, AUC rises in all 35 cells
(7 datasets $\times$ 5 conditions), e.g.\ from .727 to .776 with the JevKT input.
Rows A--C use the state with the last 25 interactions and the similar-learner statistic,
the swap rows the JevKT state with 64 shots. On the A state itself Jev reaches .761 against
.752 for A, higher on all seven datasets (by .002 to .021); the longer JevKT state helps Jev
(.776) but not Qwen3.5-9B (.727).
For the weights and readers we tested, the typed format does not produce the gain; the model does.

No other commercial System-One model is available, so we also send the byte-identical typed
request to three other LLMs through TypeSafe's official System-One adapter
(Table~\ref{tab:generality-body}; per dataset in Table~\ref{tab:generality},
Appendix~\ref{app:generality}). At $N{=}0$, GPT-4o-mini (\genGptZero), Gemini-2.5-Flash-Lite
(\genGemZero) and DeepSeek-V4-Flash (\genDsZero) are below \JevZ{} (\avgPriorZero) on every
dataset (paired bootstrap, $p{<}.01$). On identical targets the JevKT input adds a similar,
small amount to Jev (\gainJev) and to DeepSeek (\gainDs), significant on \gainJevSig{} and
\gainDsSig{} datasets, and the two gains never differ significantly; the gap between the models
(\gapZeroMatched{} at $N{=}0$) is set by the model, not by the input. Only Jev and the
Jev-distilled JEV-9B (\genNineBZero{} at $N{=}0$, \genNineBEight{} at $N{=}8$) come close to or
exceed the best deep KT model trained on 8 learners (\avgDlktEight), while Laya, an open-source
model that reproduces Jev's typed interface with a small encoder, stays near chance (\genLayaZero).
\begin{figure}[t]
\centering
\begin{tikzpicture}
\begin{axis}[
  width=\columnwidth, height=4.6cm,
  ybar=1pt, bar width=5.5pt,
  ymin=0.6, ymax=0.8,
  symbolic x coords={A,B0,C,Swap,Jev},
  xtick=data,
  xticklabels={{A\\\tiny plain}, {B0\\\tiny typed head}, {C\\\tiny JEV-9B}, {Swap\\\tiny Qwen reads}, {Jev\\\tiny commercial}},
  xticklabel style={align=center, font=\scriptsize},
  yticklabel style={/pgf/number format/fixed, /pgf/number format/precision=2, font=\scriptsize},
  ylabel={\footnotesize Mean AUC}, enlarge x limits=0.14,
  grid=major, grid style={black!8},
  legend style={font=\tiny, draw=none, at={(0.02,0.97)}, anchor=north west, legend columns=3},
]
\addplot[fill=cKNN!35, draw=cKNN] coordinates {(A,0.6877) (B0,0.6300) (C,0.6794) (Swap,0.6886) (Jev,0.7099)};
\addlegendentry{Full, $H{=}25$}
\addplot[fill=cJev!40, draw=cJev] coordinates {(A,0.7517) (B0,0.7010) (C,0.7409) (Swap,0.7273) (Jev,0.7760)};
\addlegendentry{Full, + evidence}
\addplot[fill=cDLKT!40, draw=cDLKT] coordinates {(A,0.7002) (B0,0.6588) (C,0.6994) (Swap,0.6989) (Jev,0.7270)};
\addlegendentry{Cold start $N{=}8$}
\end{axis}
\end{tikzpicture}
\caption{Model or format? On the same Qwen3.5-9B weights, a typed one-pass head (B0) is
worse than the plain readout (A) and Jev distillation (C) does not beat A, both with full
data and in cold start; on the identical state, Jev is the stronger reader (35/35 cells).}
\label{fig:modelformat}
\end{figure}
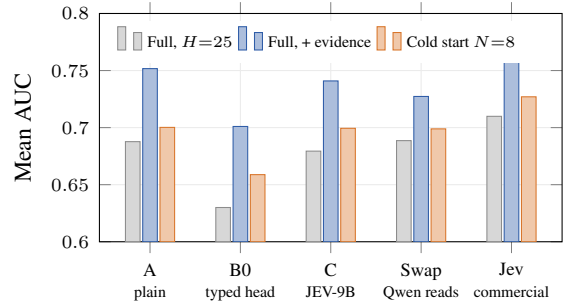
\begin{table}[t]
\centering
\footnotesize
\setlength{\tabcolsep}{3pt}
\caption{The same typed request read by other models (mean AUC over 7 datasets). \emph{Below
Jev}: datasets on which the reader is below \JevZ{} at $N{=}0$. $N{=}8$ rows of Jev and the
adapter use a 1{,}000-target subsample (budget). $^\dagger$Trained on the 8 learners.
Per dataset: Table~\ref{tab:generality}. Best in \textbf{bold}, second best \underline{underlined}.}
\label{tab:generality-body}
\begin{tabular}{l ccc}
\toprule
\textbf{Reader} & $N{=}0$ & $N{=}8$ & \textbf{Below Jev} \\
\midrule
\rowcolor{focusbg} Jev & \textbf{.706} & \textbf{.717} & -- \\
Adapter + GPT-4o-mini & .680 & -- & 7/7 \\
Adapter + Gemini-2.5-FL & .670 & -- & 7/7 \\
Adapter + DeepSeek-V4-F & .626 & .634 & 7/7 \\
JEV-9B (Jev-distilled) & \underline{.687} & \underline{.699} & 6/7 \\
Laya (open-source Jev) & .536 & .553 & 7/7 \\
\midrule
Best DLKT$^\dagger$ & -- & .689 & -- \\
\bottomrule
\end{tabular}
\end{table}

\subsection{What do the logged learners add?}
\label{sec:prior-evidence}
JevKT combines two sources, the model's prior, which \JevZ{} measures alone, and evidence from
the logged learners. At $N{=}8$ the prior alone reaches \avgPriorZero, the similar-learner
statistic alone \avgEvidenceEight, and both together \avgBothEight, against \avgDlktEight{} for
the best deep KT model (Table~\ref{tab:main-cs}). With few learners the prior carries the
prediction and the evidence adds to it; as $N$ grows the evidence becomes reliable, and
supervised KT, which learns from the same evidence, catches up. On identical targets the
evidence adds \gainJev{} AUC to Jev, significantly on \gainJevSig{} datasets (DBE-KT22 and
Algebra05), so its contribution is real but small. Figure~\ref{fig:shots} and
Appendix~\ref{app:explore} show which parts of the input matter.
Under cold start the two parts of the evidence overlap (Table~\ref{tab:ablation-n16},
Appendix~\ref{app:explore}). At $N{=}16$ the similar-learner statistic alone raises \JevZ{} from
\abZero{} to \abKnn, 64 KC-matched shots alone to \abKCsixtyfour, and both together to
\abFinal. The statistic therefore carries almost all of the gain, and the shots add to it
clearly only on DBE-KT22; we keep both because they never hurt and match the configuration
selected on full data.
Where exactly supervised KT takes over is hard to predict, since a model of the crossover fitted on
synthetic datasets does not transfer to the real ones (Appendix~\ref{app:coldstart}).
The gain is not confined to easy cases (Table~\ref{tab:errors}, Appendix~\ref{app:errors}). At
$N{=}8$ JevKT leads both on items that the logged learners answered (\errItemSeen) and on items
none of them answered (\errItemUnseen), and the lead is largest for learners whose accuracy so far
is very low or very high (\errAccLow{} and \errAccHigh), where eight learners give a supervised
model little to learn from.

\subsection{Why does System-Two LLM-KT fall behind?}
\label{sec:calibration}
\begin{figure}[t]
\centering
\begin{tikzpicture}
\begin{groupplot}[
  group style={group size=2 by 1, horizontal sep=0.35cm, y descriptions at=edge left},
  width=4.6cm, height=4.6cm, xmin=0, xmax=1, ymin=0, ymax=1,
  xtick={0,0.5,1}, ytick={0,0.5,1}, tick label style={font=\tiny},
  title style={font=\scriptsize\bfseries, yshift=-1.5mm},
  grid=major, grid style={black!8},
  legend style={font=\tiny, draw=none, fill=white, fill opacity=0.8, text opacity=1, at={(0.97,0.03)}, anchor=south east},
]
\nextgroupplot[title={(a) No learners ($N{=}0$)}, ylabel={\scriptsize Observed accuracy}]
\addplot[black!40, dashed, domain=0:1, forget plot] {x};
\addplot[cDLKT, mark=square*, mark size=1.1pt, line width=0.8pt] coordinates {(0.000,0.429) (0.000,0.423) (0.038,0.485) (0.198,0.601) (0.424,0.687) (0.638,0.750) (0.773,0.805) (0.886,0.841) (0.975,0.820) (1.000,0.843)};
\addlegendentry{Thinking-KT}
\addplot[cKNN, mark=diamond*, mark size=1.3pt, line width=0.8pt] coordinates {(0.000,0.388) (0.024,0.413) (0.135,0.497) (0.305,0.586) (0.517,0.682) (0.714,0.749) (0.847,0.806) (0.924,0.827) (1.000,0.863) (1.000,0.875)};
\addlegendentry{LOKT}
\addplot[cJev, mark=*, mark size=1.1pt, line width=1.1pt] coordinates {(0.220,0.287) (0.357,0.416) (0.451,0.517) (0.524,0.591) (0.586,0.681) (0.647,0.735) (0.705,0.811) (0.754,0.833) (0.801,0.886) (0.855,0.927)};
\addlegendentry{\JevZ{}}
\nextgroupplot[title={(b) $N{=}8$ learners}]
\addplot[black!40, dashed, domain=0:1, forget plot] {x};
\addplot[cDLKT, mark=square*, mark size=1.1pt, line width=0.8pt] coordinates {(0.277,0.349) (0.421,0.439) (0.521,0.548) (0.593,0.598) (0.665,0.661) (0.733,0.739) (0.777,0.789) (0.820,0.829) (0.878,0.846) (0.944,0.887)};
\addlegendentry{Best DLKT}
\addplot[cJev, mark=*, mark size=1.1pt, line width=1.1pt] coordinates {(0.278,0.294) (0.392,0.400) (0.475,0.499) (0.546,0.584) (0.607,0.666) (0.661,0.751) (0.713,0.805) (0.762,0.849) (0.810,0.888) (0.875,0.950)};
\addlegendentry{JevKT}
\end{groupplot}
\node[font=\scriptsize] at ($(group c1r1.south)!0.5!(group c2r1.south)+(0,-0.55)$) {Predicted probability};
\end{tikzpicture}
\caption{Reliability of the predicted probabilities (7 datasets pooled, 10 equal-mass bins for display; ECE in Table~\ref{tab:calib} uses 15;
diagonal = perfect calibration). Vote frequencies of System-Two LLM-KT pile up at 0 and 1, while
the single-pass readout of Jev follows the diagonal with a slight underconfidence. Per-dataset
ECE and Brier: Table~\ref{tab:calib}.}
\label{fig:reliability}
\end{figure}
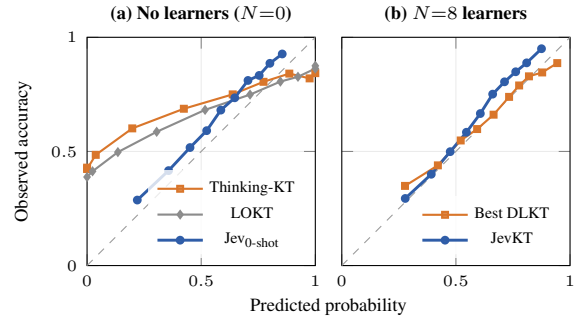

Table~\ref{tab:system} compares Thinking-KT with and without reasoning on the same Qwen3.5-9B.
A reasoning budget of 1024 or 2048 tokens never improves AUC significantly over the same model
without reasoning, while one Jev pass beats Thinking-KT on all seven datasets (Holm-corrected,
$\Delta$AUC +.038 to +.073 against the budget of 2048). Thinking-KT's probability is a vote frequency over
ten samples, so it can take only eleven values and costs ten calls per prediction
(Section~\ref{sec:cost}).
These vote frequencies are also poorly calibrated (Figure~\ref{fig:reliability},
Table~\ref{tab:calib}). They pile up at 0 and 1, and the mean ECE of Thinking-KT is \eceTkt{}
(LOKT \eceLokt), against \eceZero{} for the single \JevZ{} pass. With $N{=}8$ learners, JevKT is
as well calibrated as the best deep KT model (ECE \eceJev{} vs.\ \eceDlkt; Brier \brierJev{} vs.\
\brierDlkt), without any post-hoc scaling such as temperature scaling~\citep{guo_2017_calibration}. Jev's remaining error is a mild underconfidence, as
observed accuracy is slightly above the predicted probability across the range.
Longer reasoning does not help either. Within each dataset, the third of targets with the
longest reasoning has lower AUC than the shortest third (.668 vs.\ .719 at budget 2048; .675
vs.\ .726 at 1024), and reasoning length correlates with the absolute error (Spearman +.29),
also among targets on which all ten votes agree. Long reasoning marks hard cases rather than
fixing them.
\begin{table}[t]
\centering
\footnotesize
\setlength{\tabcolsep}{3pt}
\caption{Reasoning does not help Qwen3.5-9B on KT; a single System-One pass does better.
$\Delta$AUC (Holm-corrected paired bootstrap; $^{*}p{<}.05$).
\emph{Think $-$ No-think}: Thinking-KT with reasoning budget 1024 / 2048 minus the same
model without reasoning. \emph{Jev $-$ Thinking-KT}: one Jev call vs.\ 10 sampled
reasoning generations. \emph{Cost}: per million predictions of Thinking-KT at each budget (no-think: \$707) and of one Jev call.}
\label{tab:system}
\resizebox{\columnwidth}{!}{%
\begin{tabular}{l cc c}
\toprule
& \multicolumn{2}{c}{\textbf{Think $-$ No-think}} & \textbf{Jev $-$ Thinking-KT} \\
\cmidrule(lr){2-3}\cmidrule(lr){4-4}
\textbf{Dataset} & budget 1024 & budget 2048 & (budget 2048) \\
\midrule
ASSIST09   & $-$.013 & $-$.015 & +.038$^{*}$ \\
NIPS34     & +.014   & +.005   & +.038$^{*}$ \\
XES3G5M    & $-$.020 & $-$.007 & +.072$^{*}$ \\
DBE-KT22   & $-$.016 & $-$.026 & +.073$^{*}$ \\
Algebra05  & +.023   & +.036   & +.053$^{*}$ \\
Junyi      & $-$.034 & $-$.029 & +.068$^{*}$ \\
ASSIST17   & $-$.008 & +.010   & +.048$^{*}$ \\
\midrule
Significant & 0/7 & 0/7 & \textbf{7/7} \\
Cost / 1M pred. & \$3{,}279 & \$4{,}360 & \$44 \\
\bottomrule
\end{tabular}}
\end{table}

\subsection{How fast and cheap is it?}
\label{sec:cost}
Table~\ref{tab:cost} compares the cost of obtaining predictions. Only DLKT and CLST need
gradient training; \JevZ{} and JevKT predict for a new course immediately. One JevKT prediction
takes 325\,ms in a single call and costs \$886 per million predictions, about
1/5 of Thinking-KT, which needs \$4{,}360 and 41.7\,s per prediction even with
its ten votes issued in parallel; \JevZ{} costs \$44. Priced at the same GPU rental rate,
DLKT is far cheaper, since training on all learners costs at most \$0.08 and a million
predictions well under \$1. Its limit is not cost but data, as it must be retrained for every
course and needs logged learners to train on.
\begin{table}[t]
\centering
\scriptsize
\setlength{\tabcolsep}{2.5pt}
\caption{Efficiency. \emph{Training}: gradient training on target data is needed.
\emph{Ready after} and \emph{Train \$}: time and cost until the first prediction for a new course
with $N{=}8$ / all learners. Latency (median, one prediction), throughput, and cost per million
predictions. API methods: OpenRouter prices. DLKT, CLST and JEV-9B run on one RTX 3090, priced at
the RunPod Community Cloud on-demand rate (\$0.22/h, September 2026), which favours them.
LOKT and Thinking-KT (budget 2048) use 10 votes; CLST fine-tunes Mistral-7B with LoRA.
$^\ddagger$Distilled once from Jev, not per course.}
\label{tab:cost}
\resizebox{\columnwidth}{!}{%
\begin{tabular}{l c c c r r r}
\toprule
\textbf{Method} & \textbf{Training} & \textbf{Ready after} & \textbf{Train \$} & \textbf{Latency} & \textbf{Pred/s} & \textbf{\$/1M} \\
\midrule
\rowcolor{focusbg}
JevKT & \xmark & 0 / 0 & -- & 325 ms & 72.23 & \$886 \\
\JevZ{} & \xmark & 0 / 0 & -- & 200 ms & 105 & \$44 \\
LOKT & \xmark & 0 / 0 & -- & 1.3 s & 3.39 & \$707 \\
Thinking-KT & \xmark & 0 / 0 & -- & 41.7 s & 0.12 & \$4,360 \\
JEV-9B (vLLM) & \xmark & 0 / 0 & --$^\ddagger$ & 204 ms & 8.27 & \$7.39 \\
\midrule
CLST & \cmark & 110.8 s / -- & $<$\$0.01 / -- & 104 ms & 9.6 & \$6.37 \\
DKT & \cmark & 1.1 s / 37.6 s & $<$\$0.01 / $<$\$0.01 & 0.23 ms & 4,333 & \$0.01 \\
AKT & \cmark & 6.3 s / 22.4 min & $<$\$0.01 / \$0.08 & 3.08 ms & 325 & \$0.19 \\
simpleKT & \cmark & 2.6 s / 3.2 min & $<$\$0.01 / \$0.01 & 0.46 ms & 2,167 & \$0.03 \\
csKT & \cmark & 1.7 s / 2.8 min & $<$\$0.01 / \$0.01 & 0.54 ms & 1,857 & \$0.03 \\
\bottomrule
\end{tabular}}
\end{table}

\subsection{Could the result be an artefact?}
\label{sec:validity}
Three checks address whether the result could be an artefact. \emph{Contamination}
(Table~\ref{tab:contamination}, Appendix~\ref{app:contamination}). Asked about an item's
difficulty or skill from its ID alone, Jev and three open LLMs do no better with real IDs than
with format-matched fake IDs on any dataset, and continuing rows of the public files shows no
recall. Relabelling IDs and removing names changes \JevZ{} AUC by at most .012, never
significantly (Holm), and no dataset is flagged. On six synthetic datasets generated after
the model release, JevKT is still at or above the best deep KT model at $N{=}8$. \emph{Leakage}. A leak audit passes in every
setting, including a test that the input does not change when the target outcome is flipped.
\emph{Baseline strength}. The best deep KT model is selected on test in every cell, which
favours deep KT.
Classical KT models (BKT, PFA) fitted on the same learners are below
both (Appendix~\ref{app:coldstart}).
With hyper-parameters tuned on the validation learners of each cell (four strong pyKT models,
eight configurations each), deep KT stays below JevKT by \tunedDeltaEight, \tunedDeltaSixteen,
\tunedDeltaThirtyTwo{} and \tunedDeltaSixtyFour{} at $N{=}8$, 16, 32 and 64 (Holm-significant on
\tunedSigEight, \tunedSigSixteen, \tunedSigThirtyTwo{} and \tunedSigSixtyFour{} datasets), so default
hyper-parameters do not explain the gap.

\section{Conclusion}
\label{sec:conclusion}
Yes, an off-the-shelf System-One LLM already performs KT when few or no learners are logged.
Without any data from the target platform, Jev beats deep KT trained on 8 learners on average
and prompted System-Two LLM-KT on every dataset, at a fraction of its cost; with evidence from the logged
learners, JevKT stays ahead until supervised KT catches up between \nLead{} and \nCross{}
learners. Among the readers we tested, the gain comes from the model rather than the typed
format; the input recipe helps other LLMs as much as it helps Jev. Whether an open System-One
model can match it remains open.

\section*{Limitations}
Our finding concerns one closed System-One model. No other commercial System-One model is
available, so we test generality only with other LLMs behind the official adapter, a
Jev-distilled student and the open-source Laya (Appendix~\ref{app:generality}); the adapter
reads probabilities that the LLM writes as text, which may disadvantage it. Jev's training data
are unknown; our contamination checks find no evidence of memorisation but cannot rule it out.
Each prediction requires an API call, and results may change with future model versions; we
pin the version and cache every response. Re-querying the pinned version after two days reproduces
its predictions closely (Appendix~\ref{app:stability}), but only one Jev version is available, so
we cannot measure drift across versions. The prompted System-Two baselines run on Qwen3.5-9B, so
our conclusions about reasoning apply to this backbone; a larger reasoning model may behave
differently. Deep KT baselines use default hyper-parameters in the main comparison; a
validation-tuned run (Section~\ref{sec:validity}) does not change the conclusion, but a broader
search could narrow the gap. F1 is threshold-sensitive
on datasets with a low base rate (ASSIST17).

\section*{Use of Generative AI}
We used Claude Code (Anthropic) to assist with writing code and with language editing of the
manuscript. The human authors conceived the research ideas, verified all results and their
interpretation, and take full responsibility for the content.

\section*{Ethics Statement}
We use only public, de-identified KT datasets for research and collect no new data from people.
For inference, serialised learner histories are sent to LLMs through the OpenRouter API. KT predictions can influence what learners are shown; a model that is accurate on average
can still be wrong for individual learners or groups, so its outputs should support, not replace,
instructors' decisions.

\bibliography{references}

\appendix
\makeatletter
\setlength{\@fptop}{0pt}\setlength{\@dblfptop}{0pt}\setlength{\@fpsep}{14pt}\setlength{\@dblfpsep}{14pt}
\makeatother %
\setcounter{dbltopnumber}{3}\setcounter{topnumber}{3}
\renewcommand{\dbltopfraction}{.95}\renewcommand{\topfraction}{.95}\renewcommand{\textfraction}{.05}
\renewcommand{\dblfloatpagefraction}{.8}\renewcommand{\floatpagefraction}{.8}

\section{Datasets and Splits}
\label{app:data}
Table~\ref{tab:datasets} summarises the seven datasets: ASSIST2009~\citep{feng_2009_assistments},
NIPS34~\citep{wang_2020_eedi}, XES3G5M~\citep{liu_2023_xes3g5m},
DBE-KT22~\citep{abdelrahman_2022_dbekt22}, Algebra2005~\citep{stamper_2016_kddcup},
Junyi (2018--19 Kaggle release) and ASSIST2017~\citep{patikorn_2020_assist2017}. They differ in size, item and KC granularity, available text and base
rate, which matters for cold start: datasets with many interactions per item (e.g.,
Algebra2005) give counting evidence and supervised models more to work with.
\paragraph{Preprocessing.} We keep learners with at least 10 interactions and subsample
5{,}000 learners for XES3G5M and Junyi. An item with several KCs stays one interaction (no row
duplication, which would leak the outcome to the next row). Item text and KC names are kept
where available and used in the serialised state.
\paragraph{Splits and targets.} Learners are split 80/20 at the learner level (seed 0); 10\%
of the training learners form a dev split, used for early stopping of DLKT and for selecting
the JevKT input (Appendix~\ref{app:explore}). For P1 and P2, nested subsets of $N$ training
learners are drawn per seed (5 seeds) and the first 50 interactions of held-out learners are
scored (2{,}000 targets per dataset, identical for all methods and $N$). For P3, 20\% of the
item IDs are held out per seed (3 seeds); every interaction on them is removed from training,
examples and statistics, and up to 1{,}000 targets on these items are scored.
\begin{table}[!h]
\centering
\scriptsize
\setlength{\tabcolsep}{3pt}
\caption{Datasets. Learners with at least 10 interactions; base rate = share of correct
answers among test targets.}
\label{tab:datasets}
\begin{tabular}{l rrrrc}
\toprule
\textbf{Dataset} & \textbf{Learners} & \textbf{Interactions} & \textbf{Items} & \textbf{KCs} & \textbf{Base rate} \\
\midrule
ASSIST2009  & 3{,}002 & 277k  & 17{,}705 & 123   & .63 \\
NIPS34      & 4{,}918 & 1.38M & 948      & 57    & .54 \\
XES3G5M     & 5{,}000$^\ast$ & 1.54M & 7{,}374 & 847 & .79 \\
DBE-KT22    & 1{,}138 & 158k  & 212      & 93    & .78 \\
Algebra2005 & 567     & 607k  & 173k     & 112   & .74 \\
Junyi       & 5{,}000$^\ast$ & 1.33M & 24{,}683 & 1{,}326 & .71 \\
ASSIST2017  & 1{,}708 & 943k  & 3{,}162  & 102   & .40 \\
\bottomrule
\end{tabular}

\vspace{2pt}{\scriptsize $^\ast$Random subset of 18{,}066 (XES3G5M) and 61k (Junyi) learners.}
\end{table}

\section{Implementation Details}
\label{app:impl}
\paragraph{JevKT input.} Each prompt is a JSON state with four fields: the learner's last
$H{=}25$ interactions (item, KC names, correct/wrong), the target item, one sentence with the
similar-learner statistic, and $k{=}64$ solved examples. Each example shows another training
learner's last 10 interactions, an item sharing a KC with the target and the outcome on that
item; examples are sampled with a fixed seed per target. Jev answers one typed yes/no
question (``Will the learner answer \texttt{next\_item} correctly?''), and we use
$\pi(\text{yes})$ without post-hoc calibration. A shortened example is:
\begin{quote}\footnotesize\ttfamily
learner\_history: [\dots 25 interactions \dots]\\
next\_item: \{item, KCs: [\dots]\}\\
similar\_learners: ``the 20 learners with the most similar skill profile answered this item
correctly 72\% of the time''\\
solved\_examples\_from\_other\_learners: Example learner 1: [\dots] Outcome on that item:
CORRECT \dots
\end{quote}
\paragraph{Jev.} We query \texttt{jev-1.13-20260917} through OpenRouter's System-One
endpoint. If an input exceeds the context limit, the number of shots is halved until it fits
(3.4\% of all Jev targets, almost all on Algebra05, where 22\% of the cold-start and 40\% of the full-data targets
use 32 instead of 64 shots; no target needed fewer than 32).
\paragraph{Deep KT.} Eleven pyKT models~\citep{liu_2022_pykt}, chosen as the best-ranked models
in pyKT's reported results plus its cold-start model: DKT~\citep{piech_2015_dkt},
DKT+~\citep{yeung_2018_dktplus}, AKT~\citep{ghosh_2020_akt}, ATKT~\citep{guo_2021_atkt},
simpleKT~\citep{liu_2023_simplekt}, sparseKT~\citep{huang_2023_sparsekt},
FoLiBiKT~\citep{im_2023_folibikt}, stableKT~\citep{li_2024_stablekt},
extraKT~\citep{li_2024_extrakt}, robustKT~\citep{guo_2025_robustkt} and
csKT~\citep{bai_2025_cskt}. All use default hyper-parameters
(sequence length 200), trained on the available $N$ learners and early-stopped on their
held-out part; the best model per (dataset, $N$) is selected on test, which favours DLKT.
The pool adds \nDLKTadded{} further pyKT models (\DLKTaddedList), \nDLKT{} in total; in each
cell the best is taken over all models with at least three finished seeds.

\paragraph{LLM-KT.} LOKT~\citep{kim_2024_lokt} and Thinking-KT~\citep{lee_2026_thinkingkt}
use the released prompts with Qwen3.5-9B through OpenRouter and ten votes sampled at temperature 0.7; Thinking-KT
uses a reasoning budget of 1024 or 2048 tokens. CLST~\citep{jung_2025_clst_jedm} is re-implemented
with Mistral-7B-Instruct and LoRA (one epoch) and evaluated on the same cold-start targets
on the three datasets and the range of $N$ of the original paper ($N\le64$, plus $N{=}0$).
\paragraph{Compute.} DLKT, CLST and the open-weight analyses run on single RTX 3090 GPUs; API
methods use OpenRouter. The logged API cost of all experiments was about \$250.

\begin{table*}[t]
\centering
\scriptsize
\caption{Input ablation under cold start ($N{=}16$ training learners, mean over 5 seeds, same
early-window targets as P1). Each row adds one input to the learner's history; \JevZ{} uses no
training learners. Best in \textbf{bold}, second best \underline{underlined}.}
\label{tab:ablation-n16}
\begin{tabular}{l ccccccc c}
\toprule
\textbf{Input to Jev} & A09 & NIPS & XES & DBE & Alg05 & Junyi & A17 & \textbf{Mean} \\
\midrule
\JevZ{} (no evidence, no shots) & .738 & .691 & .725 & .742 & .698 & .700 & .648 & .706 \\
+ item/KC statistics & \textbf{.741} & .708 & \underline{.729} & .786 & .723 & \textbf{.704} & \textbf{.660} & .721 \\
+ similar-learner kNN & .738 & \underline{.733} & \textbf{.730} & \underline{.816} & \underline{.733} & .702 & \underline{.649} & \underline{.729} \\
+ 16 KC-matched shots & \underline{.739} & .709 & \underline{.729} & .797 & .723 & .702 & .641 & .720 \\
+ 64 KC-matched shots & \underline{.739} & .724 & .720 & .813 & .727 & \underline{.703} & .643 & .724 \\
\rowcolor{focusbg} + 64 shots + kNN (JevKT) & .737 & \textbf{.734} & .720 & \textbf{.834} & \textbf{.738} & \textbf{.704} & .646 & \textbf{.730} \\
\bottomrule
\end{tabular}
\end{table*}

\section{Input-Design Exploration}
\label{app:explore}
\label{app:ablation}
Before fixing JevKT we explored changes to the input only (Table~\ref{tab:explore}): history
length ($H\in\{5,\dots,100\}$), random vs.\ KC-matched shots ($k\le128$), population
statistics vs.\ the similar-learner statistic, repeated calls, state formats, paraphrased
questions and LOKT option weights. The selection rule (best mean $\Delta$AUC on the development split) was
fixed before test was evaluated, and test was evaluated once. Three findings guided the
design. First, the similar-learner statistic is the largest single gain, larger than item and
KC population statistics. Second, shots help only when they are KC-matched; random shots are
neutral or slightly harmful at every $k$. Third, repeats, paraphrases, formats, longer
histories and option weights do not change AUC, so JevKT keeps the simplest state. The development-split
ranking transfers to test (Figure~\ref{fig:shots}).
\begin{table}[t]
\centering
\scriptsize
\setlength{\tabcolsep}{3pt}
\caption{Input-design exploration for Jev. Mean $\Delta$AUC over 7 datasets relative
to the state only (AUC dev .703 / test .710). The configuration was selected on
the development split by a rule fixed in advance; test values are shown for reference and were not used for selection. 18 runs with LOKT
option weights (TCOW) did not change AUC and are omitted. Best in \textbf{bold}, second best \underline{underlined}.}
\label{tab:explore}
\begin{tabular}{l cc}
\toprule
\textbf{Input added to the state} & \textbf{dev} & \textbf{test} \\
\midrule
\rowcolor{focusbg}
\textbf{+ kNN, + 64 KC-matched shots (JevKT)} & \textbf{+.062} & \underline{+.066} \\
+ kNN, + 32 KC-matched shots & \underline{+.060} & \textbf{+.067} \\
+ statistics + kNN, + 16 KC-matched shots & +.055 & +.062 \\
+ kNN & +.050 & +.051 \\
$H{=}50$, + statistics + kNN & +.046 & +.049 \\
+ statistics + kNN & +.046 & +.048 \\
+ 128 KC-matched shots & +.029 & -- \\
+ 64 KC-matched shots & +.029 & +.032 \\
+ item/KC statistics & +.029 & +.029 \\
+ 32 KC-matched shots & +.025 & +.029 \\
+ 16 KC-matched shots (seed 1) & +.021 & +.018 \\
$H{=}100$, + 16 KC-matched shots & +.018 & +.020 \\
+ 16 KC-matched shots & +.016 & +.021 \\
+ 16 KC-matched shots (seed 2) & +.014 & +.015 \\
+ 4 KC-matched shots & +.006 & +.006 \\
$H{=}100$ & +.002 & +.002 \\
$H{=}50$ & +.002 & +.003 \\
5 repeated calls & +.000 & +.000 \\
3 repeated calls & +.000 & +.000 \\
paraphrased question set & +.000 & +.000 \\
10 repeated calls & +.000 & +.001 \\
format: recent-trend line & $-$.000 & $-$.003 \\
format: IDs only & $-$.001 & $-$.004 \\
+ 16 random shots & $-$.004 & $-$.002 \\
format: KC summary & $-$.004 & $-$.011 \\
+ 4 random shots & $-$.005 & $-$.001 \\
+ 1 random shots & $-$.005 & $-$.002 \\
+ 8 random shots & $-$.005 & $-$.001 \\
+ 2 random shots & $-$.005 & $-$.003 \\
$H{=}10$ & $-$.006 & $-$.009 \\
$H{=}5$ & $-$.008 & $-$.014 \\
\bottomrule
\end{tabular}
\end{table}

The exploration used the full training data. To check the selection under cold start, we
repeat the main input variants with $N{=}16$ training learners
(Table~\ref{tab:ablation-n16}). Every form of evidence improves on \JevZ{} (\abZero), and the
similar-learner statistic alone (\abKnn) is almost as good as the full JevKT input (\abFinal).
KC-matched shots help less than the statistic (\abKCsixteen{} with 16 shots, \abKCsixtyfour{}
with 64), and combining the two is sub-additive: the gain over the statistic alone is clear
only on DBE-KT22. The ranking of the inputs is the same as with full data.

\FloatBarrier
\section{Cold-Start Details}
\label{app:coldstart}
\label{app:fulldata}
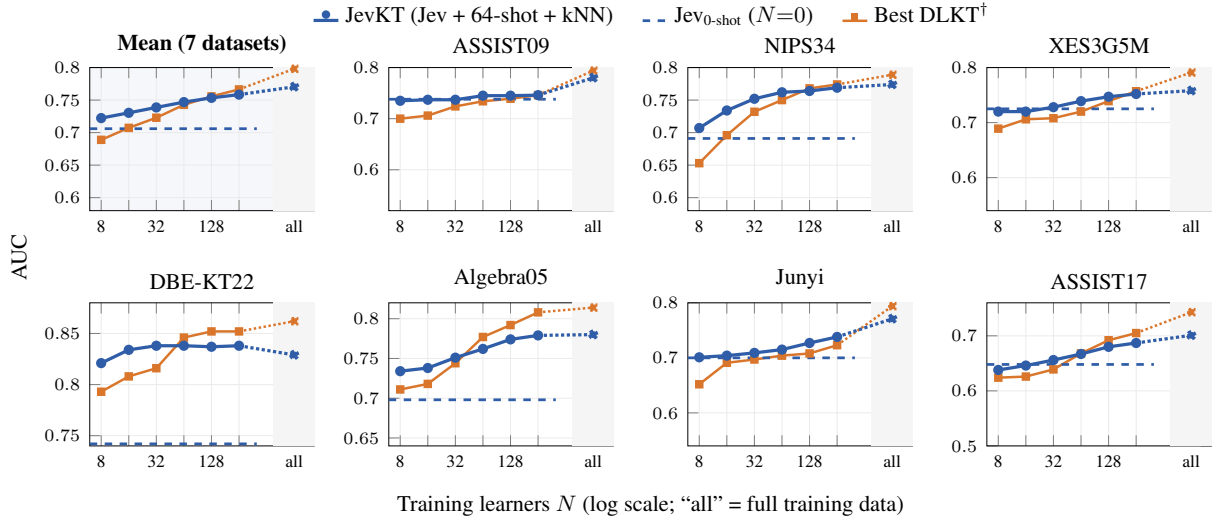
\begin{figure*}[t]
\centering
\resizebox{\textwidth}{!}{%
\begin{tikzpicture}
\begin{groupplot}[
  group style={group size=4 by 2, horizontal sep=1.05cm, vertical sep=1.3cm},
  width=4.75cm, height=3.6cm,
  xmode=log, log basis x=2, xmin=6, xmax=1700,
  xtick={8,16,32,64,128,256,1024}, xticklabels={8,,32,,128,,all},
  yticklabel style={/pgf/number format/fixed, /pgf/number format/precision=2},
  tick label style={font=\scriptsize}, title style={font=\footnotesize, yshift=-1.5mm},
  grid=major, grid style={black!8},
]
\nextgroupplot[title={\textbf{Mean (7 datasets)}}, ymin=0.58, ymax=0.80, axis background/.style={fill=cJev!5}]
\fill[black!4] (axis cs:600,0.58) rectangle (axis cs:1700,0.80);
\addplot[cDLKT, mark=square*, mark size=1.2pt, line width=1pt] coordinates {(8,0.6889) (16,0.7073) (32,0.7229) (64,0.7427) (128,0.7557) (256,0.7664)};
\addplot[cJev, mark=*, mark size=1.4pt, line width=1.3pt] coordinates {(8,0.7223) (16,0.7304) (32,0.7387) (64,0.7469) (128,0.7534) (256,0.7584)};
\addplot[cJev, dashed, line width=1.1pt, domain=6:400, samples=2] {0.706};
\addplot[cDLKT, mark=square*, mark size=1.2pt, line width=1pt, densely dotted, mark indices={2}] coordinates {(256,0.7664) (1024,0.7981)};
\addplot[cJev, mark=*, mark size=1.4pt, line width=1.3pt, densely dotted, mark indices={2}] coordinates {(256,0.7584) (1024,0.7704)};

\nextgroupplot[title={ASSIST09}, ymin=0.52, ymax=0.80]
\fill[black!4] (axis cs:600,0.52) rectangle (axis cs:1700,0.80);
\addplot[cDLKT, mark=square*, mark size=1.2pt, line width=1pt] coordinates {(8,0.7000) (16,0.7060) (32,0.7240) (64,0.7340) (128,0.7390) (256,0.7460)};
\addplot[cJev, mark=*, mark size=1.4pt, line width=1.3pt] coordinates {(8,0.7350) (16,0.7370) (32,0.7370) (64,0.7450) (128,0.7450) (256,0.7460)};
\addplot[cJev, dashed, line width=1.1pt, domain=6:400, samples=2] {0.738};
\addplot[cDLKT, mark=square*, mark size=1.2pt, line width=1pt, densely dotted, mark indices={2}] coordinates {(256,0.7460) (1024,0.7940)};
\addplot[cJev, mark=*, mark size=1.4pt, line width=1.3pt, densely dotted, mark indices={2}] coordinates {(256,0.7460) (1024,0.7800)};

\nextgroupplot[title={NIPS34}, ymin=0.58, ymax=0.80]
\fill[black!4] (axis cs:600,0.58) rectangle (axis cs:1700,0.80);
\addplot[cDLKT, mark=square*, mark size=1.2pt, line width=1pt] coordinates {(8,0.6530) (16,0.6960) (32,0.7320) (64,0.7500) (128,0.7680) (256,0.7740)};
\addplot[cJev, mark=*, mark size=1.4pt, line width=1.3pt] coordinates {(8,0.7070) (16,0.7340) (32,0.7520) (64,0.7620) (128,0.7640) (256,0.7690)};
\addplot[cJev, dashed, line width=1.1pt, domain=6:400, samples=2] {0.691};
\addplot[cDLKT, mark=square*, mark size=1.2pt, line width=1pt, densely dotted, mark indices={2}] coordinates {(256,0.7740) (1024,0.7890)};
\addplot[cJev, mark=*, mark size=1.4pt, line width=1.3pt, densely dotted, mark indices={2}] coordinates {(256,0.7690) (1024,0.7740)};

\nextgroupplot[title={XES3G5M}, ymin=0.54, ymax=0.80]
\fill[black!4] (axis cs:600,0.54) rectangle (axis cs:1700,0.80);
\addplot[cDLKT, mark=square*, mark size=1.2pt, line width=1pt] coordinates {(8,0.6890) (16,0.7060) (32,0.7080) (64,0.7200) (128,0.7390) (256,0.7570)};
\addplot[cJev, mark=*, mark size=1.4pt, line width=1.3pt] coordinates {(8,0.7200) (16,0.7200) (32,0.7280) (64,0.7390) (128,0.7470) (256,0.7520)};
\addplot[cJev, dashed, line width=1.1pt, domain=6:400, samples=2] {0.725};
\addplot[cDLKT, mark=square*, mark size=1.2pt, line width=1pt, densely dotted, mark indices={2}] coordinates {(256,0.7570) (1024,0.7910)};
\addplot[cJev, mark=*, mark size=1.4pt, line width=1.3pt, densely dotted, mark indices={2}] coordinates {(256,0.7520) (1024,0.7580)};

\nextgroupplot[title={DBE-KT22}, ymin=0.74, ymax=0.88]
\fill[black!4] (axis cs:600,0.74) rectangle (axis cs:1700,0.88);
\addplot[cDLKT, mark=square*, mark size=1.2pt, line width=1pt] coordinates {(8,0.7930) (16,0.8080) (32,0.8160) (64,0.8460) (128,0.8520) (256,0.8520)};
\addplot[cJev, mark=*, mark size=1.4pt, line width=1.3pt] coordinates {(8,0.8210) (16,0.8340) (32,0.8380) (64,0.8380) (128,0.8370) (256,0.8380)};
\addplot[cJev, dashed, line width=1.1pt, domain=6:400, samples=2] {0.742};
\addplot[cDLKT, mark=square*, mark size=1.2pt, line width=1pt, densely dotted, mark indices={2}] coordinates {(256,0.8520) (1024,0.8620)};
\addplot[cJev, mark=*, mark size=1.4pt, line width=1.3pt, densely dotted, mark indices={2}] coordinates {(256,0.8380) (1024,0.8290)};

\nextgroupplot[title={Algebra05}, ymin=0.64, ymax=0.82]
\fill[black!4] (axis cs:600,0.64) rectangle (axis cs:1700,0.82);
\addplot[cDLKT, mark=square*, mark size=1.2pt, line width=1pt] coordinates {(8,0.7110) (16,0.7180) (32,0.7440) (64,0.7770) (128,0.7920) (256,0.8080)};
\addplot[cJev, mark=*, mark size=1.4pt, line width=1.3pt] coordinates {(8,0.7340) (16,0.7380) (32,0.7510) (64,0.7620) (128,0.7740) (256,0.7790)};
\addplot[cJev, dashed, line width=1.1pt, domain=6:400, samples=2] {0.698};
\addplot[cDLKT, mark=square*, mark size=1.2pt, line width=1pt, densely dotted, mark indices={2}] coordinates {(256,0.8080) (1024,0.8140)};
\addplot[cJev, mark=*, mark size=1.4pt, line width=1.3pt, densely dotted, mark indices={2}] coordinates {(256,0.7790) (1024,0.7800)};

\nextgroupplot[title={Junyi}, ymin=0.54, ymax=0.80]
\fill[black!4] (axis cs:600,0.54) rectangle (axis cs:1700,0.80);
\addplot[cDLKT, mark=square*, mark size=1.2pt, line width=1pt] coordinates {(8,0.6520) (16,0.6910) (32,0.6970) (64,0.7040) (128,0.7080) (256,0.7230)};
\addplot[cJev, mark=*, mark size=1.4pt, line width=1.3pt] coordinates {(8,0.7010) (16,0.7040) (32,0.7090) (64,0.7150) (128,0.7270) (256,0.7380)};
\addplot[cJev, dashed, line width=1.1pt, domain=6:400, samples=2] {0.700};
\addplot[cDLKT, mark=square*, mark size=1.2pt, line width=1pt, densely dotted, mark indices={2}] coordinates {(256,0.7230) (1024,0.7940)};
\addplot[cJev, mark=*, mark size=1.4pt, line width=1.3pt, densely dotted, mark indices={2}] coordinates {(256,0.7380) (1024,0.7710)};

\nextgroupplot[title={ASSIST17}, ymin=0.50, ymax=0.76]
\fill[black!4] (axis cs:600,0.50) rectangle (axis cs:1700,0.76);
\addplot[cDLKT, mark=square*, mark size=1.2pt, line width=1pt] coordinates {(8,0.6240) (16,0.6260) (32,0.6390) (64,0.6680) (128,0.6920) (256,0.7050)};
\addplot[cJev, mark=*, mark size=1.4pt, line width=1.3pt] coordinates {(8,0.6380) (16,0.6460) (32,0.6560) (64,0.6670) (128,0.6800) (256,0.6870)};
\addplot[cJev, dashed, line width=1.1pt, domain=6:400, samples=2] {0.648};
\addplot[cDLKT, mark=square*, mark size=1.2pt, line width=1pt, densely dotted, mark indices={2}] coordinates {(256,0.7050) (1024,0.7430)};
\addplot[cJev, mark=*, mark size=1.4pt, line width=1.3pt, densely dotted, mark indices={2}] coordinates {(256,0.6870) (1024,0.7010)};

\end{groupplot}
\node[font=\small] at ($(group c2r2.south)!0.5!(group c3r2.south) + (0,-0.85)$) {Training learners $N$ (log scale; ``all'' = full training data)};
\node[font=\small, rotate=90] at ($(group c1r1.west)!0.5!(group c1r2.west) + (-1.0,0)$) {AUC};
\node[anchor=south, font=\footnotesize] at ($(group c1r1.north west)!0.5!(group c4r1.north east) + (0,0.45)$) {%
  \tikz\draw[cJev, line width=1.3pt] (0,0)--(0.35,0) node[midway,fill=cJev,circle,inner sep=1.2pt]{}; JevKT (Jev + 64-shot + kNN) \quad
  \tikz\draw[cJev, dashed, line width=1.1pt] (0,0)--(0.35,0); \JevZ{} ($N{=}0$) \quad
  \tikz\draw[cDLKT, line width=1pt] (0,0)--(0.35,0) node[midway,fill=cDLKT,inner sep=1.4pt]{}; Best DLKT$^\dagger$};
\end{tikzpicture}}
\caption{P1 per dataset (first panel: mean). AUC as a function of the number of training learners $N$ (identical early-window targets,
mean over 5 seeds); the shaded ``all'' point uses the full training data on the same
targets. With few learners, training-free JevKT outperforms the
best of \nDLKT{} DLKT models and stays at or above \JevZ{} (dashed); supervised KT catches up
between $N{=}\nLead$ and $\nCross$ on average and leads with full data. $^\dagger$Best of \nDLKT{} pyKT models selected on test per cell, which favours DLKT.}
\label{fig:curves}
\end{figure*}
This section gives the per-dataset numbers behind Table~\ref{tab:main-cs} and Figure~\ref{fig:main}.

\begin{figure}[t]
\centering
\begin{tikzpicture}
\begin{axis}[
  width=\columnwidth, height=5.2cm, scaled y ticks=false,
  xmode=log, log basis x=2,
  xtick={8,16,32,64,128,256,1024}, xticklabels={8,16,32,64,128,256,all},
  xmin=6, xmax=1400, ymin=-0.06, ymax=0.06, ytick distance=0.02,
  yticklabel style={/pgf/number format/fixed, /pgf/number format/precision=2},
  xlabel={Training learners $N$}, ylabel={$\Delta$AUC (JevKT $-$ best DLKT)},
  label style={font=\small}, tick label style={font=\footnotesize},
  grid=major, grid style={black!8},
  legend style={font=\scriptsize, draw=none, at={(0.5,-0.32)}, anchor=north,
                legend columns=4, column sep=3pt},
  cycle list name=dsets,
]
\fill[cJev!7] (axis cs:6,0) rectangle (axis cs:1400,0.06);
\fill[cDLKT!7] (axis cs:6,-0.06) rectangle (axis cs:1400,0);
\addplot coordinates {(8,0.0350) (16,0.0310) (32,0.0130) (64,0.0110) (128,0.0060) (256,0.0000) (1024,-0.0140)}; \addlegendentry{A09}
\addplot coordinates {(8,0.0540) (16,0.0380) (32,0.0200) (64,0.0120) (128,-0.0040) (256,-0.0050) (1024,-0.0150)}; \addlegendentry{NIPS}
\addplot coordinates {(8,0.0310) (16,0.0140) (32,0.0200) (64,0.0190) (128,0.0080) (256,-0.0050) (1024,-0.0330)}; \addlegendentry{XES}
\addplot coordinates {(8,0.0280) (16,0.0260) (32,0.0220) (64,-0.0080) (128,-0.0150) (256,-0.0140) (1024,-0.0330)}; \addlegendentry{DBE}
\addplot coordinates {(8,0.0230) (16,0.0200) (32,0.0070) (64,-0.0150) (128,-0.0180) (256,-0.0290) (1024,-0.0340)}; \addlegendentry{Alg05}
\addplot coordinates {(8,0.0490) (16,0.0130) (32,0.0120) (64,0.0110) (128,0.0190) (256,0.0150) (1024,-0.0230)}; \addlegendentry{Junyi}
\addplot coordinates {(8,0.0140) (16,0.0200) (32,0.0170) (64,-0.0010) (128,-0.0120) (256,-0.0180) (1024,-0.0420)}; \addlegendentry{A17}
\addplot[black, line width=1.6pt, mark=*, mark size=1.8pt] coordinates {(8,0.0334) (16,0.0231) (32,0.0159) (64,0.0041) (128,-0.0023) (256,-0.0080) (1024,-0.0277)}; \addlegendentry{\textbf{Mean}}
\draw[black!60, line width=0.6pt] (axis cs:6,0) -- (axis cs:1400,0);
\node[font=\scriptsize\itshape, text=cJev!80!black, anchor=north east] at (axis cs:1300,0.058) {JevKT better};
\node[font=\scriptsize\itshape, text=cDLKT!80!black, anchor=south east] at (axis cs:1300,-0.058) {DLKT better};
\end{axis}
\end{tikzpicture}
\caption{The cold-start advantage of training-free JevKT (Jev + 64-shot + kNN) shrinks as more learners are
logged. $\Delta$AUC against the best of \nDLKT{} pyKT models (selected on test) on identical
early-window targets; thin lines are datasets, the thick line is the mean over datasets
with all 7 cells available. }
\label{fig:crossover}
\end{figure}
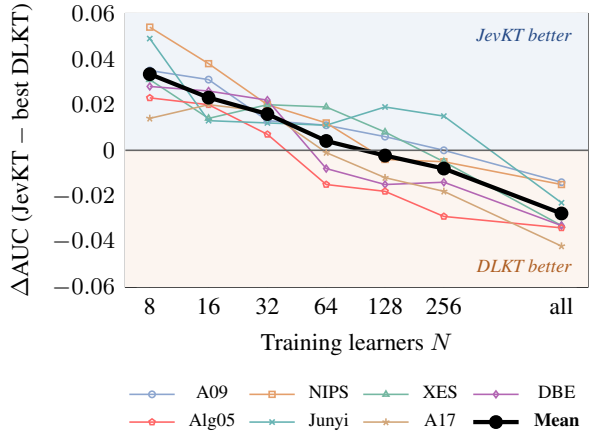

\paragraph{P1: few training learners.}
Figure~\ref{fig:curves} shows P1 per dataset and Figure~\ref{fig:crossover} the per-dataset
difference to the best deep KT model.
JevKT leads on average up to $N{=}\nLead$ (on \csWinsEight{} datasets at $N{=}8$ and \csWinsSixtyFour{} at $N{=}64$); the lead shrinks with $N$ and disappears
between $N{=}\nLead$ and $\nCross$ on average, at different $N$ for different datasets. At $N{=}0$, \JevZ{} is the strongest training-free method on every dataset.
Appendix~\ref{app:fulltables} lists, for each $N$ separately, the AUC, ACC and F1 of every model we ran
with its spread over seeds (Tables~\ref{tab:p1n0}--\ref{tab:p1fnall}): System-One LLMs, System-Two LLM-KT
(prompted: LOKT and Thinking-KT; trained: CLST and the reported LLM-KT value), the classical KT models BKT
and PFA fitted on the same $N$ learners, and all \nDLKT{} pyKT models (models with fewer than three
finished seeds are marked and are not eligible as the best deep KT model). BKT and PFA stay well
below JevKT and the best deep KT model at every $N$. ACC follows the AUC results; F1 differs
little at small $N$ and is sensitive to the .5 threshold on datasets with a low base rate (ASSIST17).

\paragraph{Trained System-Two LLM-KT (CLST).}
We re-implement CLST, which fine-tunes Mistral-7B with LoRA on the logged learners, and run it on
the three datasets of the original paper (Table~\ref{tab:clst}). At $N{=}8$ our numbers are
within about .03 AUC of those reported on the paper's own splits. CLST is below JevKT in
\clstWins{} cells with $N{\ge}8$ and below the best of the eleven core deep KT models in \clstBelowDlkt{}, so fine-tuning
an LLM on a few learners does not reach what the off-the-shelf System-One model achieves
without training.
\begin{table}[t]
\centering
\scriptsize
\setlength{\tabcolsep}{3pt}
\caption{CLST (our re-implementation, fine-tuned Mistral-7B with LoRA) on the three datasets of the
original paper, on our early-window targets (mean over 5 seeds; $N{=}0$: one run). Jev = JevKT
($N{\ge}8$) or \JevZ{} ($N{=}0$). \emph{Paper}: CLST's reported AUC at $N{=}8$ on its own splits. $^\dagger$Best of the 11 core pyKT models. Best in \textbf{bold}, second best \underline{underlined}.}
\label{tab:clst}
\begin{tabular}{lc cccc}
\toprule
\textbf{Dataset} & $N$ & \textbf{CLST} & \textbf{Jev} & \textbf{Best DLKT}$^\dagger$ & \textbf{Paper} \\
\midrule
ASSIST09 & 0 & \underline{.728} & \textbf{.738} & -- & -- \\
 & 8 & .658 & \textbf{.735} & \underline{.700} & .689 \\
 & 16 & .682 & \textbf{.737} & \underline{.706} & -- \\
 & 32 & .686 & \textbf{.737} & \underline{.724} & -- \\
 & 64 & \underline{.735} & \textbf{.745} & .734 & -- \\
\midrule
NIPS34 & 0 & \underline{.623} & \textbf{.691} & -- & -- \\
 & 8 & .605 & \textbf{.707} & \underline{.653} & .635 \\
 & 16 & .658 & \textbf{.734} & \underline{.672} & -- \\
 & 32 & .649 & \textbf{.752} & \underline{.688} & -- \\
 & 64 & .673 & \textbf{.762} & \underline{.724} & -- \\
\midrule
Algebra05 & 0 & \underline{.674} & \textbf{.698} & -- & -- \\
 & 8 & .643 & \textbf{.734} & \underline{.707} & .655 \\
 & 16 & .678 & \textbf{.738} & \underline{.718} & -- \\
 & 32 & .658 & \textbf{.751} & \underline{.744} & -- \\
 & 64 & .717 & \underline{.762} & \textbf{.777} & -- \\
\bottomrule
\end{tabular}
\end{table}

\paragraph{Can the crossover be predicted?}
We tried to predict the crossover $N^\star$ from dataset properties with six text-free
synthetic datasets from a BKT+IRT simulator (1{,}600 learners each; 48 to 3{,}072 items,
two sequence lengths and two numbers of KCs; $N\in\{8,32,128,512,\text{all}\}$, 2 seeds),
comparing JevKT with the best of four pyKT models. On these data $N^\star$ rises with the
size of the item bank relative to sequence length ($\log N^\star = 3.12 +
0.49\log(Q/\bar L)$), but the fit does not predict the real datasets (Spearman
+.11, $p{=}.82$; mean error 2.3 doublings of $N$), and no single dataset
property correlates with the observed $N^\star$. The synthetic gaps are small (at most about
.03 AUC), so their crossovers are uncertain. We therefore report the crossover as
observed and do not claim to predict it.

\begin{table*}[t]
\centering
\scriptsize
\setlength{\tabcolsep}{3pt}
\caption{P2 (new learners), every cell. AUC by the position of the target in a new learner's
sequence with $N{=}8$ training learners (\JevZ{} uses none), mean{\tiny$\pm$SD} over 5 seeds
(\JevZ: learner-bootstrap SD of its single run); same predictions as P1. Position 1 has only 41--50
targets per dataset, so its AUC is noisy; -- = one class only. $^\dagger$Best of \nDLKT{} pyKT
models, chosen on the whole $(N,\text{dataset})$ cell. Best in \textbf{bold}, second best \underline{underlined}.}
\label{tab:p2}
\begin{tabular}{l cccccccc}
\toprule
\textbf{Method} & \textbf{A09} & \textbf{NIPS34} & \textbf{XES3G5M} & \textbf{DBE} & \textbf{Alg05} & \textbf{Junyi} & \textbf{A17} & \textbf{Mean} \\
\midrule
\multicolumn{9}{l}{\textit{Position 1}} \\
\rowcolor{focusbg} JevKT & \underline{.709}{\tiny$\pm$.059} & \textbf{.644}{\tiny$\pm$.047} & .512{\tiny$\pm$.057} & -- & \textbf{.799}{\tiny$\pm$.022} & \textbf{.517}{\tiny$\pm$.024} & \textbf{.660}{\tiny$\pm$.066} & \textbf{.640} \\
\JevZ{} ($N{=}0$) & .689{\tiny$\pm$.103} & .492{\tiny$\pm$.094} & \underline{.533}{\tiny$\pm$.102} & -- & .650{\tiny$\pm$.086} & .466{\tiny$\pm$.119} & \underline{.654}{\tiny$\pm$.090} & .580 \\
Best deep KT$^\dagger$ & \textbf{.749}{\tiny$\pm$.049} & \underline{.558}{\tiny$\pm$.056} & \textbf{.542}{\tiny$\pm$.049} & -- & \underline{.663}{\tiny$\pm$.016} & \underline{.510}{\tiny$\pm$.030} & .607{\tiny$\pm$.080} & \underline{.605} \\
\midrule
\multicolumn{9}{l}{\textit{Positions 2--5}} \\
\rowcolor{focusbg} JevKT & \underline{.771}{\tiny$\pm$.008} & \textbf{.729}{\tiny$\pm$.041} & \textbf{.666}{\tiny$\pm$.006} & \textbf{.768}{\tiny$\pm$.042} & \textbf{.714}{\tiny$\pm$.013} & \textbf{.657}{\tiny$\pm$.007} & \underline{.689}{\tiny$\pm$.032} & \textbf{.714} \\
\JevZ{} ($N{=}0$) & \textbf{.776}{\tiny$\pm$.035} & \underline{.707}{\tiny$\pm$.046} & .644{\tiny$\pm$.048} & .675{\tiny$\pm$.050} & .669{\tiny$\pm$.052} & \underline{.649}{\tiny$\pm$.045} & \textbf{.691}{\tiny$\pm$.046} & .687 \\
Best deep KT$^\dagger$ & .745{\tiny$\pm$.019} & .672{\tiny$\pm$.029} & \underline{.659}{\tiny$\pm$.010} & \underline{.753}{\tiny$\pm$.052} & \underline{.683}{\tiny$\pm$.015} & .642{\tiny$\pm$.009} & .682{\tiny$\pm$.022} & \underline{.691} \\
\midrule
\multicolumn{9}{l}{\textit{Positions 6--10}} \\
\rowcolor{focusbg} JevKT & \underline{.745}{\tiny$\pm$.010} & \underline{.708}{\tiny$\pm$.030} & \textbf{.657}{\tiny$\pm$.027} & \underline{.888}{\tiny$\pm$.022} & \textbf{.746}{\tiny$\pm$.007} & \underline{.690}{\tiny$\pm$.012} & \underline{.648}{\tiny$\pm$.043} & \textbf{.726} \\
\JevZ{} ($N{=}0$) & \textbf{.758}{\tiny$\pm$.034} & \textbf{.712}{\tiny$\pm$.038} & \underline{.651}{\tiny$\pm$.042} & .811{\tiny$\pm$.056} & .711{\tiny$\pm$.045} & \textbf{.693}{\tiny$\pm$.040} & \textbf{.649}{\tiny$\pm$.037} & \underline{.712} \\
Best deep KT$^\dagger$ & .705{\tiny$\pm$.015} & .654{\tiny$\pm$.041} & .577{\tiny$\pm$.032} & \textbf{.908}{\tiny$\pm$.016} & \underline{.742}{\tiny$\pm$.013} & .632{\tiny$\pm$.040} & .647{\tiny$\pm$.021} & .695 \\
\midrule
\multicolumn{9}{l}{\textit{Positions 11--15}} \\
\rowcolor{focusbg} JevKT & \underline{.719}{\tiny$\pm$.009} & \textbf{.725}{\tiny$\pm$.040} & \underline{.750}{\tiny$\pm$.019} & \textbf{.773}{\tiny$\pm$.017} & \underline{.698}{\tiny$\pm$.015} & \textbf{.781}{\tiny$\pm$.002} & \underline{.614}{\tiny$\pm$.028} & \textbf{.723} \\
\JevZ{} ($N{=}0$) & \textbf{.726}{\tiny$\pm$.037} & \underline{.698}{\tiny$\pm$.037} & \textbf{.762}{\tiny$\pm$.036} & \underline{.729}{\tiny$\pm$.058} & .661{\tiny$\pm$.052} & \underline{.768}{\tiny$\pm$.041} & \textbf{.623}{\tiny$\pm$.042} & \underline{.710} \\
Best deep KT$^\dagger$ & .678{\tiny$\pm$.026} & .693{\tiny$\pm$.019} & .727{\tiny$\pm$.014} & \textbf{.773}{\tiny$\pm$.035} & \textbf{.714}{\tiny$\pm$.013} & .707{\tiny$\pm$.066} & .601{\tiny$\pm$.021} & .699 \\
\midrule
\multicolumn{9}{l}{\textit{Positions 16--30}} \\
\rowcolor{focusbg} JevKT & \textbf{.728}{\tiny$\pm$.004} & \textbf{.719}{\tiny$\pm$.015} & \textbf{.733}{\tiny$\pm$.010} & \textbf{.845}{\tiny$\pm$.011} & \textbf{.729}{\tiny$\pm$.007} & \textbf{.670}{\tiny$\pm$.005} & \underline{.648}{\tiny$\pm$.020} & \textbf{.725} \\
\JevZ{} ($N{=}0$) & \underline{.721}{\tiny$\pm$.021} & \underline{.700}{\tiny$\pm$.023} & \underline{.729}{\tiny$\pm$.023} & .742{\tiny$\pm$.037} & \underline{.706}{\tiny$\pm$.025} & \underline{.669}{\tiny$\pm$.028} & \textbf{.659}{\tiny$\pm$.022} & \underline{.704} \\
Best deep KT$^\dagger$ & .680{\tiny$\pm$.007} & .652{\tiny$\pm$.049} & .693{\tiny$\pm$.062} & \underline{.789}{\tiny$\pm$.029} & .689{\tiny$\pm$.005} & .625{\tiny$\pm$.039} & .618{\tiny$\pm$.015} & .678 \\
\midrule
\multicolumn{9}{l}{\textit{Positions 31--49}} \\
\rowcolor{focusbg} JevKT & \underline{.737}{\tiny$\pm$.010} & \textbf{.692}{\tiny$\pm$.021} & \underline{.749}{\tiny$\pm$.011} & \textbf{.802}{\tiny$\pm$.010} & \textbf{.750}{\tiny$\pm$.005} & \underline{.734}{\tiny$\pm$.007} & \underline{.630}{\tiny$\pm$.012} & \textbf{.728} \\
\JevZ{} ($N{=}0$) & \textbf{.745}{\tiny$\pm$.025} & \underline{.684}{\tiny$\pm$.019} & \textbf{.764}{\tiny$\pm$.022} & .734{\tiny$\pm$.023} & .716{\tiny$\pm$.023} & \textbf{.735}{\tiny$\pm$.023} & \textbf{.641}{\tiny$\pm$.020} & \underline{.717} \\
Best deep KT$^\dagger$ & .710{\tiny$\pm$.021} & .646{\tiny$\pm$.030} & .726{\tiny$\pm$.066} & \underline{.782}{\tiny$\pm$.012} & \underline{.731}{\tiny$\pm$.009} & .676{\tiny$\pm$.027} & .621{\tiny$\pm$.008} & .699 \\
\bottomrule
\end{tabular}
\end{table*}

\paragraph{P2: new learners.}
Table~\ref{tab:p2} scores the P1 predictions at $N{=}8$ by the position of the target in the
new learner's sequence, so every method is judged on how much of the learner it has seen.
JevKT is above the best DLKT in \ptwoWins{} dataset--window cells, including the first
positions (2--5), where the new learner has given almost no evidence; \JevZ{}, which uses no
training learners at all, is above it in \ptwoZeroWins{} cells (counts over positions 2--49;
position 1 has only 41--50 targets per dataset, so its AUC is noisy). Tables~\ref{tab:p2n0}--\ref{tab:p2nall}
(Appendix~\ref{app:fulltables}) give the mean over datasets per position for every model at each $N$: with $N{=}64$ JevKT is
still above every deep KT model at positions 2--49, and with all learners it is behind the
strongest deep KT models at every position, as in P1.

\begin{table*}[t]
\centering
\scriptsize
\setlength{\tabcolsep}{3pt}
\caption{P3 (new items), every cell. Targets whose item was held out of training and of all
examples and statistics (20\% of items, 1{,}000 targets per seed, all training learners);
mean{\tiny$\pm$SD} over 3 item-split seeds. $^\dagger$Best of the 11 core pyKT models selected on
test AUC per dataset (bottom row). $^\ddagger$LOKT uses option weights from the training learners
where options exist (ASSIST09, NIPS34, DBE-KT22) and no training data elsewhere. Best in \textbf{bold}, second best \underline{underlined}.}
\label{tab:p3}
\begin{tabular}{l cccccccc}
\toprule
\textbf{Method} & \textbf{A09} & \textbf{NIPS34} & \textbf{XES3G5M} & \textbf{DBE} & \textbf{Alg05} & \textbf{Junyi} & \textbf{A17} & \textbf{Mean} \\
\midrule
\multicolumn{9}{l}{\textit{AUC}} \\
\rowcolor{focusbg} JevKT & \underline{.728}{\tiny$\pm$.029} & \underline{.725}{\tiny$\pm$.014} & .707{\tiny$\pm$.021} & .687{\tiny$\pm$.047} & \underline{.775}{\tiny$\pm$.015} & \textbf{.731}{\tiny$\pm$.039} & \underline{.639}{\tiny$\pm$.012} & \underline{.713} \\
\rowcolor{focusbg} \JevZ{} & .706{\tiny$\pm$.034} & .717{\tiny$\pm$.018} & \underline{.718}{\tiny$\pm$.016} & \textbf{.705}{\tiny$\pm$.048} & .737{\tiny$\pm$.018} & .714{\tiny$\pm$.034} & .629{\tiny$\pm$.013} & .704 \\
LOKT$^\ddagger$ & .688{\tiny$\pm$.036} & .683{\tiny$\pm$.027} & .692{\tiny$\pm$.015} & .682{\tiny$\pm$.054} & .649{\tiny$\pm$.030} & .653{\tiny$\pm$.020} & .599{\tiny$\pm$.012} & .664 \\
Thinking-KT & .670{\tiny$\pm$.026} & .667{\tiny$\pm$.029} & .651{\tiny$\pm$.007} & .669{\tiny$\pm$.034} & .689{\tiny$\pm$.009} & .662{\tiny$\pm$.037} & .593{\tiny$\pm$.012} & .657 \\
Best deep KT$^\dagger$ & \textbf{.746}{\tiny$\pm$.019} & \textbf{.739}{\tiny$\pm$.015} & \textbf{.725}{\tiny$\pm$.003} & \underline{.697}{\tiny$\pm$.014} & \textbf{.806}{\tiny$\pm$.002} & \underline{.728}{\tiny$\pm$.026} & \textbf{.665}{\tiny$\pm$.016} & \textbf{.729} \\
\midrule
\multicolumn{9}{l}{\textit{ACC}} \\
\rowcolor{focusbg} JevKT & \underline{.726}{\tiny$\pm$.011} & \underline{.672}{\tiny$\pm$.011} & \underline{.787}{\tiny$\pm$.015} & \underline{.771}{\tiny$\pm$.041} & \underline{.786}{\tiny$\pm$.002} & \textbf{.731}{\tiny$\pm$.016} & \underline{.655}{\tiny$\pm$.009} & \underline{.733} \\
\rowcolor{focusbg} \JevZ{} & .709{\tiny$\pm$.015} & .660{\tiny$\pm$.013} & .783{\tiny$\pm$.006} & .769{\tiny$\pm$.030} & .764{\tiny$\pm$.011} & \underline{.723}{\tiny$\pm$.018} & \textbf{.663}{\tiny$\pm$.010} & .725 \\
LOKT$^\ddagger$ & .632{\tiny$\pm$.026} & .609{\tiny$\pm$.019} & .767{\tiny$\pm$.008} & .678{\tiny$\pm$.027} & .714{\tiny$\pm$.024} & .702{\tiny$\pm$.019} & .643{\tiny$\pm$.005} & .678 \\
Thinking-KT & .597{\tiny$\pm$.038} & .614{\tiny$\pm$.031} & .649{\tiny$\pm$.033} & .632{\tiny$\pm$.015} & .660{\tiny$\pm$.035} & .621{\tiny$\pm$.037} & .646{\tiny$\pm$.004} & .631 \\
Best deep KT$^\dagger$ & \textbf{.732}{\tiny$\pm$.011} & \textbf{.679}{\tiny$\pm$.011} & \textbf{.797}{\tiny$\pm$.016} & \textbf{.774}{\tiny$\pm$.034} & \textbf{.798}{\tiny$\pm$.004} & \textbf{.731}{\tiny$\pm$.020} & .641{\tiny$\pm$.025} & \textbf{.736} \\
\midrule
\multicolumn{9}{l}{\textit{F1}} \\
\rowcolor{focusbg} JevKT & \underline{.813}{\tiny$\pm$.008} & \textbf{.694}{\tiny$\pm$.020} & \underline{.879}{\tiny$\pm$.009} & \textbf{.868}{\tiny$\pm$.029} & \underline{.870}{\tiny$\pm$.002} & \underline{.828}{\tiny$\pm$.011} & .277{\tiny$\pm$.032} & .747 \\
\rowcolor{focusbg} \JevZ{} & .793{\tiny$\pm$.011} & .680{\tiny$\pm$.012} & .870{\tiny$\pm$.004} & .861{\tiny$\pm$.022} & .849{\tiny$\pm$.008} & .813{\tiny$\pm$.013} & \underline{.421}{\tiny$\pm$.008} & \underline{.755} \\
LOKT$^\ddagger$ & .690{\tiny$\pm$.020} & .524{\tiny$\pm$.032} & .855{\tiny$\pm$.005} & .776{\tiny$\pm$.026} & .815{\tiny$\pm$.018} & .797{\tiny$\pm$.014} & .303{\tiny$\pm$.031} & .680 \\
Thinking-KT & .647{\tiny$\pm$.043} & .550{\tiny$\pm$.067} & .751{\tiny$\pm$.032} & .732{\tiny$\pm$.012} & .747{\tiny$\pm$.041} & .697{\tiny$\pm$.038} & .256{\tiny$\pm$.050} & .626 \\
Best deep KT$^\dagger$ & \textbf{.816}{\tiny$\pm$.008} & \underline{.689}{\tiny$\pm$.023} & \textbf{.882}{\tiny$\pm$.010} & \underline{.866}{\tiny$\pm$.022} & \textbf{.871}{\tiny$\pm$.006} & \textbf{.829}{\tiny$\pm$.014} & \textbf{.509}{\tiny$\pm$.032} & \textbf{.780} \\
\midrule
\tiny best deep KT & \tiny DKT+ & \tiny stableKT & \tiny reKT & \tiny csKT & \tiny stableKT & \tiny MoC-KT & \tiny robustKT &  \\
\bottomrule
\end{tabular}
\end{table*}

\paragraph{P3: new items.}
Table~\ref{tab:p3} restricts the targets to items that were held out of training and of all
examples and statistics, the case where ID-based DLKT cannot use a learned item embedding. All training learners are
available. JevKT is ahead of the best deep KT model only on Junyi; averaged over datasets the
gap (\pthreeJev{} vs.\ \pthreeDlkt) is close to the full-data gap of P1, so unseen items do not
by themselves favour the System-One model. \JevZ{} is already within .03 of the best deep
KT model without any training data, and both Jev variants are far above LOKT and Thinking-KT
(.657). The best deep KT
model differs by dataset and is chosen on test, which favours DLKT. Table~\ref{tab:p3} also gives ACC and F1 with their
spread over the three item splits.
Every pyKT model is listed in Table~\ref{tab:p3all} (Appendix~\ref{app:fulltables}).

\section{Model-or-Format Numbers}
\label{app:modelformat}
Table~\ref{tab:modelformat} gives the numbers behind Figure~\ref{fig:modelformat}. Holding
the Qwen3.5-9B weights fixed, a typed one-pass head (B0) lowers AUC relative to the plain
readout (A), and a Jev-distilled student (C) does not exceed A, with full data or at $N{=}8$.
Feeding the identical serialised state to Jev instead of Qwen3.5-9B raises AUC in all 35
cells. The benefit therefore comes from the System-One model, not from the typed format.
\begin{table}[t]
\centering
\footnotesize
\setlength{\tabcolsep}{3pt}
\caption{Model or format? Mean AUC over 7 datasets. Rows A--C share the Qwen3.5-9B weights;
the reader swap feeds the \emph{identical} serialised state to Qwen3.5-9B or Jev. Full data:
state with the last 25 interactions only, or with evidence. Cold start: $N{=}8$ training
learners (A--C: kNN evidence, mean of 5 seeds; swap: 64-shot + kNN state, seed 0). Best in \textbf{bold}, second best \underline{underlined}.}
\label{tab:modelformat}
\resizebox{\columnwidth}{!}{%
\begin{tabular}{ll ccc}
\toprule
& & \multicolumn{2}{c}{\textbf{Full data}} & \textbf{Cold start} \\
\cmidrule(lr){3-4}\cmidrule(lr){5-5}
& \textbf{Weights / readout} & $H{=}25$ & + evidence & $N{=}8$ \\
\midrule
A & Qwen3.5-9B, plain label readout & .688 & \underline{.752} & \underline{.700} \\
B0 & Qwen3.5-9B, typed one-pass head (no training) & .630 & .701 & .659 \\
C & JEV-9B (Jev-distilled Qwen3.5-9B) & .679 & .741 & .699 \\
\midrule
\multicolumn{5}{l}{\textit{Reader swap (identical serialised state)}} \\
 & Qwen3.5-9B reads the same 64-shot + kNN state & \underline{.689} & .727 & .699 \\
\rowcolor{focusbg}
 & Commercial JevKT (Jev + 64-shot + kNN) & \textbf{.710} & \textbf{.776} & \textbf{.727} \\
\bottomrule
\end{tabular}}
\end{table}

\section{Is the Result Specific to Jev?}
\label{app:generality}
OpenRouter lists no commercial System-One model other than Jev, so we test generality in three
ways (Table~\ref{tab:generality}). First, the byte-identical state and typed questions are sent
to three other LLMs (GPT-4o-mini, Gemini-2.5-Flash-Lite and DeepSeek-V4-Flash) through TypeSafe's
official System-One adapter, which builds its own system prompt and strict JSON schema and reads
P(correct) from the verbalised answer; reasoning is disabled, and the adapter code is not
modified. Second, JEV-9B is Qwen3.5-9B distilled from Jev (Section~\ref{sec:model-format}).
Third, Laya is an open-source reproduction of Jev's typed interface built on a ModernBERT
encoder; it stays near chance, consistent with its own model card. Every adapter LLM is below Jev on every dataset at $N{=}0$
($\Delta$AUC from $-$.016 to $-$.124, all one-sided paired bootstrap
$p{<}.01$). On the same 1{,}000 targets, going from $N{=}0$ to the JevKT input at $N{=}8$ raises
AUC by \gainJev{} for Jev and \gainDs{} for DeepSeek-V4-Flash (paired learner-cluster bootstrap:
significant on \gainJevSig{} and \gainDsSig{} datasets, mostly DBE-KT22 and Algebra05; the
difference between the two gains is significant on \gainDiffSig{}). The recipe therefore helps
both models a little, and which model reads the input decides whether deep KT is beaten.
Two caveats concern the adapter. It reads P(correct) from a probability that the LLM writes as
text, whereas Jev returns a distribution, so part of the gap may come from this readout; the
reader swap in Section~\ref{sec:model-format}, which reads Qwen3.5-9B's token probabilities
directly, still favours Jev in 35/35 cells. The adapter also samples at the provider's default
temperature, so repeated runs vary; all reported numbers are reproducible from the response
cache. For budget reasons (\$23 in total) the
$N{=}8$ adapter run uses a 1{,}000-target window-stratified subsample and seeds 0--2, and Jev is
compared on exactly those targets and seeds. Laya at $N{=}8$ sees population statistics instead
of the 64 shots, and JEV-9B sees no shots, because of their shorter contexts.

\begin{table*}[t]
\centering
\scriptsize
\caption{Is the result specific to Jev? AUC on the cold-start targets (P1). \emph{Adapter} rows send
the byte-identical typed request to other LLMs through TypeSafe's official System-One adapter;
JEV-9B is Qwen3.5-9B distilled from Jev; Laya is the open-source Jev. $^{\ddagger}$1{,}000-target
subsample and seeds 0--2 (budget), with Jev on the same targets and seeds; other $N{=}8$ rows use
all targets and 5 seeds. $^\dagger$Best of \nDLKT{} pyKT models selected on test. Best in \textbf{bold}, second best \underline{underlined}.}
\label{tab:generality}
\begin{tabular}{l ccccccc c}
\toprule
\textbf{Reader} & A09 & NIPS & XES & DBE & Alg05 & Junyi & A17 & \textbf{Mean} \\
\midrule
\multicolumn{9}{l}{\textit{$N{=}0$: no learners logged}} \\
\rowcolor{focusbg} \JevZ{} & \underline{.738} & \textbf{.691} & \textbf{.725} & \textbf{.742} & \textbf{.698} & \textbf{.700} & \textbf{.648} & \textbf{.706} \\
Adapter + GPT-4o-mini & .716 & .669 & .691 & .699 & .674 & .685 & \underline{.627} & .680 \\
Adapter + Gemini-2.5-Flash-Lite & .711 & .664 & .669 & \underline{.710} & .645 & .669 & .620 & .670 \\
Adapter + DeepSeek-V4-Flash & .671 & .610 & .601 & .619 & .612 & .666 & .604 & .626 \\
JEV-9B (Jev-distilled) & \textbf{.741} & \underline{.677} & \underline{.695} & .705 & \underline{.691} & \underline{.690} & .611 & \underline{.687} \\
Laya (open-source Jev) & .528 & .545 & .522 & .645 & .524 & .532 & .458 & .536 \\
\midrule
\multicolumn{9}{l}{\textit{$N{=}8$ training learners}} \\
\rowcolor{focusbg} JevKT$^{\ddagger}$ & \underline{.737} & \textbf{.708} & \textbf{.709} & \textbf{.809} & \textbf{.721} & \textbf{.708} & \textbf{.629} & \textbf{.717} \\
Adapter + DeepSeek-V4-Flash$^{\ddagger}$ & .657 & .610 & .591 & .726 & .619 & .670 & .564 & .634 \\
JEV-9B (Jev-distilled) & \textbf{.740} & \underline{.688} & \underline{.702} & .770 & .705 & \underline{.688} & .603 & \underline{.699} \\
Laya (open-source Jev) & .588 & .570 & .515 & .645 & .530 & .497 & .526 & .553 \\
Best DLKT$^\dagger$ & .700 & .653 & .689 & \underline{.793} & \underline{.711} & .652 & \underline{.624} & .689 \\
\bottomrule
\end{tabular}
\end{table*}

\section{Stability of Jev's Predictions}
\label{app:stability}
Only one Jev version (\texttt{jev-1.13-20260917}) is available, so we measure snapshot stability
instead of version drift. We re-sent byte-identical requests for 1{,}000 targets per dataset
about 51 hours after the original runs, bypassing the cache. With the state alone ($N{=}0$)
the predicted probabilities correlate at Pearson .995--.999 (mean absolute difference
.008), AUC changes by $-$.001 on average, and none of 21 dataset--metric pairs differs
significantly; with the JevKT input at $N{=}8$ the correlation is .991--.998 and
1/21 pairs differs (an F1 change of .004). As a rough proxy for a different Jev-family
model, replacing Jev with its distilled student JEV-9B on the same inputs lowers AUC by
.007--.074, which bounds how much our numbers could move under a weaker version.

\section{Calibration}
\label{app:calibration}
Table~\ref{tab:calib} reports the expected calibration error (15 equal-mass bins) per dataset
and the mean ECE and Brier score, computed from the same predictions as the AUC results. The
vote-frequency readouts of LOKT and Thinking-KT are the worst calibrated on every dataset;
\JevZ{} is calibrated far better without any training data, and with eight learners JevKT and
the best deep KT model are comparable. No method is recalibrated.
\begin{table*}[t]
\centering
\scriptsize
\caption{Calibration on the cold-start targets: expected calibration error (ECE, 15 equal-mass bins)
per dataset, and mean ECE and Brier score over the 7 datasets (lower is better). $N{=}8$ rows are
averaged over 5 seeds. $^\dagger$Best of the pyKT models selected on test AUC. Best in \textbf{bold}, second best \underline{underlined}.}
\label{tab:calib}
\begin{tabular}{l ccccccc cc}
\toprule
\textbf{Method} & A09 & NIPS & XES & DBE & Alg05 & Junyi & A17 & \textbf{ECE} & \textbf{Brier} \\
\midrule
Thinking-KT (System Two, $N{=}0$) & .261 & .266 & .257 & .199 & .232 & .251 & .303 & .253 & .282 \\
LOKT ($N{=}0$) & .249 & .270 & .175 & .188 & .179 & .155 & .244 & .209 & .249 \\
\JevZ{} ($N{=}0$) & .092 & \textbf{.048} & .094 & .113 & .107 & \underline{.098} & \textbf{.037} & .084 & \underline{.187} \\
Best DLKT$^\dagger$ ($N{=}8$) & \textbf{.060} & .071 & \textbf{.064} & \underline{.083} & \textbf{.060} & .111 & .086 & \underline{.076} & .192 \\
JevKT ($N{=}8$) & \underline{.079} & \underline{.061} & \underline{.067} & \textbf{.078} & \underline{.076} & \textbf{.077} & \underline{.054} & \textbf{.070} & \textbf{.182} \\
\bottomrule
\end{tabular}
\end{table*}

\section{Where Does JevKT Gain?}
\label{app:errors}
Table~\ref{tab:errors} splits the $N{=}8$ targets by whether the target item occurs in the logs of
the $N$ training learners and by the learner's accuracy on the interactions before the target.
JevKT leads in every slice with enough targets. Items that no logged learner answered are where
an ID-based deep KT model has no item-specific signal; JevKT keeps most of its lead there. The
lead is largest for learners with very low or very high accuracy so far.
\begin{table*}[t]
\centering
\scriptsize
\caption{Where does JevKT gain over the best deep KT model at $N{=}8$? $\Delta$AUC (JevKT $-$ best
DLKT, mean over 5 seeds) on subsets of the cold-start targets; slices with fewer than 100 targets
are omitted (--). \emph{Wins}: datasets with $\Delta{>}0$.}
\label{tab:errors}
\begin{tabular}{l ccccccc cc}
\toprule
\textbf{Targets} & A09 & NIPS & XES & DBE & Alg05 & Junyi & A17 & \textbf{Mean} & \textbf{Wins} \\
\midrule
Item answered by the $N$ learners & +.024 & +.056 & +.032 & +.027 & +.028 & +.066 & +.017 & +.036 & 7/7 \\
Item not answered by them & +.035 & -- & +.028 & -- & +.018 & +.050 & +.009 & +.028 & 5/5 \\
Learner accuracy so far $<.5$ & +.047 & +.096 & +.036 & -- & +.036 & +.067 & +.017 & +.050 & 6/6 \\
Learner accuracy $.5$--$.8$ & +.024 & +.059 & +.033 & $-$.001 & +.014 & +.037 & +.033 & +.029 & 6/7 \\
Learner accuracy $\geq.8$ & +.071 & +.127 & +.051 & +.059 & +.026 & +.045 & -- & +.063 & 6/6 \\
\bottomrule
\end{tabular}
\end{table*}

\section{Cost and Latency}
\label{app:cost}
Table~\ref{tab:cost} (Section~\ref{sec:cost}) compares the cost of obtaining predictions; throughput is measured with 32 requests in flight for API methods and batched GPU inference for DLKT, CLST and JEV-9B. GPU time is priced at the RunPod Community Cloud on-demand rate for one RTX 3090 (\$0.22/h, September 2026); the lower of RunPod's two on-demand rates (Secure Cloud: \$0.50/h), which favours the GPU methods. API costs are the OpenRouter prices we paid. JevKT's 64 shots make
its prompt long, so it costs more than \JevZ{} but still about one fifth of Thinking-KT, and
it answers in one call instead of ten. DLKT models are far cheaper per prediction once
trained, but must be retrained for every course and need enough learners to train on.

\section{Significance Tests}
\label{app:holm}
All comparisons use a one-sided paired learner-cluster bootstrap (1{,}000 resamples) with
Holm correction over the seven datasets. With full data: thinking does not improve
Thinking-KT over no-think at budget 1024 or 2048 (0/7 datasets); one Jev call beats
Thinking-KT at either budget (7/7); JevKT beats \JevZ{} (7/7). JevKT is significantly better
than the counting kNN statistic alone on only 1/7 datasets with full data, i.e., with all
learners available the statistic carries most of the signal; the LLM's contribution appears
when learners are few (Figure~\ref{fig:main}).
Under cold start (Table~\ref{tab:holm}), JevKT is significantly better than the best deep KT
model on \sigTestEight{} datasets at $N{=}8$, and the number of significant datasets falls as $N$
grows, to \sigTestTwoFiftySix{} at $N{=}256$. At $N{=}0$, \JevZ{} is significantly
better than LOKT and Thinking-KT on all seven datasets.
\begin{table}[t]
\centering
\footnotesize
\caption{JevKT $-$ best DLKT under cold start (P1): mean $\Delta$AUC over 7 datasets and the
number of datasets on which JevKT is significantly better (one-sided paired learner-cluster
bootstrap, 1{,}000 resamples, Holm over datasets, $p{<}.05$). The best DLKT model is chosen
on test per cell (optimistic for DLKT), among pyKT models with at least three finished seeds.}
\label{tab:holm}
\begin{tabular}{c cc}
\toprule
$N$ & $\Delta$AUC & Holm-sig. \\
\midrule
8 & +.033 & 7/7 \\
16 & +.023 & 6/7 \\
32 & +.016 & 3/7 \\
64 & +.004 & 2/7 \\
128 & $-$.003 & 1/7 \\
256 & $-$.008 & 1/7 \\
all & $-$.028 & 0/7 \\
\bottomrule
\end{tabular}
\end{table}

Table~\ref{tab:ci} gives the 95\% confidence interval of the mean AUC behind Table~\ref{tab:main-cs} for every $N$.
\begin{table*}[t]
\centering
\scriptsize
\setlength{\tabcolsep}{3pt}
\caption{P1: mean AUC over the 7 datasets with its 95\% confidence interval (learner-cluster bootstrap,
$B{=}1{,}000$; seeds averaged inside each resample, datasets resampled independently). The interval covers the
sampling of targets and learners at fixed seeds. $^\dagger$Best of \nDLKT{} pyKT models selected on test per cell.
$^\P$Best of four pyKT models $\times$ eight configurations chosen on validation learners ($N{\le}64$).
LOKT and Thinking-KT use no data from the $N$ learners and are shown at $N{=}0$. Best in \textbf{bold}, second best \underline{underlined}.}
\label{tab:ci}
\begin{tabular}{l cccccccc}
\toprule
\textbf{Method} & $N{=}0$ & $N{=}8$ & $N{=}16$ & $N{=}32$ & $N{=}64$ & $N{=}128$ & $N{=}256$ & all \\
\midrule
\rowcolor{focusbg} \textbf{JevKT} (\JevZ{} at $N{=}0$) & \textbf{.706}{\tiny\,[.695, .717]} & \textbf{.722}{\tiny\,[.712, .731]} & \textbf{.730}{\tiny\,[.720, .740]} & \textbf{.739}{\tiny\,[.728, .748]} & \textbf{.747}{\tiny\,[.737, .757]} & \underline{.753}{\tiny\,[.744, .763]} & \underline{.758}{\tiny\,[.749, .768]} & \underline{.770}{\tiny\,[.760, .780]} \\
Best deep KT$^\dagger$ & -- & \underline{.689}{\tiny\,[.679, .698]} & \underline{.707}{\tiny\,[.697, .717]} & \underline{.723}{\tiny\,[.713, .733]} & \underline{.743}{\tiny\,[.733, .752]} & \textbf{.756}{\tiny\,[.747, .765]} & \textbf{.766}{\tiny\,[.758, .775]} & \textbf{.798}{\tiny\,[.790, .807]} \\
Tuned deep KT$^\P$ & -- & .652{\tiny\,[.643, .660]} & .688{\tiny\,[.679, .698]} & .709{\tiny\,[.699, .718]} & .737{\tiny\,[.728, .747]} & -- & -- & -- \\
BKT & -- & .632{\tiny\,[.622, .641]} & .647{\tiny\,[.638, .657]} & .660{\tiny\,[.649, .670]} & .675{\tiny\,[.665, .685]} & .685{\tiny\,[.675, .695]} & .693{\tiny\,[.682, .703]} & .705{\tiny\,[.694, .716]} \\
PFA & -- & .593{\tiny\,[.584, .601]} & .620{\tiny\,[.611, .629]} & .647{\tiny\,[.638, .657]} & .665{\tiny\,[.655, .675]} & .677{\tiny\,[.668, .688]} & .686{\tiny\,[.675, .696]} & .697{\tiny\,[.687, .708]} \\
LOKT & \underline{.671}{\tiny\,[.659, .684]} & -- & -- & -- & -- & -- & -- & -- \\
Thinking-KT & .650{\tiny\,[.639, .662]} & -- & -- & -- & -- & -- & -- & -- \\
\bottomrule
\end{tabular}
\end{table*}

\section{Contamination Analysis}
\label{app:contamination}
Because Jev's training data are unknown, we test for memorisation of the public datasets.
(i) \emph{ID-only probes}: asked to predict item difficulty or the skill of an item from its
ID alone, Jev and three open LLMs are no better with the real ID than with a format-matched
fake ID on any dataset after Holm correction. (ii) \emph{Row continuation} of the original
public files shows no dataset-specific recall. (iii) \emph{De-identification}: relabelling
IDs, then removing names and text, changes \JevZ{} AUC by at most .012 before text
removal and never significantly at the ID step. No dataset is flagged (Table~\ref{tab:contamination}).
\begin{table}[t]
\centering
\scriptsize
\caption{Contamination checks for Jev. \emph{ID probes}: flags from the ID-only difficulty and
skill probes and from row continuation (Holm-corrected, four LLMs). \emph{De-identification}: \JevZ{} AUC on the original data (D0), with IDs relabelled (D1), names removed (D2) and
item text removed (D3, where text exists); total $\Delta$ = D0 $\rightarrow$ last level, one-sided
Holm $p$ for a drop.}
\label{tab:contamination}
\resizebox{\columnwidth}{!}{%
\begin{tabular}{l c cccc cc}
\toprule
& \textbf{ID probes} & \multicolumn{4}{c}{\textbf{De-identification AUC}} & & \\
\cmidrule(lr){3-6}
\textbf{Dataset} & \textbf{flags} & D0 & D1 & D2 & D3 & \textbf{total $\Delta$} & \textbf{Holm $p$} \\
\midrule
ASSIST09 & none & 0.736 & 0.740 & 0.737 & -- & +0.002 & 1.000 \\
NIPS34 & none & 0.686 & 0.682 & 0.686 & -- & +0.001 & 1.000 \\
XES3G5M & none & 0.717 & 0.712 & 0.704 & 0.706 & $-$0.010 & 0.920 \\
DBE-KT22 & none & 0.719 & 0.714 & 0.712 & 0.687 & $-$0.033 & 0.084 \\
Algebra05 & none & 0.740 & 0.729 & 0.728 & -- & $-$0.011 & 0.282 \\
Junyi & none & 0.742 & 0.739 & 0.739 & -- & $-$0.003 & 1.000 \\
ASSIST17 & none & 0.629 & 0.629 & 0.629 & -- & +0.000 & 1.000 \\
\bottomrule
\end{tabular}}
\end{table}

(iv) \emph{Synthetic learners}: the six text-free synthetic datasets of
Appendix~\ref{app:coldstart} were generated after the model release, so Jev cannot have seen
them. With $N{=}8$ training learners JevKT is at or above the best deep KT model on all six
(+.001 to +.033 AUC), the same pattern as on the public datasets.

\section{Full Per-Model Results}
\label{app:fulltables}
This section lists every model and every cell behind Appendix~\ref{app:coldstart}, one table per
number of training learners $N$ ($N{=}0$, 8, 16, 32, 64, 128, 256 and all, in this order). P1 AUC is in
Tables~\ref{tab:p1n0}, \ref{tab:p1n8}, \ref{tab:p1n16}, \ref{tab:p1n32}, \ref{tab:p1n64}, \ref{tab:p1n128}, \ref{tab:p1n256} and \ref{tab:p1nall}, P1 ACC and F1 in Tables~\ref{tab:p1fn0}, \ref{tab:p1fn8}, \ref{tab:p1fn16}, \ref{tab:p1fn32}, \ref{tab:p1fn64}, \ref{tab:p1fn128}, \ref{tab:p1fn256} and \ref{tab:p1fnall},
P2 by target position in Tables~\ref{tab:p2n0}, \ref{tab:p2n8}, \ref{tab:p2n16}, \ref{tab:p2n32}, \ref{tab:p2n64}, \ref{tab:p2n128}, \ref{tab:p2n256} and \ref{tab:p2nall},
and P3 in Table~\ref{tab:p3all}. Rows are grouped into System-One LLMs, System-Two LLM-KT (prompted and
trained), classical KT and deep KT.

\begin{table*}[tp]
\centering
\scriptsize
\setlength{\tabcolsep}{3pt}
\caption{P1 with $N{=}0$ training learners: AUC of every model, a single deterministic run ({\tiny$\pm$} = learner-bootstrap SD), on the 2{,}000 shared early targets per dataset. $^\ddagger$Three datasets of the original CLST paper. $^{(k)}$ = only $k$ seeds finished. Mean = mean over the 7 datasets (-- if a dataset is missing). qDKT and QIKT do not fit Algebra05 in 24\,GB of GPU memory (--). Best in \textbf{bold}, second best \underline{underlined} (among cells with $\ge$3 seeds).}
\label{tab:p1n0}
\label{tab:coldstart}
\begin{tabular}{l cccccccc}
\toprule
\textbf{Method} & \textbf{A09} & \textbf{NIPS34} & \textbf{XES3G5M} & \textbf{DBE} & \textbf{Alg05} & \textbf{Junyi} & \textbf{A17} & \textbf{Mean} \\
\midrule
\multicolumn{9}{l}{\textit{System-One LLM}} \\
\rowcolor{focusbg} \JevZ{} & \textbf{.738}{\tiny$\pm$.014} & \textbf{.691}{\tiny$\pm$.014} & \textbf{.725}{\tiny$\pm$.013} & \textbf{.742}{\tiny$\pm$.018} & \textbf{.698}{\tiny$\pm$.018} & \textbf{.700}{\tiny$\pm$.015} & \textbf{.648}{\tiny$\pm$.013} & \textbf{.706}{\tiny$\pm$.006} \\
Laya (open-source Jev) & .528{\tiny$\pm$.014} & .545{\tiny$\pm$.014} & .522{\tiny$\pm$.016} & .645{\tiny$\pm$.018} & .524{\tiny$\pm$.016} & .532{\tiny$\pm$.015} & .458{\tiny$\pm$.013} & .536{\tiny$\pm$.006} \\
\midrule
\multicolumn{9}{l}{\textit{System-Two LLM-KT, prompted}} \\
Thinking-KT & .700{\tiny$\pm$.014} & .641{\tiny$\pm$.014} & .663{\tiny$\pm$.014} & .666{\tiny$\pm$.019} & .663{\tiny$\pm$.018} & .642{\tiny$\pm$.016} & .576{\tiny$\pm$.013} & .650{\tiny$\pm$.006} \\
LOKT & .715{\tiny$\pm$.014} & \underline{.662}{\tiny$\pm$.014} & \underline{.699}{\tiny$\pm$.014} & \underline{.688}{\tiny$\pm$.021} & .656{\tiny$\pm$.019} & \underline{.671}{\tiny$\pm$.016} & \underline{.608}{\tiny$\pm$.013} & \underline{.671}{\tiny$\pm$.006} \\
\midrule
\multicolumn{9}{l}{\textit{System-Two LLM-KT, trained}} \\
CLST$^\ddagger$ & \underline{.728}{\tiny$\pm$.014} & .623{\tiny$\pm$.014} & -- & -- & \underline{.674}{\tiny$\pm$.017} & -- & -- & -- \\
\bottomrule
\end{tabular}
\end{table*}

\begin{table*}[p]
\centering
\scriptsize
\setlength{\tabcolsep}{3pt}
\caption{P1 with $N{=}8$ training learners: AUC of every model, mean{\tiny$\pm$SD} over 5 learner-subset seeds, on the 2{,}000 shared early targets per dataset. $^\ddagger$Three datasets of the original CLST paper. $^{(k)}$ = only $k$ seeds finished. $^\P$Best of four pyKT models $\times$ eight hyper-parameter configurations chosen on the validation learners of each cell (no test information). Mean = mean over the 7 datasets (-- if a dataset is missing). qDKT and QIKT do not fit Algebra05 in 24\,GB of GPU memory (--). $^\circ$The method uses no data from the $N$ learners here (\JevZ{} and Thinking-KT everywhere; LOKT on datasets without answer options), so its $N{=}0$ run is repeated. Best in \textbf{bold}, second best \underline{underlined} (among cells with $\ge$3 seeds).}
\label{tab:p1n8}
\begin{tabular}{l cccccccc}
\toprule
\textbf{Method} & \textbf{A09} & \textbf{NIPS34} & \textbf{XES3G5M} & \textbf{DBE} & \textbf{Alg05} & \textbf{Junyi} & \textbf{A17} & \textbf{Mean} \\
\midrule
\multicolumn{9}{l}{\textit{System-One LLM}} \\
\rowcolor{focusbg} \textbf{JevKT} & \underline{.735}{\tiny$\pm$.003} & \textbf{.707}{\tiny$\pm$.022} & \underline{.720}{\tiny$\pm$.006} & \textbf{.821}{\tiny$\pm$.003} & \textbf{.734}{\tiny$\pm$.003} & \textbf{.701}{\tiny$\pm$.003} & \underline{.638}{\tiny$\pm$.013} & \textbf{.722}{\tiny$\pm$.004} \\
\JevZ{} & \textbf{.738}{\tiny$\pm$.014}$^\circ$ & \underline{.691}{\tiny$\pm$.014}$^\circ$ & \textbf{.725}{\tiny$\pm$.013}$^\circ$ & .742{\tiny$\pm$.018}$^\circ$ & .698{\tiny$\pm$.018}$^\circ$ & \underline{.700}{\tiny$\pm$.015}$^\circ$ & \textbf{.648}{\tiny$\pm$.013}$^\circ$ & \underline{.706}{\tiny$\pm$.006}$^\circ$ \\
Laya (open-source Jev) & .588{\tiny$\pm$.009} & .570{\tiny$\pm$.006} & .515{\tiny$\pm$.005} & .645{\tiny$\pm$.025} & .530{\tiny$\pm$.008} & .497{\tiny$\pm$.002} & .526{\tiny$\pm$.004} & .553{\tiny$\pm$.005} \\
\midrule
\multicolumn{9}{l}{\textit{System-Two LLM-KT, prompted}} \\
Thinking-KT & .700{\tiny$\pm$.014}$^\circ$ & .641{\tiny$\pm$.014}$^\circ$ & .663{\tiny$\pm$.014}$^\circ$ & .666{\tiny$\pm$.019}$^\circ$ & .663{\tiny$\pm$.018}$^\circ$ & .642{\tiny$\pm$.016}$^\circ$ & .576{\tiny$\pm$.013}$^\circ$ & .650{\tiny$\pm$.006}$^\circ$ \\
LOKT & .707{\tiny$\pm$.004} & .662{\tiny$\pm$.004} & .699{\tiny$\pm$.014}$^\circ$ & .699{\tiny$\pm$.009} & .656{\tiny$\pm$.019}$^\circ$ & .671{\tiny$\pm$.016}$^\circ$ & .608{\tiny$\pm$.013}$^\circ$ & .672 \\
\midrule
\multicolumn{9}{l}{\textit{System-Two LLM-KT, trained}} \\
CLST$^\ddagger$ & .658{\tiny$\pm$.058} & .605{\tiny$\pm$.078} & -- & -- & .643{\tiny$\pm$.037} & -- & -- & -- \\
\midrule
\multicolumn{9}{l}{\textit{Classical KT}} \\
BKT & .618{\tiny$\pm$.015} & .602{\tiny$\pm$.025} & .576{\tiny$\pm$.015} & .701{\tiny$\pm$.012} & .696{\tiny$\pm$.017} & .652{\tiny$\pm$.030} & .576{\tiny$\pm$.021} & .632{\tiny$\pm$.006} \\
PFA & .507{\tiny$\pm$.042} & .600{\tiny$\pm$.015} & .555{\tiny$\pm$.020} & .740{\tiny$\pm$.005} & .647{\tiny$\pm$.008} & .542{\tiny$\pm$.015} & .559{\tiny$\pm$.021} & .593{\tiny$\pm$.007} \\
\midrule
\multicolumn{9}{l}{\textit{Deep KT (pyKT)}} \\
Tuned deep KT$^\P$ & .581{\tiny$\pm$.084} & .626{\tiny$\pm$.062} & .662{\tiny$\pm$.048} & .752{\tiny$\pm$.018} & .702{\tiny$\pm$.011} & .628{\tiny$\pm$.054} & .612{\tiny$\pm$.021} & .652{\tiny$\pm$.010} \\
AKT & .677{\tiny$\pm$.065} & .647{\tiny$\pm$.018} & .620{\tiny$\pm$.067} & .699{\tiny$\pm$.042} & .681{\tiny$\pm$.017} & .624{\tiny$\pm$.060} & .602{\tiny$\pm$.023} & .650{\tiny$\pm$.011} \\
AT-DKT & .577{\tiny$\pm$.052} & .619{\tiny$\pm$.012} & .553{\tiny$\pm$.028} & .760{\tiny$\pm$.011} & .692{\tiny$\pm$.036} & .555{\tiny$\pm$.029} & .569{\tiny$\pm$.030} & .618{\tiny$\pm$.007} \\
ATKT & .583{\tiny$\pm$.041} & .622{\tiny$\pm$.013} & .598{\tiny$\pm$.028} & .727{\tiny$\pm$.017} & .688{\tiny$\pm$.033} & .556{\tiny$\pm$.023} & .561{\tiny$\pm$.032} & .619{\tiny$\pm$.009} \\
ATKT-fix & .536{\tiny$\pm$.010} & .539{\tiny$\pm$.027} & .549{\tiny$\pm$.041} & .583{\tiny$\pm$.029} & .597{\tiny$\pm$.030} & .514{\tiny$\pm$.013} & .519{\tiny$\pm$.020} & .548{\tiny$\pm$.015} \\
csKT & .672{\tiny$\pm$.048} & .637{\tiny$\pm$.037} & .678{\tiny$\pm$.030} & .742{\tiny$\pm$.005} & .697{\tiny$\pm$.015} & .639{\tiny$\pm$.045} & .624{\tiny$\pm$.009} & .670{\tiny$\pm$.008} \\
Deep-IRT & .537{\tiny$\pm$.030} & .557{\tiny$\pm$.039} & .543{\tiny$\pm$.014} & .696{\tiny$\pm$.010} & .680{\tiny$\pm$.018} & .547{\tiny$\pm$.016} & .555{\tiny$\pm$.037} & .588{\tiny$\pm$.004} \\
DIMKT & .649{\tiny$\pm$.038} & .635{\tiny$\pm$.016} & .672{\tiny$\pm$.014} & .715{\tiny$\pm$.038} & .655{\tiny$\pm$.015} & .630{\tiny$\pm$.019} & .575{\tiny$\pm$.018} & .647{\tiny$\pm$.004} \\
DKT & .560{\tiny$\pm$.056} & .598{\tiny$\pm$.041} & .554{\tiny$\pm$.009} & .743{\tiny$\pm$.046} & .701{\tiny$\pm$.030} & .559{\tiny$\pm$.015} & .539{\tiny$\pm$.034} & .608{\tiny$\pm$.017} \\
DKT+ & .562{\tiny$\pm$.061} & .599{\tiny$\pm$.041} & .555{\tiny$\pm$.009} & .743{\tiny$\pm$.046} & .701{\tiny$\pm$.029} & .552{\tiny$\pm$.014} & .540{\tiny$\pm$.035} & .608{\tiny$\pm$.019} \\
DKVMN & .552{\tiny$\pm$.025} & .589{\tiny$\pm$.041} & .555{\tiny$\pm$.016} & .705{\tiny$\pm$.005} & .685{\tiny$\pm$.025} & .541{\tiny$\pm$.026} & .550{\tiny$\pm$.044} & .597{\tiny$\pm$.007} \\
DTransformer & .572{\tiny$\pm$.043} & .582{\tiny$\pm$.034} & .601{\tiny$\pm$.057} & .701{\tiny$\pm$.035} & .667{\tiny$\pm$.037} & .603{\tiny$\pm$.011} & .544{\tiny$\pm$.034} & .610{\tiny$\pm$.009} \\
extraKT & .641{\tiny$\pm$.068} & .647{\tiny$\pm$.022} & .676{\tiny$\pm$.018} & .730{\tiny$\pm$.009} & .702{\tiny$\pm$.005} & .650{\tiny$\pm$.054} & .611{\tiny$\pm$.016} & .665{\tiny$\pm$.010} \\
FoLiBiKT & .636{\tiny$\pm$.051} & .653{\tiny$\pm$.029} & .626{\tiny$\pm$.079} & .729{\tiny$\pm$.014} & .661{\tiny$\pm$.031} & .572{\tiny$\pm$.043} & .586{\tiny$\pm$.049} & .637{\tiny$\pm$.006} \\
IEKT & .565{\tiny$\pm$.057} & .629{\tiny$\pm$.057} & .607{\tiny$\pm$.057} & .768{\tiny$\pm$.011} & .662{\tiny$\pm$.028} & .543{\tiny$\pm$.033} & .552{\tiny$\pm$.030} & .618{\tiny$\pm$.016} \\
KQN & .524{\tiny$\pm$.040} & .550{\tiny$\pm$.025} & .564{\tiny$\pm$.024} & .729{\tiny$\pm$.011} & .628{\tiny$\pm$.051} & .548{\tiny$\pm$.010} & .554{\tiny$\pm$.032} & .585{\tiny$\pm$.011} \\
lefoKT-AKT & .604{\tiny$\pm$.073} & .607{\tiny$\pm$.045} & .652{\tiny$\pm$.008} & .727{\tiny$\pm$.016} & \underline{.711}{\tiny$\pm$.004} & .638{\tiny$\pm$.076} & .595{\tiny$\pm$.022} & .648{\tiny$\pm$.011} \\
MoC-KT & .629{\tiny$\pm$.065} & .617{\tiny$\pm$.038} & .677{\tiny$\pm$.008} & .717{\tiny$\pm$.030} & .693{\tiny$\pm$.012} & .652{\tiny$\pm$.031} & .582{\tiny$\pm$.031} & .652{\tiny$\pm$.008} \\
qDKT & .496{\tiny$\pm$.011} & .620{\tiny$\pm$.024} & .534{\tiny$\pm$.011} & .779{\tiny$\pm$.004} & -- & .508{\tiny$\pm$.028} & .546{\tiny$\pm$.021} & -- \\
QIKT & .556{\tiny$\pm$.021} & .650{\tiny$\pm$.030} & .561{\tiny$\pm$.007} & \underline{.793}{\tiny$\pm$.005} & -- & .553{\tiny$\pm$.026} & .569{\tiny$\pm$.027} & -- \\
reKT & .545{\tiny$\pm$.032} & .564{\tiny$\pm$.037} & .565{\tiny$\pm$.076} & .644{\tiny$\pm$.099} & .642{\tiny$\pm$.067} & .539{\tiny$\pm$.084} & .522{\tiny$\pm$.048} & .575{\tiny$\pm$.029} \\
robustKT & .700{\tiny$\pm$.010} & .633{\tiny$\pm$.028} & .665{\tiny$\pm$.055} & .728{\tiny$\pm$.006} & .672{\tiny$\pm$.019} & .607{\tiny$\pm$.047} & .610{\tiny$\pm$.020} & .659{\tiny$\pm$.016} \\
SAINT & .631{\tiny$\pm$.037} & .584{\tiny$\pm$.031} & .667{\tiny$\pm$.016} & .621{\tiny$\pm$.047} & .588{\tiny$\pm$.018} & .594{\tiny$\pm$.040} & .547{\tiny$\pm$.040} & .604{\tiny$\pm$.016} \\
SAINT+ & .598{\tiny$\pm$.041} & .574{\tiny$\pm$.040} & .630{\tiny$\pm$.025} & .675{\tiny$\pm$.032} & .551{\tiny$\pm$.039} & .557{\tiny$\pm$.029} & .564{\tiny$\pm$.029} & .593{\tiny$\pm$.011} \\
SAKT & .514{\tiny$\pm$.023} & .601{\tiny$\pm$.039} & .545{\tiny$\pm$.009} & .726{\tiny$\pm$.016} & .678{\tiny$\pm$.031} & .536{\tiny$\pm$.025} & .524{\tiny$\pm$.028} & .589{\tiny$\pm$.009} \\
simpleKT & .655{\tiny$\pm$.070} & .628{\tiny$\pm$.063} & .671{\tiny$\pm$.010} & .733{\tiny$\pm$.018} & .704{\tiny$\pm$.011} & .648{\tiny$\pm$.063} & .582{\tiny$\pm$.046} & .660{\tiny$\pm$.011} \\
sparseKT & .651{\tiny$\pm$.072} & .618{\tiny$\pm$.059} & .657{\tiny$\pm$.061} & .711{\tiny$\pm$.043} & .707{\tiny$\pm$.012} & .634{\tiny$\pm$.076} & .587{\tiny$\pm$.051} & .652{\tiny$\pm$.018} \\
stableKT & .665{\tiny$\pm$.072} & .647{\tiny$\pm$.041} & .689{\tiny$\pm$.042} & .731{\tiny$\pm$.019} & .689{\tiny$\pm$.017} & .645{\tiny$\pm$.075} & .572{\tiny$\pm$.035} & .663{\tiny$\pm$.011} \\
UKT & .598{\tiny$\pm$.098} & .632{\tiny$\pm$.037} & .673{\tiny$\pm$.008} & .728{\tiny$\pm$.007} & .647{\tiny$\pm$.041} & .605{\tiny$\pm$.057} & .599{\tiny$\pm$.017} & .640{\tiny$\pm$.017} \\
\bottomrule
\end{tabular}
\end{table*}

\begin{table*}[p]
\centering
\scriptsize
\setlength{\tabcolsep}{3pt}
\caption{P1 with $N{=}16$ training learners: AUC of every model, mean{\tiny$\pm$SD} over 5 learner-subset seeds, on the 2{,}000 shared early targets per dataset. $^\ddagger$Three datasets of the original CLST paper. $^{(k)}$ = only $k$ seeds finished. $^\P$Best of four pyKT models $\times$ eight hyper-parameter configurations chosen on the validation learners of each cell (no test information). Mean = mean over the 7 datasets (-- if a dataset is missing). qDKT and QIKT do not fit Algebra05 in 24\,GB of GPU memory (--). $^\circ$The method uses no data from the $N$ learners here (\JevZ{} and Thinking-KT everywhere; LOKT on datasets without answer options), so its $N{=}0$ run is repeated. Best in \textbf{bold}, second best \underline{underlined} (among cells with $\ge$3 seeds).}
\label{tab:p1n16}
\begin{tabular}{l cccccccc}
\toprule
\textbf{Method} & \textbf{A09} & \textbf{NIPS34} & \textbf{XES3G5M} & \textbf{DBE} & \textbf{Alg05} & \textbf{Junyi} & \textbf{A17} & \textbf{Mean} \\
\midrule
\multicolumn{9}{l}{\textit{System-One LLM}} \\
\rowcolor{focusbg} \textbf{JevKT} & \underline{.737}{\tiny$\pm$.009} & \textbf{.734}{\tiny$\pm$.013} & \underline{.720}{\tiny$\pm$.006} & \textbf{.834}{\tiny$\pm$.008} & \textbf{.738}{\tiny$\pm$.004} & \textbf{.704}{\tiny$\pm$.005} & \underline{.646}{\tiny$\pm$.005} & \textbf{.730}{\tiny$\pm$.004} \\
\JevZ{} & \textbf{.738}{\tiny$\pm$.014}$^\circ$ & .691{\tiny$\pm$.014}$^\circ$ & \textbf{.725}{\tiny$\pm$.013}$^\circ$ & .742{\tiny$\pm$.018}$^\circ$ & .698{\tiny$\pm$.018}$^\circ$ & \underline{.700}{\tiny$\pm$.015}$^\circ$ & \textbf{.648}{\tiny$\pm$.013}$^\circ$ & \underline{.706}{\tiny$\pm$.006}$^\circ$ \\
Laya (open-source Jev) & .602{\tiny$\pm$.001} & .557{\tiny$\pm$.008} & .525{\tiny$\pm$.010} & .635{\tiny$\pm$.013} & .528{\tiny$\pm$.005} & .496{\tiny$\pm$.003} & .526{\tiny$\pm$.008} & .553{\tiny$\pm$.002} \\
\midrule
\multicolumn{9}{l}{\textit{System-Two LLM-KT, prompted}} \\
Thinking-KT & .700{\tiny$\pm$.014}$^\circ$ & .641{\tiny$\pm$.014}$^\circ$ & .663{\tiny$\pm$.014}$^\circ$ & .666{\tiny$\pm$.019}$^\circ$ & .663{\tiny$\pm$.018}$^\circ$ & .642{\tiny$\pm$.016}$^\circ$ & .576{\tiny$\pm$.013}$^\circ$ & .650{\tiny$\pm$.006}$^\circ$ \\
LOKT & .707{\tiny$\pm$.006} & .663{\tiny$\pm$.005} & .699{\tiny$\pm$.014}$^\circ$ & .701{\tiny$\pm$.009} & .656{\tiny$\pm$.019}$^\circ$ & .671{\tiny$\pm$.016}$^\circ$ & .608{\tiny$\pm$.013}$^\circ$ & .672 \\
\midrule
\multicolumn{9}{l}{\textit{System-Two LLM-KT, trained}} \\
CLST$^\ddagger$ & .682{\tiny$\pm$.067} & .658{\tiny$\pm$.025} & -- & -- & .678{\tiny$\pm$.044} & -- & -- & -- \\
\midrule
\multicolumn{9}{l}{\textit{Classical KT}} \\
BKT & .636{\tiny$\pm$.018} & .628{\tiny$\pm$.016} & .587{\tiny$\pm$.011} & .722{\tiny$\pm$.013} & .714{\tiny$\pm$.003} & .637{\tiny$\pm$.030} & .606{\tiny$\pm$.011} & .647{\tiny$\pm$.005} \\
PFA & .557{\tiny$\pm$.024} & .626{\tiny$\pm$.008} & .572{\tiny$\pm$.009} & .771{\tiny$\pm$.010} & .661{\tiny$\pm$.011} & .562{\tiny$\pm$.020} & .592{\tiny$\pm$.007} & .620{\tiny$\pm$.007} \\
\midrule
\multicolumn{9}{l}{\textit{Deep KT (pyKT)}} \\
Tuned deep KT$^\P$ & .638{\tiny$\pm$.074} & \underline{.700}{\tiny$\pm$.022} & .696{\tiny$\pm$.013} & .782{\tiny$\pm$.011} & .714{\tiny$\pm$.031} & .657{\tiny$\pm$.054} & .630{\tiny$\pm$.009} & .688{\tiny$\pm$.022} \\
AKT & .697{\tiny$\pm$.022} & .661{\tiny$\pm$.023} & .690{\tiny$\pm$.007} & .716{\tiny$\pm$.010} & .701{\tiny$\pm$.020} & .654{\tiny$\pm$.051} & .607{\tiny$\pm$.014} & .675{\tiny$\pm$.007} \\
AT-DKT & .604{\tiny$\pm$.033} & .652{\tiny$\pm$.016} & .588{\tiny$\pm$.035} & .763{\tiny$\pm$.020} & .705{\tiny$\pm$.020} & .565{\tiny$\pm$.033} & .587{\tiny$\pm$.017} & .638{\tiny$\pm$.015} \\
ATKT & .602{\tiny$\pm$.037} & .653{\tiny$\pm$.015} & .643{\tiny$\pm$.023} & .751{\tiny$\pm$.022} & .712{\tiny$\pm$.021} & .559{\tiny$\pm$.047} & .614{\tiny$\pm$.008} & .648{\tiny$\pm$.011} \\
ATKT-fix & .545{\tiny$\pm$.022} & .558{\tiny$\pm$.015} & .541{\tiny$\pm$.044} & .615{\tiny$\pm$.034} & .600{\tiny$\pm$.032} & .516{\tiny$\pm$.010} & .520{\tiny$\pm$.029} & .556{\tiny$\pm$.011} \\
csKT & .692{\tiny$\pm$.019} & .672{\tiny$\pm$.026} & .700{\tiny$\pm$.007} & .748{\tiny$\pm$.003} & .706{\tiny$\pm$.028} & .674{\tiny$\pm$.018} & .626{\tiny$\pm$.015} & .688{\tiny$\pm$.008} \\
Deep-IRT & .584{\tiny$\pm$.025} & .625{\tiny$\pm$.039} & .571{\tiny$\pm$.027} & .707{\tiny$\pm$.017} & .705{\tiny$\pm$.015} & .546{\tiny$\pm$.024} & .590{\tiny$\pm$.014} & .618{\tiny$\pm$.009} \\
DIMKT & .667{\tiny$\pm$.025} & .678{\tiny$\pm$.019} & .676{\tiny$\pm$.008} & .766{\tiny$\pm$.022} & .674{\tiny$\pm$.019} & .643{\tiny$\pm$.019} & .585{\tiny$\pm$.007} & .670{\tiny$\pm$.010} \\
DKT & .602{\tiny$\pm$.038} & .651{\tiny$\pm$.016} & .589{\tiny$\pm$.012} & .774{\tiny$\pm$.015} & .716{\tiny$\pm$.007} & .567{\tiny$\pm$.026} & .601{\tiny$\pm$.009} & .643{\tiny$\pm$.012} \\
DKT+ & .604{\tiny$\pm$.038} & .652{\tiny$\pm$.014} & .591{\tiny$\pm$.013} & .775{\tiny$\pm$.015} & \underline{.718}{\tiny$\pm$.008} & .564{\tiny$\pm$.021} & .603{\tiny$\pm$.009} & .644{\tiny$\pm$.011} \\
DKVMN & .570{\tiny$\pm$.020} & .632{\tiny$\pm$.043} & .592{\tiny$\pm$.016} & .713{\tiny$\pm$.014} & .691{\tiny$\pm$.036} & .556{\tiny$\pm$.025} & .584{\tiny$\pm$.026} & .620{\tiny$\pm$.015} \\
DTransformer & .564{\tiny$\pm$.065} & .615{\tiny$\pm$.053} & .634{\tiny$\pm$.033} & .741{\tiny$\pm$.009} & .705{\tiny$\pm$.014} & .591{\tiny$\pm$.025} & .588{\tiny$\pm$.011} & .634{\tiny$\pm$.015} \\
extraKT & .688{\tiny$\pm$.016} & .651{\tiny$\pm$.008} & .677{\tiny$\pm$.026} & .720{\tiny$\pm$.010} & .702{\tiny$\pm$.019} & .691{\tiny$\pm$.008} & .616{\tiny$\pm$.008} & .678{\tiny$\pm$.006} \\
FoLiBiKT & .639{\tiny$\pm$.054} & .669{\tiny$\pm$.010} & .630{\tiny$\pm$.071} & .712{\tiny$\pm$.034} & .689{\tiny$\pm$.015} & .630{\tiny$\pm$.039} & .612{\tiny$\pm$.021} & .655{\tiny$\pm$.015} \\
IEKT & .566{\tiny$\pm$.034} & .696{\tiny$\pm$.019} & .632{\tiny$\pm$.033} & .803{\tiny$\pm$.013} & .689{\tiny$\pm$.040} & .604{\tiny$\pm$.027} & .602{\tiny$\pm$.021} & .656{\tiny$\pm$.011} \\
KQN & .535{\tiny$\pm$.035} & .626{\tiny$\pm$.058} & .611{\tiny$\pm$.009} & .703{\tiny$\pm$.079} & .666{\tiny$\pm$.069} & .538{\tiny$\pm$.031} & .610{\tiny$\pm$.013} & .613{\tiny$\pm$.023} \\
lefoKT-AKT & .622{\tiny$\pm$.084} & .642{\tiny$\pm$.022} & .661{\tiny$\pm$.011} & .738{\tiny$\pm$.013} & .714{\tiny$\pm$.007} & .678{\tiny$\pm$.015} & .611{\tiny$\pm$.023} & .667{\tiny$\pm$.014} \\
MoC-KT & .687{\tiny$\pm$.018} & .640{\tiny$\pm$.038} & .664{\tiny$\pm$.047} & .726{\tiny$\pm$.013} & .704{\tiny$\pm$.004} & .648{\tiny$\pm$.058} & .614{\tiny$\pm$.012} & .669{\tiny$\pm$.010} \\
qDKT & .498{\tiny$\pm$.012} & .656{\tiny$\pm$.015} & .568{\tiny$\pm$.013} & .803{\tiny$\pm$.010} & -- & .537{\tiny$\pm$.016} & .568{\tiny$\pm$.013} & -- \\
QIKT & .582{\tiny$\pm$.021} & .692{\tiny$\pm$.019} & .592{\tiny$\pm$.009} & \underline{.808}{\tiny$\pm$.008} & -- & .576{\tiny$\pm$.021} & .605{\tiny$\pm$.013} & -- \\
reKT & .525{\tiny$\pm$.059} & .562{\tiny$\pm$.048} & .648{\tiny$\pm$.048} & .752{\tiny$\pm$.040} & .697{\tiny$\pm$.038} & .564{\tiny$\pm$.077} & .575{\tiny$\pm$.068} & .618{\tiny$\pm$.020} \\
robustKT & .692{\tiny$\pm$.048} & .665{\tiny$\pm$.016} & .692{\tiny$\pm$.006} & .733{\tiny$\pm$.006} & .687{\tiny$\pm$.030} & .681{\tiny$\pm$.015} & .620{\tiny$\pm$.004} & .682{\tiny$\pm$.007} \\
SAINT & .631{\tiny$\pm$.031} & .630{\tiny$\pm$.029} & .676{\tiny$\pm$.011} & .622{\tiny$\pm$.044} & .561{\tiny$\pm$.024} & .600{\tiny$\pm$.055} & .559{\tiny$\pm$.025} & .612{\tiny$\pm$.013} \\
SAINT+ & .611{\tiny$\pm$.030} & .602{\tiny$\pm$.033} & .642{\tiny$\pm$.023} & .679{\tiny$\pm$.034} & .569{\tiny$\pm$.016} & .600{\tiny$\pm$.020} & .559{\tiny$\pm$.025} & .609{\tiny$\pm$.010} \\
SAKT & .551{\tiny$\pm$.040} & .654{\tiny$\pm$.012} & .554{\tiny$\pm$.020} & .746{\tiny$\pm$.009} & .691{\tiny$\pm$.038} & .552{\tiny$\pm$.024} & .570{\tiny$\pm$.010} & .617{\tiny$\pm$.007} \\
simpleKT & .706{\tiny$\pm$.019} & .664{\tiny$\pm$.022} & .685{\tiny$\pm$.005} & .752{\tiny$\pm$.010} & .698{\tiny$\pm$.045} & .643{\tiny$\pm$.081} & .619{\tiny$\pm$.033} & .681{\tiny$\pm$.025} \\
sparseKT & .671{\tiny$\pm$.077} & .661{\tiny$\pm$.018} & .692{\tiny$\pm$.009} & .751{\tiny$\pm$.014} & .705{\tiny$\pm$.044} & .641{\tiny$\pm$.083} & .625{\tiny$\pm$.022} & .678{\tiny$\pm$.022} \\
stableKT & .679{\tiny$\pm$.050} & .669{\tiny$\pm$.014} & .706{\tiny$\pm$.005} & .735{\tiny$\pm$.007} & .709{\tiny$\pm$.015} & .677{\tiny$\pm$.023} & .618{\tiny$\pm$.016} & .685{\tiny$\pm$.009} \\
UKT & .686{\tiny$\pm$.036} & .663{\tiny$\pm$.005} & .683{\tiny$\pm$.013} & .691{\tiny$\pm$.087} & .697{\tiny$\pm$.022} & .680{\tiny$\pm$.008} & .606{\tiny$\pm$.010} & .672{\tiny$\pm$.012} \\
\bottomrule
\end{tabular}
\end{table*}

\begin{table*}[p]
\centering
\scriptsize
\setlength{\tabcolsep}{3pt}
\caption{P1 with $N{=}32$ training learners: AUC of every model, mean{\tiny$\pm$SD} over 5 learner-subset seeds, on the 2{,}000 shared early targets per dataset. $^\ddagger$Three datasets of the original CLST paper. $^{(k)}$ = only $k$ seeds finished. $^\P$Best of four pyKT models $\times$ eight hyper-parameter configurations chosen on the validation learners of each cell (no test information). Mean = mean over the 7 datasets (-- if a dataset is missing). qDKT and QIKT do not fit Algebra05 in 24\,GB of GPU memory (--). $^\circ$The method uses no data from the $N$ learners here (\JevZ{} and Thinking-KT everywhere; LOKT on datasets without answer options), so its $N{=}0$ run is repeated. Best in \textbf{bold}, second best \underline{underlined} (among cells with $\ge$3 seeds).}
\label{tab:p1n32}
\begin{tabular}{l cccccccc}
\toprule
\textbf{Method} & \textbf{A09} & \textbf{NIPS34} & \textbf{XES3G5M} & \textbf{DBE} & \textbf{Alg05} & \textbf{Junyi} & \textbf{A17} & \textbf{Mean} \\
\midrule
\multicolumn{9}{l}{\textit{System-One LLM}} \\
\rowcolor{focusbg} \textbf{JevKT} & \underline{.737}{\tiny$\pm$.006} & \textbf{.752}{\tiny$\pm$.005} & \textbf{.728}{\tiny$\pm$.007} & \textbf{.838}{\tiny$\pm$.004} & \textbf{.751}{\tiny$\pm$.002} & \textbf{.709}{\tiny$\pm$.003} & \textbf{.656}{\tiny$\pm$.009} & \textbf{.739}{\tiny$\pm$.002} \\
\JevZ{} & \textbf{.738}{\tiny$\pm$.014}$^\circ$ & .691{\tiny$\pm$.014}$^\circ$ & \underline{.725}{\tiny$\pm$.013}$^\circ$ & .742{\tiny$\pm$.018}$^\circ$ & .698{\tiny$\pm$.018}$^\circ$ & \underline{.700}{\tiny$\pm$.015}$^\circ$ & .648{\tiny$\pm$.013}$^\circ$ & .706{\tiny$\pm$.006}$^\circ$ \\
Laya (open-source Jev) & .602{\tiny$\pm$.004} & .558{\tiny$\pm$.006} & .522{\tiny$\pm$.007} & .629{\tiny$\pm$.020} & .523{\tiny$\pm$.005} & .498{\tiny$\pm$.005} & .517{\tiny$\pm$.002} & .550{\tiny$\pm$.003} \\
\midrule
\multicolumn{9}{l}{\textit{System-Two LLM-KT, prompted}} \\
Thinking-KT & .700{\tiny$\pm$.014}$^\circ$ & .641{\tiny$\pm$.014}$^\circ$ & .663{\tiny$\pm$.014}$^\circ$ & .666{\tiny$\pm$.019}$^\circ$ & .663{\tiny$\pm$.018}$^\circ$ & .642{\tiny$\pm$.016}$^\circ$ & .576{\tiny$\pm$.013}$^\circ$ & .650{\tiny$\pm$.006}$^\circ$ \\
LOKT & .707{\tiny$\pm$.005} & .660{\tiny$\pm$.007} & .699{\tiny$\pm$.014}$^\circ$ & .701{\tiny$\pm$.006} & .656{\tiny$\pm$.019}$^\circ$ & .671{\tiny$\pm$.016}$^\circ$ & .608{\tiny$\pm$.013}$^\circ$ & .672 \\
\midrule
\multicolumn{9}{l}{\textit{System-Two LLM-KT, trained}} \\
CLST$^\ddagger$ & .686{\tiny$\pm$.100} & .649{\tiny$\pm$.065} & -- & -- & .658{\tiny$\pm$.067} & -- & -- & -- \\
\midrule
\multicolumn{9}{l}{\textit{Classical KT}} \\
BKT & .673{\tiny$\pm$.024} & .639{\tiny$\pm$.004} & .603{\tiny$\pm$.013} & .732{\tiny$\pm$.009} & .719{\tiny$\pm$.005} & .625{\tiny$\pm$.023} & .627{\tiny$\pm$.005} & .660{\tiny$\pm$.005} \\
PFA & .641{\tiny$\pm$.017} & .638{\tiny$\pm$.010} & .592{\tiny$\pm$.016} & .788{\tiny$\pm$.011} & .665{\tiny$\pm$.008} & .590{\tiny$\pm$.023} & .616{\tiny$\pm$.006} & .647{\tiny$\pm$.006} \\
\midrule
\multicolumn{9}{l}{\textit{Deep KT (pyKT)}} \\
Tuned deep KT$^\P$ & .687{\tiny$\pm$.031} & .700{\tiny$\pm$.021} & .692{\tiny$\pm$.022} & .795{\tiny$\pm$.012} & \underline{.749}{\tiny$\pm$.017} & .687{\tiny$\pm$.004} & \underline{.649}{\tiny$\pm$.020} & \underline{.709}{\tiny$\pm$.005} \\
AKT & .710{\tiny$\pm$.017} & .671{\tiny$\pm$.007} & .695{\tiny$\pm$.009} & .717{\tiny$\pm$.006} & .733{\tiny$\pm$.027} & .687{\tiny$\pm$.006} & .617{\tiny$\pm$.005} & .690{\tiny$\pm$.005} \\
AT-DKT & .627{\tiny$\pm$.044} & .673{\tiny$\pm$.008} & .622{\tiny$\pm$.015} & .785{\tiny$\pm$.004} & .734{\tiny$\pm$.020} & .595{\tiny$\pm$.026} & .610{\tiny$\pm$.016} & .664{\tiny$\pm$.011} \\
ATKT & .655{\tiny$\pm$.036} & .681{\tiny$\pm$.006} & .674{\tiny$\pm$.005} & .785{\tiny$\pm$.011} & .738{\tiny$\pm$.020} & .598{\tiny$\pm$.023} & .617{\tiny$\pm$.018} & .678{\tiny$\pm$.009} \\
ATKT-fix & .555{\tiny$\pm$.033} & .575{\tiny$\pm$.019} & .597{\tiny$\pm$.014} & .620{\tiny$\pm$.055} & .622{\tiny$\pm$.037} & .535{\tiny$\pm$.017} & .532{\tiny$\pm$.023} & .576{\tiny$\pm$.014} \\
csKT & .724{\tiny$\pm$.011} & .685{\tiny$\pm$.004} & .703{\tiny$\pm$.006} & .753{\tiny$\pm$.017} & .722{\tiny$\pm$.021} & .693{\tiny$\pm$.005} & .634{\tiny$\pm$.004} & .702{\tiny$\pm$.006} \\
Deep-IRT & .638{\tiny$\pm$.017} & .663{\tiny$\pm$.011} & .594{\tiny$\pm$.031} & .730{\tiny$\pm$.023} & .716{\tiny$\pm$.017} & .555{\tiny$\pm$.027} & .618{\tiny$\pm$.019} & .645{\tiny$\pm$.007} \\
DIMKT & .681{\tiny$\pm$.015} & .703{\tiny$\pm$.006} & .681{\tiny$\pm$.008} & .784{\tiny$\pm$.015} & .722{\tiny$\pm$.014} & .648{\tiny$\pm$.021} & .591{\tiny$\pm$.018} & .687{\tiny$\pm$.006} \\
DKT & .636{\tiny$\pm$.064} & .674{\tiny$\pm$.010} & .622{\tiny$\pm$.010} & .791{\tiny$\pm$.015} & .738{\tiny$\pm$.011} & .600{\tiny$\pm$.024} & .616{\tiny$\pm$.011} & .668{\tiny$\pm$.015} \\
DKT+ & .638{\tiny$\pm$.064} & .675{\tiny$\pm$.011} & .628{\tiny$\pm$.010} & .796{\tiny$\pm$.015} & .744{\tiny$\pm$.011} & .597{\tiny$\pm$.025} & .619{\tiny$\pm$.009} & .671{\tiny$\pm$.014} \\
DKVMN & .641{\tiny$\pm$.029} & .667{\tiny$\pm$.005} & .606{\tiny$\pm$.017} & .743{\tiny$\pm$.030} & .727{\tiny$\pm$.016} & .564{\tiny$\pm$.016} & .623{\tiny$\pm$.013} & .653{\tiny$\pm$.004} \\
DTransformer & .652{\tiny$\pm$.016} & .660{\tiny$\pm$.007} & .657{\tiny$\pm$.015} & .762{\tiny$\pm$.016} & .729{\tiny$\pm$.011} & .620{\tiny$\pm$.021} & .615{\tiny$\pm$.009} & .671{\tiny$\pm$.004} \\
extraKT & .705{\tiny$\pm$.011} & .664{\tiny$\pm$.006} & .690{\tiny$\pm$.014} & .750{\tiny$\pm$.009} & .712{\tiny$\pm$.004} & .697{\tiny$\pm$.013} & .627{\tiny$\pm$.004} & .692{\tiny$\pm$.005} \\
FoLiBiKT & .711{\tiny$\pm$.009} & .677{\tiny$\pm$.005} & .686{\tiny$\pm$.009} & .736{\tiny$\pm$.019} & .710{\tiny$\pm$.020} & .667{\tiny$\pm$.006} & .628{\tiny$\pm$.010} & .688{\tiny$\pm$.005} \\
IEKT & .656{\tiny$\pm$.031} & \underline{.732}{\tiny$\pm$.016} & .686{\tiny$\pm$.018} & .804{\tiny$\pm$.027} & .734{\tiny$\pm$.007} & .613{\tiny$\pm$.026} & .638{\tiny$\pm$.008} & .695{\tiny$\pm$.008} \\
KQN & .582{\tiny$\pm$.062} & .676{\tiny$\pm$.004} & .638{\tiny$\pm$.017} & .767{\tiny$\pm$.013} & .707{\tiny$\pm$.010} & .563{\tiny$\pm$.021} & .619{\tiny$\pm$.007} & .650{\tiny$\pm$.012} \\
lefoKT-AKT & .683{\tiny$\pm$.058} & .662{\tiny$\pm$.005} & .666{\tiny$\pm$.015} & .749{\tiny$\pm$.005} & .716{\tiny$\pm$.003} & .691{\tiny$\pm$.008} & .619{\tiny$\pm$.010} & .684{\tiny$\pm$.009} \\
MoC-KT & .682{\tiny$\pm$.044} & .660{\tiny$\pm$.016} & .674{\tiny$\pm$.036} & .727{\tiny$\pm$.018} & .705{\tiny$\pm$.004} & .691{\tiny$\pm$.005} & .621{\tiny$\pm$.010} & .680{\tiny$\pm$.007} \\
qDKT & .502{\tiny$\pm$.010} & .675{\tiny$\pm$.008} & .591{\tiny$\pm$.007} & .806{\tiny$\pm$.016} & -- & .566{\tiny$\pm$.011} & .604{\tiny$\pm$.030} & -- \\
QIKT & .621{\tiny$\pm$.027} & .729{\tiny$\pm$.013} & .610{\tiny$\pm$.010} & \underline{.816}{\tiny$\pm$.010} & -- & .604{\tiny$\pm$.032} & .639{\tiny$\pm$.021} & -- \\
reKT & .613{\tiny$\pm$.092} & .560{\tiny$\pm$.074} & .633{\tiny$\pm$.079} & .741{\tiny$\pm$.094} & .711{\tiny$\pm$.042} & .643{\tiny$\pm$.087} & .546{\tiny$\pm$.055} & .635{\tiny$\pm$.026} \\
robustKT & .713{\tiny$\pm$.010} & .669{\tiny$\pm$.007} & .659{\tiny$\pm$.069} & .737{\tiny$\pm$.006} & .742{\tiny$\pm$.016} & .686{\tiny$\pm$.010} & .619{\tiny$\pm$.006} & .689{\tiny$\pm$.012} \\
SAINT & .660{\tiny$\pm$.035} & .635{\tiny$\pm$.025} & .682{\tiny$\pm$.009} & .652{\tiny$\pm$.042} & .587{\tiny$\pm$.030} & .607{\tiny$\pm$.058} & .582{\tiny$\pm$.008} & .629{\tiny$\pm$.019} \\
SAINT+ & .645{\tiny$\pm$.026} & .637{\tiny$\pm$.030} & .641{\tiny$\pm$.018} & .686{\tiny$\pm$.028} & .583{\tiny$\pm$.043} & .583{\tiny$\pm$.047} & .585{\tiny$\pm$.014} & .623{\tiny$\pm$.020} \\
SAKT & .588{\tiny$\pm$.047} & .665{\tiny$\pm$.005} & .568{\tiny$\pm$.045} & .765{\tiny$\pm$.011} & .714{\tiny$\pm$.021} & .545{\tiny$\pm$.028} & .589{\tiny$\pm$.011} & .634{\tiny$\pm$.010} \\
simpleKT & .678{\tiny$\pm$.082} & .679{\tiny$\pm$.002} & .689{\tiny$\pm$.009} & .757{\tiny$\pm$.007} & .721{\tiny$\pm$.004} & .687{\tiny$\pm$.009} & .637{\tiny$\pm$.008} & .692{\tiny$\pm$.013} \\
sparseKT & .706{\tiny$\pm$.015} & .671{\tiny$\pm$.006} & .697{\tiny$\pm$.002} & .765{\tiny$\pm$.012} & .721{\tiny$\pm$.004} & .691{\tiny$\pm$.011} & .634{\tiny$\pm$.007} & .698{\tiny$\pm$.002} \\
stableKT & .695{\tiny$\pm$.061} & .688{\tiny$\pm$.003} & .708{\tiny$\pm$.003} & .741{\tiny$\pm$.009} & .740{\tiny$\pm$.008} & .689{\tiny$\pm$.009} & .629{\tiny$\pm$.006} & .699{\tiny$\pm$.009} \\
UKT & .691{\tiny$\pm$.026} & .670{\tiny$\pm$.008} & .681{\tiny$\pm$.015} & .734{\tiny$\pm$.011} & .688{\tiny$\pm$.025} & .681{\tiny$\pm$.010} & .613{\tiny$\pm$.008} & .680{\tiny$\pm$.007} \\
\bottomrule
\end{tabular}
\end{table*}

\begin{table*}[p]
\centering
\scriptsize
\setlength{\tabcolsep}{3pt}
\caption{P1 with $N{=}64$ training learners: AUC of every model, mean{\tiny$\pm$SD} over 5 learner-subset seeds, on the 2{,}000 shared early targets per dataset. $^\ddagger$Three datasets of the original CLST paper. $^{(k)}$ = only $k$ seeds finished. $^\P$Best of four pyKT models $\times$ eight hyper-parameter configurations chosen on the validation learners of each cell (no test information). Mean = mean over the 7 datasets (-- if a dataset is missing). qDKT and QIKT do not fit Algebra05 in 24\,GB of GPU memory (--). $^\circ$The method uses no data from the $N$ learners here (\JevZ{} and Thinking-KT everywhere; LOKT on datasets without answer options), so its $N{=}0$ run is repeated. Best in \textbf{bold}, second best \underline{underlined} (among cells with $\ge$3 seeds).}
\label{tab:p1n64}
\begin{tabular}{l cccccccc}
\toprule
\textbf{Method} & \textbf{A09} & \textbf{NIPS34} & \textbf{XES3G5M} & \textbf{DBE} & \textbf{Alg05} & \textbf{Junyi} & \textbf{A17} & \textbf{Mean} \\
\midrule
\multicolumn{9}{l}{\textit{System-One LLM}} \\
\rowcolor{focusbg} \textbf{JevKT} & \textbf{.745}{\tiny$\pm$.005} & \textbf{.762}{\tiny$\pm$.006} & \textbf{.739}{\tiny$\pm$.007} & \underline{.838}{\tiny$\pm$.002} & .762{\tiny$\pm$.004} & \textbf{.715}{\tiny$\pm$.005} & .667{\tiny$\pm$.009} & \textbf{.747}{\tiny$\pm$.002} \\
\JevZ{} & \underline{.738}{\tiny$\pm$.014}$^\circ$ & .691{\tiny$\pm$.014}$^\circ$ & \underline{.725}{\tiny$\pm$.013}$^\circ$ & .742{\tiny$\pm$.018}$^\circ$ & .698{\tiny$\pm$.018}$^\circ$ & .700{\tiny$\pm$.015}$^\circ$ & .648{\tiny$\pm$.013}$^\circ$ & .706{\tiny$\pm$.006}$^\circ$ \\
Laya (open-source Jev) & .600{\tiny$\pm$.003} & .571{\tiny$\pm$.004} & .515{\tiny$\pm$.008} & .607{\tiny$\pm$.007} & .533{\tiny$\pm$.004} & .509{\tiny$\pm$.004} & .506{\tiny$\pm$.006} & .549{\tiny$\pm$.002} \\
\midrule
\multicolumn{9}{l}{\textit{System-Two LLM-KT, prompted}} \\
Thinking-KT & .700{\tiny$\pm$.014}$^\circ$ & .641{\tiny$\pm$.014}$^\circ$ & .663{\tiny$\pm$.014}$^\circ$ & .666{\tiny$\pm$.019}$^\circ$ & .663{\tiny$\pm$.018}$^\circ$ & .642{\tiny$\pm$.016}$^\circ$ & .576{\tiny$\pm$.013}$^\circ$ & .650{\tiny$\pm$.006}$^\circ$ \\
LOKT & .703{\tiny$\pm$.004} & .660{\tiny$\pm$.005} & .699{\tiny$\pm$.014}$^\circ$ & .699{\tiny$\pm$.011} & .656{\tiny$\pm$.019}$^\circ$ & .671{\tiny$\pm$.016}$^\circ$ & .608{\tiny$\pm$.013}$^\circ$ & .671 \\
\midrule
\multicolumn{9}{l}{\textit{System-Two LLM-KT, trained}} \\
CLST$^\ddagger$ & .735{\tiny$\pm$.011} & .673{\tiny$\pm$.028} & -- & -- & .717{\tiny$\pm$.016} & -- & -- & -- \\
\midrule
\multicolumn{9}{l}{\textit{Classical KT}} \\
BKT & .711{\tiny$\pm$.010} & .651{\tiny$\pm$.004} & .627{\tiny$\pm$.007} & .742{\tiny$\pm$.006} & .726{\tiny$\pm$.004} & .635{\tiny$\pm$.012} & .636{\tiny$\pm$.008} & .675{\tiny$\pm$.003} \\
PFA & .675{\tiny$\pm$.017} & .648{\tiny$\pm$.001} & .615{\tiny$\pm$.003} & .802{\tiny$\pm$.006} & .667{\tiny$\pm$.005} & .619{\tiny$\pm$.028} & .627{\tiny$\pm$.006} & .665{\tiny$\pm$.004} \\
\midrule
\multicolumn{9}{l}{\textit{Deep KT (pyKT)}} \\
Tuned deep KT$^\P$ & .731{\tiny$\pm$.004} & .738{\tiny$\pm$.028} & .711{\tiny$\pm$.008} & .827{\tiny$\pm$.006} & .770{\tiny$\pm$.007} & .701{\tiny$\pm$.003} & \textbf{.681}{\tiny$\pm$.012} & \underline{.737}{\tiny$\pm$.003} \\
AKT & .723{\tiny$\pm$.006} & .679{\tiny$\pm$.005} & .706{\tiny$\pm$.006} & .742{\tiny$\pm$.035} & \underline{.774}{\tiny$\pm$.004} & .697{\tiny$\pm$.007} & .619{\tiny$\pm$.004} & .706{\tiny$\pm$.007} \\
AT-DKT & .680{\tiny$\pm$.028} & .685{\tiny$\pm$.004} & .664{\tiny$\pm$.012} & .811{\tiny$\pm$.006} & .771{\tiny$\pm$.009} & .633{\tiny$\pm$.004} & .620{\tiny$\pm$.012} & .695{\tiny$\pm$.007} \\
ATKT & .713{\tiny$\pm$.014} & .690{\tiny$\pm$.004} & .701{\tiny$\pm$.005} & .805{\tiny$\pm$.005} & .754{\tiny$\pm$.011} & .652{\tiny$\pm$.004} & .638{\tiny$\pm$.015} & .708{\tiny$\pm$.002} \\
ATKT-fix & .577{\tiny$\pm$.023} & .616{\tiny$\pm$.020} & .629{\tiny$\pm$.013} & .627{\tiny$\pm$.036} & .740{\tiny$\pm$.007} & .565{\tiny$\pm$.031} & .555{\tiny$\pm$.018} & .616{\tiny$\pm$.007} \\
csKT & .734{\tiny$\pm$.005} & .724{\tiny$\pm$.034} & .712{\tiny$\pm$.007} & .749{\tiny$\pm$.003} & .758{\tiny$\pm$.005} & .700{\tiny$\pm$.005} & .635{\tiny$\pm$.005} & .716{\tiny$\pm$.005} \\
Deep-IRT & .670{\tiny$\pm$.025} & .675{\tiny$\pm$.007} & .653{\tiny$\pm$.010} & .765{\tiny$\pm$.023} & .743{\tiny$\pm$.008} & .573{\tiny$\pm$.027} & .640{\tiny$\pm$.009} & .674{\tiny$\pm$.007} \\
DIMKT & .688{\tiny$\pm$.016} & .726{\tiny$\pm$.006} & .693{\tiny$\pm$.004} & .816{\tiny$\pm$.006} & .746{\tiny$\pm$.007} & .676{\tiny$\pm$.013} & .642{\tiny$\pm$.022} & .712{\tiny$\pm$.006} \\
DKT & .680{\tiny$\pm$.024} & .687{\tiny$\pm$.005} & .661{\tiny$\pm$.012} & .813{\tiny$\pm$.003} & .771{\tiny$\pm$.005} & .637{\tiny$\pm$.008} & .629{\tiny$\pm$.017} & .697{\tiny$\pm$.006} \\
DKT+ & .683{\tiny$\pm$.024} & .687{\tiny$\pm$.005} & .668{\tiny$\pm$.012} & .817{\tiny$\pm$.001} & .769{\tiny$\pm$.005} & .638{\tiny$\pm$.007} & .632{\tiny$\pm$.018} & .699{\tiny$\pm$.005} \\
DKVMN & .657{\tiny$\pm$.036} & .676{\tiny$\pm$.007} & .658{\tiny$\pm$.006} & .771{\tiny$\pm$.013} & .740{\tiny$\pm$.009} & .580{\tiny$\pm$.010} & .639{\tiny$\pm$.009} & .674{\tiny$\pm$.006} \\
DTransformer & .680{\tiny$\pm$.004} & .683{\tiny$\pm$.015} & .683{\tiny$\pm$.015} & .787{\tiny$\pm$.015} & .751{\tiny$\pm$.015} & .650{\tiny$\pm$.008} & .633{\tiny$\pm$.005} & .695{\tiny$\pm$.002} \\
extraKT & .724{\tiny$\pm$.003} & .673{\tiny$\pm$.002} & .701{\tiny$\pm$.011} & .752{\tiny$\pm$.006} & .726{\tiny$\pm$.020} & \underline{.704}{\tiny$\pm$.008} & .629{\tiny$\pm$.004} & .701{\tiny$\pm$.004} \\
FoLiBiKT & .711{\tiny$\pm$.018} & .690{\tiny$\pm$.023} & .701{\tiny$\pm$.007} & .759{\tiny$\pm$.038} & .754{\tiny$\pm$.008} & .682{\tiny$\pm$.008} & .631{\tiny$\pm$.009} & .704{\tiny$\pm$.008} \\
IEKT & .703{\tiny$\pm$.008} & \underline{.750}{\tiny$\pm$.011} & .720{\tiny$\pm$.005} & \textbf{.846}{\tiny$\pm$.006} & .749{\tiny$\pm$.014} & .664{\tiny$\pm$.034} & .657{\tiny$\pm$.014} & .727{\tiny$\pm$.002} \\
KQN & .672{\tiny$\pm$.020} & .681{\tiny$\pm$.003} & .680{\tiny$\pm$.017} & .773{\tiny$\pm$.010} & .722{\tiny$\pm$.012} & .595{\tiny$\pm$.029} & .629{\tiny$\pm$.004} & .679{\tiny$\pm$.008} \\
lefoKT-AKT & .729{\tiny$\pm$.004} & .672{\tiny$\pm$.011} & .675{\tiny$\pm$.010} & .766{\tiny$\pm$.015} & .714{\tiny$\pm$.003} & .695{\tiny$\pm$.005} & .615{\tiny$\pm$.016} & .695{\tiny$\pm$.004} \\
MoC-KT & .728{\tiny$\pm$.008} & .672{\tiny$\pm$.007} & .694{\tiny$\pm$.006} & .699{\tiny$\pm$.074} & .705{\tiny$\pm$.005} & .691{\tiny$\pm$.010} & .621{\tiny$\pm$.009} & .687{\tiny$\pm$.011} \\
qDKT & .513{\tiny$\pm$.013} & .695{\tiny$\pm$.005} & .613{\tiny$\pm$.012} & .829{\tiny$\pm$.009} & -- & .600{\tiny$\pm$.013} & .637{\tiny$\pm$.010} & -- \\
QIKT & .650{\tiny$\pm$.022} & .743{\tiny$\pm$.009} & .650{\tiny$\pm$.010} & .832{\tiny$\pm$.007} & -- & .640{\tiny$\pm$.007} & \underline{.668}{\tiny$\pm$.010} & -- \\
reKT & .661{\tiny$\pm$.071} & .545{\tiny$\pm$.038} & .675{\tiny$\pm$.078} & .775{\tiny$\pm$.051} & .736{\tiny$\pm$.045} & .672{\tiny$\pm$.036} & .585{\tiny$\pm$.052} & .664{\tiny$\pm$.029} \\
robustKT & .723{\tiny$\pm$.007} & .677{\tiny$\pm$.009} & .701{\tiny$\pm$.006} & .766{\tiny$\pm$.037} & \textbf{.777}{\tiny$\pm$.011} & .692{\tiny$\pm$.013} & .622{\tiny$\pm$.006} & .708{\tiny$\pm$.006} \\
SAINT & .694{\tiny$\pm$.008} & .650{\tiny$\pm$.028} & .685{\tiny$\pm$.007} & .655{\tiny$\pm$.046} & .614{\tiny$\pm$.014} & .631{\tiny$\pm$.024} & .588{\tiny$\pm$.028} & .645{\tiny$\pm$.009} \\
SAINT+ & .661{\tiny$\pm$.015} & .628{\tiny$\pm$.026} & .663{\tiny$\pm$.016} & .703{\tiny$\pm$.039} & .600{\tiny$\pm$.071} & .610{\tiny$\pm$.030} & .569{\tiny$\pm$.039} & .633{\tiny$\pm$.012} \\
SAKT & .629{\tiny$\pm$.038} & .675{\tiny$\pm$.005} & .627{\tiny$\pm$.009} & .788{\tiny$\pm$.011} & .731{\tiny$\pm$.006} & .584{\tiny$\pm$.019} & .614{\tiny$\pm$.010} & .664{\tiny$\pm$.004} \\
simpleKT & .732{\tiny$\pm$.010} & .687{\tiny$\pm$.003} & .701{\tiny$\pm$.003} & .769{\tiny$\pm$.007} & .735{\tiny$\pm$.011} & .701{\tiny$\pm$.005} & .643{\tiny$\pm$.007} & .710{\tiny$\pm$.002} \\
sparseKT & .724{\tiny$\pm$.009} & .676{\tiny$\pm$.006} & .704{\tiny$\pm$.003} & .783{\tiny$\pm$.022} & .757{\tiny$\pm$.007} & .698{\tiny$\pm$.006} & .641{\tiny$\pm$.007} & .712{\tiny$\pm$.003} \\
stableKT & .728{\tiny$\pm$.007} & .695{\tiny$\pm$.005} & .715{\tiny$\pm$.001} & .754{\tiny$\pm$.023} & .761{\tiny$\pm$.006} & .700{\tiny$\pm$.007} & .632{\tiny$\pm$.006} & .712{\tiny$\pm$.003} \\
UKT & .719{\tiny$\pm$.012} & .675{\tiny$\pm$.013} & .682{\tiny$\pm$.039} & .735{\tiny$\pm$.010} & .708{\tiny$\pm$.033} & .690{\tiny$\pm$.016} & .614{\tiny$\pm$.008} & .689{\tiny$\pm$.006} \\
\bottomrule
\end{tabular}
\end{table*}

\begin{table*}[p]
\centering
\scriptsize
\setlength{\tabcolsep}{3pt}
\caption{P1 with $N{=}128$ training learners: AUC of every model, mean{\tiny$\pm$SD} over 5 learner-subset seeds, on the 2{,}000 shared early targets per dataset. $^\ddagger$Three datasets of the original CLST paper. $^{(k)}$ = only $k$ seeds finished. Mean = mean over the 7 datasets (-- if a dataset is missing). qDKT and QIKT do not fit Algebra05 in 24\,GB of GPU memory (--). $^\circ$The method uses no data from the $N$ learners here (\JevZ{} and Thinking-KT everywhere; LOKT on datasets without answer options), so its $N{=}0$ run is repeated. Best in \textbf{bold}, second best \underline{underlined} (among cells with $\ge$3 seeds).}
\label{tab:p1n128}
\begin{tabular}{l cccccccc}
\toprule
\textbf{Method} & \textbf{A09} & \textbf{NIPS34} & \textbf{XES3G5M} & \textbf{DBE} & \textbf{Alg05} & \textbf{Junyi} & \textbf{A17} & \textbf{Mean} \\
\midrule
\multicolumn{9}{l}{\textit{System-One LLM}} \\
\rowcolor{focusbg} \textbf{JevKT} & \textbf{.745}{\tiny$\pm$.002} & \underline{.764}{\tiny$\pm$.005} & \textbf{.747}{\tiny$\pm$.006} & .837{\tiny$\pm$.004} & .774{\tiny$\pm$.006} & \textbf{.727}{\tiny$\pm$.005} & \underline{.680}{\tiny$\pm$.003} & \textbf{.753}{\tiny$\pm$.001} \\
\JevZ{} & .738{\tiny$\pm$.014}$^\circ$ & .691{\tiny$\pm$.014}$^\circ$ & .725{\tiny$\pm$.013}$^\circ$ & .742{\tiny$\pm$.018}$^\circ$ & .698{\tiny$\pm$.018}$^\circ$ & .700{\tiny$\pm$.015}$^\circ$ & .648{\tiny$\pm$.013}$^\circ$ & .706{\tiny$\pm$.006}$^\circ$ \\
Laya (open-source Jev) & .596{\tiny$\pm$.003} & .575{\tiny$\pm$.002} & .511{\tiny$\pm$.007} & .605{\tiny$\pm$.003} & .530{\tiny$\pm$.004} & .520{\tiny$\pm$.006} & .507{\tiny$\pm$.004} & .549{\tiny$\pm$.002} \\
\midrule
\multicolumn{9}{l}{\textit{System-Two LLM-KT, prompted}} \\
Thinking-KT & .700{\tiny$\pm$.014}$^\circ$ & .641{\tiny$\pm$.014}$^\circ$ & .663{\tiny$\pm$.014}$^\circ$ & .666{\tiny$\pm$.019}$^\circ$ & .663{\tiny$\pm$.018}$^\circ$ & .642{\tiny$\pm$.016}$^\circ$ & .576{\tiny$\pm$.013}$^\circ$ & .650{\tiny$\pm$.006}$^\circ$ \\
LOKT & .704{\tiny$\pm$.004} & .658{\tiny$\pm$.005} & .699{\tiny$\pm$.014}$^\circ$ & .697{\tiny$\pm$.008} & .656{\tiny$\pm$.019}$^\circ$ & .671{\tiny$\pm$.016}$^\circ$ & .608{\tiny$\pm$.013}$^\circ$ & .670 \\
\midrule
\multicolumn{9}{l}{\textit{Classical KT}} \\
BKT & .730{\tiny$\pm$.006} & .650{\tiny$\pm$.004} & .646{\tiny$\pm$.007} & .748{\tiny$\pm$.003} & .730{\tiny$\pm$.003} & .655{\tiny$\pm$.010} & .639{\tiny$\pm$.003} & .685{\tiny$\pm$.001} \\
PFA & .701{\tiny$\pm$.004} & .648{\tiny$\pm$.003} & .634{\tiny$\pm$.011} & .814{\tiny$\pm$.002} & .663{\tiny$\pm$.002} & .649{\tiny$\pm$.015} & .631{\tiny$\pm$.007} & .677{\tiny$\pm$.003} \\
\midrule
\multicolumn{9}{l}{\textit{Deep KT (pyKT)}} \\
AKT & .720{\tiny$\pm$.011} & .681{\tiny$\pm$.009} & .709{\tiny$\pm$.005} & .803{\tiny$\pm$.040} & .785{\tiny$\pm$.005} & .697{\tiny$\pm$.008} & .621{\tiny$\pm$.003} & .717{\tiny$\pm$.005} \\
AT-DKT & .720{\tiny$\pm$.006} & .699{\tiny$\pm$.005} & .695{\tiny$\pm$.011} & .829{\tiny$\pm$.011} & .784{\tiny$\pm$.004} & .657{\tiny$\pm$.012} & .642{\tiny$\pm$.017} & .718{\tiny$\pm$.004} \\
ATKT & .732{\tiny$\pm$.005} & .696{\tiny$\pm$.005} & .716{\tiny$\pm$.008} & .822{\tiny$\pm$.006} & .769{\tiny$\pm$.004} & .679{\tiny$\pm$.005} & .647{\tiny$\pm$.011} & .723{\tiny$\pm$.003} \\
ATKT-fix & .600{\tiny$\pm$.036} & .621{\tiny$\pm$.014} & .647{\tiny$\pm$.015} & .736{\tiny$\pm$.053} & .749{\tiny$\pm$.002} & .597{\tiny$\pm$.015} & .598{\tiny$\pm$.008} & .650{\tiny$\pm$.012} \\
csKT & \underline{.739}{\tiny$\pm$.007} & .747{\tiny$\pm$.034} & .723{\tiny$\pm$.008} & .751{\tiny$\pm$.004} & .772{\tiny$\pm$.006} & .705{\tiny$\pm$.004} & .633{\tiny$\pm$.002} & .725{\tiny$\pm$.004} \\
Deep-IRT & .708{\tiny$\pm$.011} & .681{\tiny$\pm$.008} & .681{\tiny$\pm$.008} & .798{\tiny$\pm$.012} & .771{\tiny$\pm$.006} & .614{\tiny$\pm$.010} & .648{\tiny$\pm$.009} & .700{\tiny$\pm$.004} \\
DIMKT & .702{\tiny$\pm$.009} & .748{\tiny$\pm$.013} & .716{\tiny$\pm$.011} & .837{\tiny$\pm$.009} & .761{\tiny$\pm$.009} & .692{\tiny$\pm$.012} & .675{\tiny$\pm$.007} & .733{\tiny$\pm$.005} \\
DKT & .722{\tiny$\pm$.004} & .698{\tiny$\pm$.006} & .695{\tiny$\pm$.008} & .832{\tiny$\pm$.009} & .782{\tiny$\pm$.005} & .656{\tiny$\pm$.009} & .646{\tiny$\pm$.021} & .719{\tiny$\pm$.005} \\
DKT+ & .723{\tiny$\pm$.004} & .700{\tiny$\pm$.004} & .703{\tiny$\pm$.008} & .836{\tiny$\pm$.005} & .779{\tiny$\pm$.007} & .657{\tiny$\pm$.009} & .651{\tiny$\pm$.012} & .721{\tiny$\pm$.003} \\
DKVMN & .707{\tiny$\pm$.018} & .680{\tiny$\pm$.010} & .680{\tiny$\pm$.011} & .812{\tiny$\pm$.009} & .762{\tiny$\pm$.020} & .619{\tiny$\pm$.017} & .647{\tiny$\pm$.008} & .701{\tiny$\pm$.005} \\
DTransformer & .709{\tiny$\pm$.013} & .727{\tiny$\pm$.013} & .699{\tiny$\pm$.015} & .817{\tiny$\pm$.008} & .766{\tiny$\pm$.012} & .660{\tiny$\pm$.012} & .644{\tiny$\pm$.010} & .718{\tiny$\pm$.007} \\
extraKT & .727{\tiny$\pm$.005} & .685{\tiny$\pm$.001} & .710{\tiny$\pm$.003} & .754{\tiny$\pm$.006} & .773{\tiny$\pm$.008} & \underline{.708}{\tiny$\pm$.008} & .629{\tiny$\pm$.001} & .712{\tiny$\pm$.002} \\
FoLiBiKT & .725{\tiny$\pm$.009} & .736{\tiny$\pm$.035} & .722{\tiny$\pm$.007} & .746{\tiny$\pm$.015} & .773{\tiny$\pm$.006} & .695{\tiny$\pm$.003} & .633{\tiny$\pm$.011} & .718{\tiny$\pm$.004} \\
IEKT & .714{\tiny$\pm$.009} & \textbf{.768}{\tiny$\pm$.010} & \underline{.739}{\tiny$\pm$.002} & \textbf{.852}{\tiny$\pm$.003} & .772{\tiny$\pm$.008} & .702{\tiny$\pm$.006} & .678{\tiny$\pm$.009} & \underline{.747}{\tiny$\pm$.001} \\
KQN & .694{\tiny$\pm$.019} & .685{\tiny$\pm$.003} & .702{\tiny$\pm$.017} & .763{\tiny$\pm$.006} & .757{\tiny$\pm$.004} & .635{\tiny$\pm$.012} & .629{\tiny$\pm$.005} & .695{\tiny$\pm$.002} \\
lefoKT-AKT & .736{\tiny$\pm$.006} & .694{\tiny$\pm$.004} & .681{\tiny$\pm$.007} & .786{\tiny$\pm$.020} & .723{\tiny$\pm$.020} & .697{\tiny$\pm$.005} & .629{\tiny$\pm$.008} & .707{\tiny$\pm$.004} \\
MoC-KT & .730{\tiny$\pm$.005} & .671{\tiny$\pm$.010} & .697{\tiny$\pm$.007} & .708{\tiny$\pm$.067} & .686{\tiny$\pm$.032} & .698{\tiny$\pm$.006} & .625{\tiny$\pm$.007} & .688{\tiny$\pm$.016} \\
qDKT & .544{\tiny$\pm$.010} & .719{\tiny$\pm$.010} & .647{\tiny$\pm$.011} & .838{\tiny$\pm$.005} & -- & .630{\tiny$\pm$.013} & .665{\tiny$\pm$.013} & -- \\
QIKT & .680{\tiny$\pm$.012} & .759{\tiny$\pm$.010} & .685{\tiny$\pm$.013} & \underline{.845}{\tiny$\pm$.004} & -- & .663{\tiny$\pm$.009} & \textbf{.692}{\tiny$\pm$.004} & -- \\
reKT & .665{\tiny$\pm$.046} & .588{\tiny$\pm$.046} & .722{\tiny$\pm$.009} & .795{\tiny$\pm$.056} & .774{\tiny$\pm$.005} & .666{\tiny$\pm$.075} & .622{\tiny$\pm$.016} & .690{\tiny$\pm$.012} \\
robustKT & .724{\tiny$\pm$.007} & .683{\tiny$\pm$.008} & .710{\tiny$\pm$.004} & .757{\tiny$\pm$.039} & .789{\tiny$\pm$.007} & .698{\tiny$\pm$.009} & .623{\tiny$\pm$.005} & .712{\tiny$\pm$.006} \\
SAINT & .686{\tiny$\pm$.010} & .676{\tiny$\pm$.009} & .696{\tiny$\pm$.016} & .692{\tiny$\pm$.082} & .623{\tiny$\pm$.034} & .623{\tiny$\pm$.043} & .586{\tiny$\pm$.026} & .654{\tiny$\pm$.013} \\
SAINT+ & .661{\tiny$\pm$.019} & .671{\tiny$\pm$.032} & .692{\tiny$\pm$.030} & .680{\tiny$\pm$.034} & .651{\tiny$\pm$.057} & .586{\tiny$\pm$.019} & .625{\tiny$\pm$.016} & .652{\tiny$\pm$.020} \\
SAKT & .685{\tiny$\pm$.014} & .682{\tiny$\pm$.008} & .663{\tiny$\pm$.011} & .797{\tiny$\pm$.005} & .741{\tiny$\pm$.005} & .595{\tiny$\pm$.017} & .624{\tiny$\pm$.010} & .684{\tiny$\pm$.005} \\
simpleKT & \underline{.739}{\tiny$\pm$.007} & .690{\tiny$\pm$.007} & .709{\tiny$\pm$.006} & .790{\tiny$\pm$.008} & .761{\tiny$\pm$.024} & .705{\tiny$\pm$.004} & .646{\tiny$\pm$.005} & .720{\tiny$\pm$.003} \\
sparseKT & .732{\tiny$\pm$.008} & .703{\tiny$\pm$.031} & .710{\tiny$\pm$.007} & .814{\tiny$\pm$.009} & .770{\tiny$\pm$.009} & \underline{.708}{\tiny$\pm$.005} & .643{\tiny$\pm$.007} & .726{\tiny$\pm$.007} \\
stableKT & .732{\tiny$\pm$.005} & .700{\tiny$\pm$.006} & .721{\tiny$\pm$.001} & .768{\tiny$\pm$.034} & \underline{.790}{\tiny$\pm$.005} & .702{\tiny$\pm$.006} & .632{\tiny$\pm$.005} & .721{\tiny$\pm$.005} \\
UKT & .719{\tiny$\pm$.008} & .683{\tiny$\pm$.009} & .673{\tiny$\pm$.035} & .738{\tiny$\pm$.009} & \textbf{.792}{\tiny$\pm$.006} & .698{\tiny$\pm$.012} & .612{\tiny$\pm$.012} & .702{\tiny$\pm$.007} \\
\bottomrule
\end{tabular}
\end{table*}

\begin{table*}[p]
\centering
\scriptsize
\setlength{\tabcolsep}{3pt}
\caption{P1 with $N{=}256$ training learners: AUC of every model, mean{\tiny$\pm$SD} over 5 learner-subset seeds, on the 2{,}000 shared early targets per dataset. $^\ddagger$Three datasets of the original CLST paper. $^{(k)}$ = only $k$ seeds finished. Mean = mean over the 7 datasets (-- if a dataset is missing). qDKT and QIKT do not fit Algebra05 in 24\,GB of GPU memory (--). $^\circ$The method uses no data from the $N$ learners here (\JevZ{} and Thinking-KT everywhere; LOKT on datasets without answer options), so its $N{=}0$ run is repeated. Best in \textbf{bold}, second best \underline{underlined} (among cells with $\ge$3 seeds).}
\label{tab:p1n256}
\begin{tabular}{l cccccccc}
\toprule
\textbf{Method} & \textbf{A09} & \textbf{NIPS34} & \textbf{XES3G5M} & \textbf{DBE} & \textbf{Alg05} & \textbf{Junyi} & \textbf{A17} & \textbf{Mean} \\
\midrule
\multicolumn{9}{l}{\textit{System-One LLM}} \\
\rowcolor{focusbg} \textbf{JevKT} & \textbf{.746}{\tiny$\pm$.005} & \underline{.769}{\tiny$\pm$.004} & \underline{.752}{\tiny$\pm$.005} & .838{\tiny$\pm$.002} & .779{\tiny$\pm$.001} & \textbf{.738}{\tiny$\pm$.002} & .687{\tiny$\pm$.004} & \underline{.758}{\tiny$\pm$.001} \\
\JevZ{} & .738{\tiny$\pm$.014}$^\circ$ & .691{\tiny$\pm$.014}$^\circ$ & .725{\tiny$\pm$.013}$^\circ$ & .742{\tiny$\pm$.018}$^\circ$ & .698{\tiny$\pm$.018}$^\circ$ & .700{\tiny$\pm$.015}$^\circ$ & .648{\tiny$\pm$.013}$^\circ$ & .706{\tiny$\pm$.006}$^\circ$ \\
Laya (open-source Jev) & .586{\tiny$\pm$.006} & .579{\tiny$\pm$.005} & .512{\tiny$\pm$.005} & .602{\tiny$\pm$.002} & .533{\tiny$\pm$.005} & .534{\tiny$\pm$.004} & .504{\tiny$\pm$.001} & .550{\tiny$\pm$.001} \\
\midrule
\multicolumn{9}{l}{\textit{System-Two LLM-KT, prompted}} \\
Thinking-KT & .700{\tiny$\pm$.014}$^\circ$ & .641{\tiny$\pm$.014}$^\circ$ & .663{\tiny$\pm$.014}$^\circ$ & .666{\tiny$\pm$.019}$^\circ$ & .663{\tiny$\pm$.018}$^\circ$ & .642{\tiny$\pm$.016}$^\circ$ & .576{\tiny$\pm$.013}$^\circ$ & .650{\tiny$\pm$.006}$^\circ$ \\
LOKT & .706{\tiny$\pm$.003} & .659{\tiny$\pm$.004} & .699{\tiny$\pm$.014}$^\circ$ & .692{\tiny$\pm$.007} & .656{\tiny$\pm$.019}$^\circ$ & .671{\tiny$\pm$.016}$^\circ$ & .608{\tiny$\pm$.013}$^\circ$ & .670 \\
\midrule
\multicolumn{9}{l}{\textit{Classical KT}} \\
BKT & .736{\tiny$\pm$.003} & .650{\tiny$\pm$.002} & .657{\tiny$\pm$.009} & .754{\tiny$\pm$.003} & .731{\tiny$\pm$.001} & .679{\tiny$\pm$.006} & .642{\tiny$\pm$.002} & .693{\tiny$\pm$.001} \\
PFA & .717{\tiny$\pm$.004} & .648{\tiny$\pm$.001} & .649{\tiny$\pm$.006} & .817{\tiny$\pm$.001} & .660{\tiny$\pm$.001} & .672{\tiny$\pm$.014} & .635{\tiny$\pm$.005} & .686{\tiny$\pm$.002} \\
\midrule
\multicolumn{9}{l}{\textit{Deep KT (pyKT)}} \\
AKT & .728{\tiny$\pm$.004} & .709{\tiny$\pm$.025} & .715{\tiny$\pm$.006} & .834{\tiny$\pm$.003} & .800{\tiny$\pm$.004} & .703{\tiny$\pm$.005} & .620{\tiny$\pm$.003} & .730{\tiny$\pm$.004} \\
AT-DKT & .731{\tiny$\pm$.013} & .701{\tiny$\pm$.004} & .723{\tiny$\pm$.004} & .840{\tiny$\pm$.005} & .789{\tiny$\pm$.002} & .671{\tiny$\pm$.015} & .670{\tiny$\pm$.007} & .732{\tiny$\pm$.001} \\
ATKT & \textbf{.746}{\tiny$\pm$.006} & .698{\tiny$\pm$.004} & .721{\tiny$\pm$.005} & .824{\tiny$\pm$.004} & .777{\tiny$\pm$.006} & .696{\tiny$\pm$.004} & .666{\tiny$\pm$.012} & .733{\tiny$\pm$.002} \\
ATKT-fix & .641{\tiny$\pm$.050} & .678{\tiny$\pm$.009} & .687{\tiny$\pm$.004} & .798{\tiny$\pm$.009} & .765{\tiny$\pm$.004} & .620{\tiny$\pm$.017} & .609{\tiny$\pm$.011} & .685{\tiny$\pm$.009} \\
csKT & \textbf{.746}{\tiny$\pm$.014} & .767{\tiny$\pm$.007} & .743{\tiny$\pm$.004} & .774{\tiny$\pm$.042} & .782{\tiny$\pm$.003} & \underline{.723}{\tiny$\pm$.011} & .635{\tiny$\pm$.004} & .739{\tiny$\pm$.007} \\
Deep-IRT & .732{\tiny$\pm$.007} & .686{\tiny$\pm$.007} & .703{\tiny$\pm$.011} & .799{\tiny$\pm$.026} & .783{\tiny$\pm$.005} & .654{\tiny$\pm$.019} & .645{\tiny$\pm$.010} & .715{\tiny$\pm$.005} \\
DIMKT & .728{\tiny$\pm$.011} & .764{\tiny$\pm$.006} & .739{\tiny$\pm$.007} & \underline{.846}{\tiny$\pm$.005} & .794{\tiny$\pm$.005} & .703{\tiny$\pm$.013} & \underline{.693}{\tiny$\pm$.009} & .752{\tiny$\pm$.002} \\
DKT & .735{\tiny$\pm$.009} & .705{\tiny$\pm$.002} & .725{\tiny$\pm$.006} & .839{\tiny$\pm$.003} & .792{\tiny$\pm$.006} & .671{\tiny$\pm$.011} & .668{\tiny$\pm$.009} & .733{\tiny$\pm$.001} \\
DKT+ & .738{\tiny$\pm$.008} & .705{\tiny$\pm$.001} & .732{\tiny$\pm$.006} & .838{\tiny$\pm$.004} & .790{\tiny$\pm$.005} & .673{\tiny$\pm$.010} & .663{\tiny$\pm$.006} & .734{\tiny$\pm$.001} \\
DKVMN & .733{\tiny$\pm$.009} & .687{\tiny$\pm$.009} & .704{\tiny$\pm$.007} & .828{\tiny$\pm$.007} & .785{\tiny$\pm$.006} & .658{\tiny$\pm$.021} & .648{\tiny$\pm$.007} & .720{\tiny$\pm$.003} \\
DTransformer & .714{\tiny$\pm$.021} & .748{\tiny$\pm$.008} & .729{\tiny$\pm$.008} & .829{\tiny$\pm$.010} & .778{\tiny$\pm$.004} & .694{\tiny$\pm$.004} & .681{\tiny$\pm$.004} & .739{\tiny$\pm$.006} \\
extraKT & .729{\tiny$\pm$.004} & .724{\tiny$\pm$.033} & .714{\tiny$\pm$.005} & .791{\tiny$\pm$.030} & .783{\tiny$\pm$.005} & .710{\tiny$\pm$.007} & .630{\tiny$\pm$.003} & .726{\tiny$\pm$.008} \\
FoLiBiKT & .736{\tiny$\pm$.009} & .761{\tiny$\pm$.011} & .745{\tiny$\pm$.008} & .844{\tiny$\pm$.004} & .786{\tiny$\pm$.005} & .703{\tiny$\pm$.005} & .686{\tiny$\pm$.013} & .752{\tiny$\pm$.001} \\
IEKT & .733{\tiny$\pm$.009} & \textbf{.774}{\tiny$\pm$.006} & \textbf{.757}{\tiny$\pm$.005} & .842{\tiny$\pm$.013} & .793{\tiny$\pm$.005} & .720{\tiny$\pm$.007} & \textbf{.705}{\tiny$\pm$.009} & \textbf{.761}{\tiny$\pm$.003} \\
KQN & .717{\tiny$\pm$.017} & .686{\tiny$\pm$.002} & .728{\tiny$\pm$.002} & .804{\tiny$\pm$.028} & .774{\tiny$\pm$.009} & .659{\tiny$\pm$.016} & .632{\tiny$\pm$.003} & .714{\tiny$\pm$.004} \\
lefoKT-AKT & .739{\tiny$\pm$.007} & .739{\tiny$\pm$.008} & .703{\tiny$\pm$.008} & .805{\tiny$\pm$.005} & .772{\tiny$\pm$.012} & .706{\tiny$\pm$.006} & .636{\tiny$\pm$.006} & .729{\tiny$\pm$.002} \\
MoC-KT & .733{\tiny$\pm$.004} & .667{\tiny$\pm$.013} & .698{\tiny$\pm$.007} & .746{\tiny$\pm$.026} & .719{\tiny$\pm$.010} & .704{\tiny$\pm$.005} & .629{\tiny$\pm$.003} & .699{\tiny$\pm$.003} \\
qDKT & .572{\tiny$\pm$.012} & .741{\tiny$\pm$.004} & .679{\tiny$\pm$.010} & \underline{.846}{\tiny$\pm$.002} & -- & .663{\tiny$\pm$.011} & .692{\tiny$\pm$.009} & -- \\
QIKT & .709{\tiny$\pm$.010} & \underline{.769}{\tiny$\pm$.005} & .728{\tiny$\pm$.005} & \textbf{.852}{\tiny$\pm$.002} & -- & .683{\tiny$\pm$.016} & \textbf{.705}{\tiny$\pm$.006} & -- \\
reKT & .696{\tiny$\pm$.038} & .655{\tiny$\pm$.036} & .722{\tiny$\pm$.027} & .840{\tiny$\pm$.005} & .786{\tiny$\pm$.005} & .719{\tiny$\pm$.009} & .690{\tiny$\pm$.022} & .730{\tiny$\pm$.009} \\
robustKT & .728{\tiny$\pm$.004} & .693{\tiny$\pm$.019} & .718{\tiny$\pm$.004} & .839{\tiny$\pm$.005} & .803{\tiny$\pm$.002} & .702{\tiny$\pm$.008} & .624{\tiny$\pm$.004} & .730{\tiny$\pm$.003} \\
SAINT & .707{\tiny$\pm$.007} & .678{\tiny$\pm$.006} & .702{\tiny$\pm$.024} & .764{\tiny$\pm$.026} & .669{\tiny$\pm$.041} & .662{\tiny$\pm$.011} & .615{\tiny$\pm$.002} & .685{\tiny$\pm$.007} \\
SAINT+ & .659{\tiny$\pm$.025} & .673{\tiny$\pm$.085} & .700{\tiny$\pm$.037} & .820{\tiny$\pm$.006} & .736{\tiny$\pm$.009} & .666{\tiny$\pm$.003} & .637{\tiny$\pm$.057} & .699{\tiny$\pm$.019} \\
SAKT & .712{\tiny$\pm$.008} & .694{\tiny$\pm$.004} & .689{\tiny$\pm$.011} & .805{\tiny$\pm$.008} & .759{\tiny$\pm$.006} & .621{\tiny$\pm$.018} & .627{\tiny$\pm$.010} & .701{\tiny$\pm$.003} \\
simpleKT & \textbf{.746}{\tiny$\pm$.004} & .741{\tiny$\pm$.027} & .717{\tiny$\pm$.007} & .810{\tiny$\pm$.007} & .788{\tiny$\pm$.004} & .713{\tiny$\pm$.002} & .649{\tiny$\pm$.004} & .738{\tiny$\pm$.002} \\
sparseKT & \underline{.744}{\tiny$\pm$.005} & .751{\tiny$\pm$.015} & .728{\tiny$\pm$.006} & .827{\tiny$\pm$.006} & .777{\tiny$\pm$.005} & .718{\tiny$\pm$.005} & .674{\tiny$\pm$.009} & .746{\tiny$\pm$.002} \\
stableKT & .737{\tiny$\pm$.001} & .720{\tiny$\pm$.008} & .727{\tiny$\pm$.003} & .821{\tiny$\pm$.005} & \textbf{.808}{\tiny$\pm$.003} & .710{\tiny$\pm$.003} & .661{\tiny$\pm$.018} & .740{\tiny$\pm$.002} \\
UKT & .727{\tiny$\pm$.005} & .692{\tiny$\pm$.003} & .717{\tiny$\pm$.003} & .751{\tiny$\pm$.023} & \underline{.804}{\tiny$\pm$.004} & .715{\tiny$\pm$.004} & .616{\tiny$\pm$.009} & .717{\tiny$\pm$.005} \\
\bottomrule
\end{tabular}
\end{table*}

\begin{table*}[p]
\centering
\scriptsize
\setlength{\tabcolsep}{3pt}
\caption{P1 with $N{=}\mathrm{all}$ training learners: AUC of every model, mean{\tiny$\pm$SD} over seeds (pyKT; Jev, LLM and CLST rows are single runs), on the 2{,}000 shared early targets per dataset. $^\S$Reported by the fine-tuned LLM-KT~\citep{wang_2026_llmkt_acl_findings} (all training data, its own split; F1 not reported); not comparable to our targets and not ranked. $^\ddagger$Three datasets of the original CLST paper. $^{(k)}$ = only $k$ seeds finished. Mean = mean over the 7 datasets (-- if a dataset is missing). qDKT and QIKT do not fit Algebra05 in 24\,GB of GPU memory (--). $^\circ$The method uses no data from the $N$ learners here (\JevZ{} and Thinking-KT everywhere; LOKT on datasets without answer options), so its $N{=}0$ run is repeated. Best in \textbf{bold}, second best \underline{underlined} (among cells with $\ge$3 seeds).}
\label{tab:p1nall}
\begin{tabular}{l cccccccc}
\toprule
\textbf{Method} & \textbf{A09} & \textbf{NIPS34} & \textbf{XES3G5M} & \textbf{DBE} & \textbf{Alg05} & \textbf{Junyi} & \textbf{A17} & \textbf{Mean} \\
\midrule
\multicolumn{9}{l}{\textit{System-One LLM}} \\
\rowcolor{focusbg} \textbf{JevKT} & .780{\tiny$\pm$.012} & .774{\tiny$\pm$.011} & .758{\tiny$\pm$.013} & .829{\tiny$\pm$.014} & .780{\tiny$\pm$.016} & .771{\tiny$\pm$.013} & .701{\tiny$\pm$.013} & .770{\tiny$\pm$.005} \\
\JevZ{} & .738{\tiny$\pm$.014}$^\circ$ & .691{\tiny$\pm$.014}$^\circ$ & .725{\tiny$\pm$.013}$^\circ$ & .742{\tiny$\pm$.018}$^\circ$ & .698{\tiny$\pm$.018}$^\circ$ & .700{\tiny$\pm$.015}$^\circ$ & .648{\tiny$\pm$.013}$^\circ$ & .706{\tiny$\pm$.006}$^\circ$ \\
Laya (open-source Jev) & .581{\tiny$\pm$.015} & .579{\tiny$\pm$.014} & .509{\tiny$\pm$.015} & .596{\tiny$\pm$.018} & .538{\tiny$\pm$.018} & .512{\tiny$\pm$.015} & .505{\tiny$\pm$.014} & .546{\tiny$\pm$.006} \\
\midrule
\multicolumn{9}{l}{\textit{System-Two LLM-KT, prompted}} \\
Thinking-KT & .700{\tiny$\pm$.014}$^\circ$ & .641{\tiny$\pm$.014}$^\circ$ & .663{\tiny$\pm$.014}$^\circ$ & .666{\tiny$\pm$.019}$^\circ$ & .663{\tiny$\pm$.018}$^\circ$ & .642{\tiny$\pm$.016}$^\circ$ & .576{\tiny$\pm$.013}$^\circ$ & .650{\tiny$\pm$.006}$^\circ$ \\
LOKT & .705{\tiny$\pm$.015} & .666{\tiny$\pm$.014} & .699{\tiny$\pm$.014}$^\circ$ & .700{\tiny$\pm$.021} & .656{\tiny$\pm$.019}$^\circ$ & .671{\tiny$\pm$.016}$^\circ$ & .608{\tiny$\pm$.013}$^\circ$ & .672{\tiny$\pm$.006} \\
\bottomrule
\end{tabular}
\end{table*}

\begin{table*}[tp]
\centering
\scriptsize
\setlength{\tabcolsep}{3pt}
\caption{P1 with $N{=}0$ training learners: ACC\,/\,F1 of every model (threshold .5, F1 of the correct class; mean over seeds), same cells as Table~\ref{tab:p1n0}. Best in \textbf{bold}, second best \underline{underlined}.}
\label{tab:p1fn0}
\label{tab:accf1}
\begin{tabular}{l cccccccc}
\toprule
\textbf{Method} & \textbf{A09} & \textbf{NIPS34} & \textbf{XES3G5M} & \textbf{DBE} & \textbf{Alg05} & \textbf{Junyi} & \textbf{A17} & \textbf{Mean} \\
\midrule
\multicolumn{9}{l}{\textit{System-One LLM}} \\
\rowcolor{focusbg} \JevZ{} & \textbf{.707}/\textbf{.779} & \textbf{.636}/\textbf{.688} & \textbf{.751}/\textbf{.840} & \textbf{.836}/\textbf{.908} & \textbf{.734}/\textbf{.827} & \textbf{.751}/\textbf{.840} & \textbf{.646}/\textbf{.423} & \textbf{.723}/\textbf{.758} \\
Laya (open-source Jev) & .394/.203 & .442/.057 & .429/.470 & .723/.828 & .373/.355 & .341/.269 & .473/\underline{.385} & .454/.367 \\
\midrule
\multicolumn{9}{l}{\textit{System-Two LLM-KT, prompted}} \\
Thinking-KT & .655/.704 & \underline{.611}/.615 & .662/.753 & .756/.849 & .678/.769 & .658/.755 & .617/.265 & .663/.673 \\
LOKT & .637/.677 & .604/.575 & \underline{.713}/\underline{.803} & \underline{.777}/\underline{.863} & .703/.804 & \underline{.742}/\underline{.833} & \underline{.636}/.349 & \underline{.687}/\underline{.700} \\
\midrule
\multicolumn{9}{l}{\textit{System-Two LLM-KT, trained}} \\
CLST$^\ddagger$ & \underline{.686}/\underline{.752} & .581/\underline{.626} & -- & -- & \underline{.726}/\underline{.822} & -- & -- & -- \\
\bottomrule
\end{tabular}
\end{table*}

\begin{table*}[p]
\centering
\scriptsize
\setlength{\tabcolsep}{3pt}
\caption{P1 with $N{=}8$ training learners: ACC\,/\,F1 of every model (threshold .5, F1 of the correct class; mean over seeds), same cells as Table~\ref{tab:p1n8}. Best in \textbf{bold}, second best \underline{underlined}.}
\label{tab:p1fn8}
\begin{tabular}{l cccccccc}
\toprule
\textbf{Method} & \textbf{A09} & \textbf{NIPS34} & \textbf{XES3G5M} & \textbf{DBE} & \textbf{Alg05} & \textbf{Junyi} & \textbf{A17} & \textbf{Mean} \\
\midrule
\multicolumn{9}{l}{\textit{System-One LLM}} \\
\rowcolor{focusbg} \textbf{JevKT} & \textbf{.711}/\textbf{.790} & \textbf{.650}/\textbf{.694} & \underline{.750}/.850 & \textbf{.855}/\textbf{.920} & \textbf{.765}/\textbf{.856} & \textbf{.760}/.848 & .635/.317 & \textbf{.732}/.753 \\
\JevZ{} & \underline{.707}/\underline{.779}$^\circ$ & \underline{.636}/.688$^\circ$ & \textbf{.751}/.840$^\circ$ & .836/.908$^\circ$ & .734/.827$^\circ$ & \underline{.751}/.840$^\circ$ & \textbf{.646}/.423$^\circ$ & \underline{.723}/\textbf{.758}$^\circ$ \\
Laya (open-source Jev) & .371/.058 & .448/.091 & .326/.226 & .699/.808 & .334/.228 & .282/.080 & .609/.080 & .438/.224 \\
\midrule
\multicolumn{9}{l}{\textit{System-Two LLM-KT, prompted}} \\
Thinking-KT & .655/.704$^\circ$ & .611/.615$^\circ$ & .662/.753$^\circ$ & .756/.849$^\circ$ & .678/.769$^\circ$ & .658/.755$^\circ$ & .617/.265$^\circ$ & .663/.673$^\circ$ \\
LOKT & .632/.673 & .608/.590 & .713/.803$^\circ$ & .769/.857 & .703/.804$^\circ$ & .742/.833$^\circ$ & \underline{.636}/.349$^\circ$ & .686/.701 \\
\bottomrule
\end{tabular}
\end{table*}

\begin{table*}[p]
\centering
\scriptsize
\setlength{\tabcolsep}{3pt}
\caption{P1 with $N{=}16$ training learners: ACC\,/\,F1 of every model (threshold .5, F1 of the correct class; mean over seeds), same cells as Table~\ref{tab:p1n16}. Best in \textbf{bold}, second best \underline{underlined}.}
\label{tab:p1fn16}
\begin{tabular}{l cccccccc}
\toprule
\textbf{Method} & \textbf{A09} & \textbf{NIPS34} & \textbf{XES3G5M} & \textbf{DBE} & \textbf{Alg05} & \textbf{Junyi} & \textbf{A17} & \textbf{Mean} \\
\midrule
\multicolumn{9}{l}{\textit{System-One LLM}} \\
\rowcolor{focusbg} \textbf{JevKT} & \textbf{.716}/\textbf{.800} & \textbf{.667}/\underline{.721} & \textbf{.753}/\textbf{.853} & \textbf{.860}/\textbf{.923} & \textbf{.770}/\textbf{.859} & \textbf{.760}/.849 & \underline{.642}/.300 & \textbf{.738}/\underline{.758} \\
\JevZ{} & \underline{.707}/.779$^\circ$ & .636/.688$^\circ$ & .751/.840$^\circ$ & .836/.908$^\circ$ & .734/.827$^\circ$ & \underline{.751}/.840$^\circ$ & \textbf{.646}/.423$^\circ$ & \underline{.723}/\underline{.758}$^\circ$ \\
Laya (open-source Jev) & .380/.088 & .452/.117 & .347/.283 & .743/.844 & .336/.239 & .279/.073 & .609/.088 & .449/.247 \\
\midrule
\multicolumn{9}{l}{\textit{System-Two LLM-KT, prompted}} \\
Thinking-KT & .655/.704$^\circ$ & .611/.615$^\circ$ & .662/.753$^\circ$ & .756/.849$^\circ$ & .678/.769$^\circ$ & .658/.755$^\circ$ & .617/.265$^\circ$ & .663/.673$^\circ$ \\
LOKT & .634/.675 & .603/.578 & .713/.803$^\circ$ & .766/.855 & .703/.804$^\circ$ & .742/.833$^\circ$ & .636/.349$^\circ$ & .685/.700 \\
\bottomrule
\end{tabular}
\end{table*}

\begin{table*}[p]
\centering
\scriptsize
\setlength{\tabcolsep}{3pt}
\caption{P1 with $N{=}32$ training learners: ACC\,/\,F1 of every model (threshold .5, F1 of the correct class; mean over seeds), same cells as Table~\ref{tab:p1n32}. Best in \textbf{bold}, second best \underline{underlined}.}
\label{tab:p1fn32}
\begin{tabular}{l cccccccc}
\toprule
\textbf{Method} & \textbf{A09} & \textbf{NIPS34} & \textbf{XES3G5M} & \textbf{DBE} & \textbf{Alg05} & \textbf{Junyi} & \textbf{A17} & \textbf{Mean} \\
\midrule
\multicolumn{9}{l}{\textit{System-One LLM}} \\
\rowcolor{focusbg} \textbf{JevKT} & \underline{.713}/\textbf{.801} & \textbf{.684}/\textbf{.732} & \underline{.751}/\textbf{.853} & \textbf{.861}/\textbf{.923} & \underline{.775}/\underline{.861} & \textbf{.755}/.845 & \underline{.646}/.342 & \textbf{.741}/.765 \\
\JevZ{} & .707/.779$^\circ$ & .636/.688$^\circ$ & \underline{.751}/.840$^\circ$ & .836/.908$^\circ$ & .734/.827$^\circ$ & \underline{.751}/.840$^\circ$ & \underline{.646}/.423$^\circ$ & .723/.758$^\circ$ \\
Laya (open-source Jev) & .380/.091 & .443/.064 & .364/.326 & .692/.804 & .347/.269 & .280/.074 & .607/.103 & .445/.247 \\
\midrule
\multicolumn{9}{l}{\textit{System-Two LLM-KT, prompted}} \\
Thinking-KT & .655/.704$^\circ$ & .611/.615$^\circ$ & .662/.753$^\circ$ & .756/.849$^\circ$ & .678/.769$^\circ$ & .658/.755$^\circ$ & .617/.265$^\circ$ & .663/.673$^\circ$ \\
LOKT & .634/.674 & .600/.567 & .713/.803$^\circ$ & .766/.855 & .703/.804$^\circ$ & .742/.833$^\circ$ & .636/.349$^\circ$ & .685/.698 \\
\bottomrule
\end{tabular}
\end{table*}

\begin{table*}[p]
\centering
\scriptsize
\setlength{\tabcolsep}{3pt}
\caption{P1 with $N{=}64$ training learners: ACC\,/\,F1 of every model (threshold .5, F1 of the correct class; mean over seeds), same cells as Table~\ref{tab:p1n64}. Best in \textbf{bold}, second best \underline{underlined}.}
\label{tab:p1fn64}
\begin{tabular}{l cccccccc}
\toprule
\textbf{Method} & \textbf{A09} & \textbf{NIPS34} & \textbf{XES3G5M} & \textbf{DBE} & \textbf{Alg05} & \textbf{Junyi} & \textbf{A17} & \textbf{Mean} \\
\midrule
\multicolumn{9}{l}{\textit{System-One LLM}} \\
\rowcolor{focusbg} \textbf{JevKT} & \textbf{.716}/\textbf{.806} & \textbf{.695}/\textbf{.747} & \underline{.753}/\textbf{.854} & \underline{.862}/\textbf{.924} & .779/.863 & \underline{.756}/.847 & \underline{.650}/.373 & \textbf{.744}/.773 \\
\JevZ{} & .707/.779$^\circ$ & .636/.688$^\circ$ & .751/.840$^\circ$ & .836/.908$^\circ$ & .734/.827$^\circ$ & .751/.840$^\circ$ & .646/.423$^\circ$ & .723/.758$^\circ$ \\
Laya (open-source Jev) & .384/.105 & .443/.069 & .374/.356 & .630/.752 & .357/.286 & .279/.071 & .605/.100 & .439/.248 \\
\midrule
\multicolumn{9}{l}{\textit{System-Two LLM-KT, prompted}} \\
Thinking-KT & .655/.704$^\circ$ & .611/.615$^\circ$ & .662/.753$^\circ$ & .756/.849$^\circ$ & .678/.769$^\circ$ & .658/.755$^\circ$ & .617/.265$^\circ$ & .663/.673$^\circ$ \\
LOKT & .637/.680 & .602/.570 & .713/.803$^\circ$ & .762/.852 & .703/.804$^\circ$ & .742/.833$^\circ$ & .636/.349$^\circ$ & .685/.699 \\
\bottomrule
\end{tabular}
\end{table*}

\begin{table*}[p]
\centering
\scriptsize
\setlength{\tabcolsep}{3pt}
\caption{P1 with $N{=}128$ training learners: ACC\,/\,F1 of every model (threshold .5, F1 of the correct class; mean over seeds), same cells as Table~\ref{tab:p1n128}. Best in \textbf{bold}, second best \underline{underlined}.}
\label{tab:p1fn128}
\begin{tabular}{l cccccccc}
\toprule
\textbf{Method} & \textbf{A09} & \textbf{NIPS34} & \textbf{XES3G5M} & \textbf{DBE} & \textbf{Alg05} & \textbf{Junyi} & \textbf{A17} & \textbf{Mean} \\
\midrule
\multicolumn{9}{l}{\textit{System-One LLM}} \\
\rowcolor{focusbg} \textbf{JevKT} & \textbf{.716}/\textbf{.803} & \underline{.695}/\underline{.744} & \underline{.756}/\textbf{.855} & .864/\underline{.925} & .782/.865 & \underline{.760}/\underline{.852} & \textbf{.661}/.388 & \textbf{.748}/.776 \\
\JevZ{} & .707/.779$^\circ$ & .636/.688$^\circ$ & .751/.840$^\circ$ & .836/.908$^\circ$ & .734/.827$^\circ$ & .751/.840$^\circ$ & .646/.423$^\circ$ & .723/.758$^\circ$ \\
Laya (open-source Jev) & .387/.115 & .444/.071 & .381/.370 & .634/.755 & .357/.288 & .285/.086 & .604/.106 & .442/.256 \\
\midrule
\multicolumn{9}{l}{\textit{System-Two LLM-KT, prompted}} \\
Thinking-KT & .655/.704$^\circ$ & .611/.615$^\circ$ & .662/.753$^\circ$ & .756/.849$^\circ$ & .678/.769$^\circ$ & .658/.755$^\circ$ & .617/.265$^\circ$ & .663/.673$^\circ$ \\
LOKT & .638/.682 & .595/.558 & .713/.803$^\circ$ & .771/.859 & .703/.804$^\circ$ & .742/.833$^\circ$ & .636/.349$^\circ$ & .685/.698 \\
\bottomrule
\end{tabular}
\end{table*}

\begin{table*}[p]
\centering
\scriptsize
\setlength{\tabcolsep}{3pt}
\caption{P1 with $N{=}256$ training learners: ACC\,/\,F1 of every model (threshold .5, F1 of the correct class; mean over seeds), same cells as Table~\ref{tab:p1n256}. Best in \textbf{bold}, second best \underline{underlined}.}
\label{tab:p1fn256}
\begin{tabular}{l cccccccc}
\toprule
\textbf{Method} & \textbf{A09} & \textbf{NIPS34} & \textbf{XES3G5M} & \textbf{DBE} & \textbf{Alg05} & \textbf{Junyi} & \textbf{A17} & \textbf{Mean} \\
\midrule
\multicolumn{9}{l}{\textit{System-One LLM}} \\
\rowcolor{focusbg} \textbf{JevKT} & .717/.800 & .699/\textbf{.746} & \underline{.760}/\underline{.857} & .865/\underline{.925} & .787/.868 & \textbf{.763}/\textbf{.853} & \underline{.668}/.399 & \underline{.751}/.778 \\
\JevZ{} & .707/.779$^\circ$ & .636/.688$^\circ$ & .751/.840$^\circ$ & .836/.908$^\circ$ & .734/.827$^\circ$ & .751/.840$^\circ$ & .646/.423$^\circ$ & .723/.758$^\circ$ \\
Laya (open-source Jev) & .387/.115 & .443/.065 & .380/.363 & .631/.753 & .346/.260 & .290/.103 & .603/.104 & .440/.252 \\
\midrule
\multicolumn{9}{l}{\textit{System-Two LLM-KT, prompted}} \\
Thinking-KT & .655/.704$^\circ$ & .611/.615$^\circ$ & .662/.753$^\circ$ & .756/.849$^\circ$ & .678/.769$^\circ$ & .658/.755$^\circ$ & .617/.265$^\circ$ & .663/.673$^\circ$ \\
LOKT & .642/.687 & .603/.570 & .713/.803$^\circ$ & .766/.856 & .703/.804$^\circ$ & .742/.833$^\circ$ & .636/.349$^\circ$ & .686/.700 \\
\bottomrule
\end{tabular}
\end{table*}

\begin{table*}[p]
\centering
\scriptsize
\setlength{\tabcolsep}{3pt}
\caption{P1 with $N{=}\mathrm{all}$ training learners: ACC\,/\,F1 of every model (threshold .5, F1 of the correct class; mean over seeds), same cells as Table~\ref{tab:p1nall}. Best in \textbf{bold}, second best \underline{underlined}.}
\label{tab:p1fnall}
\begin{tabular}{l cccccccc}
\toprule
\textbf{Method} & \textbf{A09} & \textbf{NIPS34} & \textbf{XES3G5M} & \textbf{DBE} & \textbf{Alg05} & \textbf{Junyi} & \textbf{A17} & \textbf{Mean} \\
\midrule
\multicolumn{9}{l}{\textit{System-One LLM}} \\
\rowcolor{focusbg} \textbf{JevKT} & .738/\textbf{.813} & .699/.743 & .769/.860 & .868/\textbf{.927} & .784/.866 & \textbf{.779}/\textbf{.862} & .674/.418 & .759/.784 \\
\JevZ{} & .707/.779$^\circ$ & .636/.688$^\circ$ & .751/.840$^\circ$ & .836/.908$^\circ$ & .734/.827$^\circ$ & .751/.840$^\circ$ & .646/.423$^\circ$ & .723/.758$^\circ$ \\
Laya (open-source Jev) & .389/.116 & .447/.081 & .367/.337 & .624/.747 & .344/.250 & .281/.082 & .603/.106 & .436/.246 \\
\midrule
\multicolumn{9}{l}{\textit{System-Two LLM-KT, prompted}} \\
Thinking-KT & .655/.704$^\circ$ & .611/.615$^\circ$ & .662/.753$^\circ$ & .756/.849$^\circ$ & .678/.769$^\circ$ & .658/.755$^\circ$ & .617/.265$^\circ$ & .663/.673$^\circ$ \\
LOKT & .641/.688 & .607/.577 & .713/.803$^\circ$ & .759/.851 & .703/.804$^\circ$ & .742/.833$^\circ$ & .636/.349$^\circ$ & .686/.701 \\
\bottomrule
\end{tabular}
\end{table*}

\begin{table*}[p]
\centering
\scriptsize
\setlength{\tabcolsep}{6pt}
\caption{P2 with $N{=}0$: AUC by the position of the target in a new learner's sequence, mean over the 7 datasets ({\tiny$\pm$} = SD over seeds), for every model with all 7 datasets. $^{(k)}$ = only $k$ seeds finished; Best in \textbf{bold}, second best \underline{underlined} (among cells with $\ge$3 seeds).}
\label{tab:p2n0}
\begin{tabular}{l cccccc}
\toprule
\textbf{Method} & 1 & 2--5 & 6--10 & 11--15 & 16--30 & 31--49 \\
\midrule
\multicolumn{7}{l}{\textit{System-One LLM}} \\
\rowcolor{focusbg} \JevZ{} & .580 & \textbf{.687} & \textbf{.712} & \textbf{.710} & \textbf{.704} & \textbf{.717} \\
Laya (open-source Jev) & .572 & .539 & .615 & .547 & .533 & .528 \\
\midrule
\multicolumn{7}{l}{\textit{System-Two LLM-KT, prompted}} \\
Thinking-KT & \underline{.598} & .641 & .614 & .659 & .662 & .662 \\
LOKT & \textbf{.603} & \underline{.664} & \underline{.683} & \underline{.684} & \underline{.668} & \underline{.676} \\
\bottomrule
\end{tabular}
\end{table*}

\begin{table*}[p]
\centering
\scriptsize
\setlength{\tabcolsep}{6pt}
\caption{P2 with $N{=}8$: AUC by the position of the target in a new learner's sequence, mean over the 7 datasets ({\tiny$\pm$} = SD over seeds), for every model with all 7 datasets. $^{(k)}$ = only $k$ seeds finished; $^\circ$Uses no data from the $N$ learners (LOKT: on 4 of the 7 datasets), so the $N{=}0$ run is repeated (fully or in part). Best in \textbf{bold}, second best \underline{underlined} (among cells with $\ge$3 seeds).}
\label{tab:p2n8}
\begin{tabular}{l cccccc}
\toprule
\textbf{Method} & 1 & 2--5 & 6--10 & 11--15 & 16--30 & 31--49 \\
\midrule
\multicolumn{7}{l}{\textit{System-One LLM}} \\
\rowcolor{focusbg} \textbf{JevKT} & \textbf{.640}{\tiny$\pm$.023} & \textbf{.714}{\tiny$\pm$.012} & \textbf{.726}{\tiny$\pm$.007} & \textbf{.723}{\tiny$\pm$.008} & \textbf{.725}{\tiny$\pm$.006} & \textbf{.728}{\tiny$\pm$.003} \\
\JevZ{}$^\circ$ & .580 & \underline{.687} & \underline{.712} & \underline{.710} & \underline{.704} & \underline{.717} \\
Laya (open-source Jev) & .556{\tiny$\pm$.012} & .599{\tiny$\pm$.014} & .639{\tiny$\pm$.014} & .574{\tiny$\pm$.005} & .545{\tiny$\pm$.002} & .522{\tiny$\pm$.005} \\
\midrule
\multicolumn{7}{l}{\textit{System-Two LLM-KT, prompted}} \\
Thinking-KT$^\circ$ & .598 & .641 & .614 & .659 & .662 & .662 \\
LOKT$^\circ$ & \underline{.602} & .661 & .679 & .679 & .667 & .677 \\
\bottomrule
\end{tabular}
\end{table*}

\begin{table*}[p]
\centering
\scriptsize
\setlength{\tabcolsep}{6pt}
\caption{P2 with $N{=}16$: AUC by the position of the target in a new learner's sequence, mean over the 7 datasets ({\tiny$\pm$} = SD over seeds), for every model with all 7 datasets. $^{(k)}$ = only $k$ seeds finished; $^\circ$Uses no data from the $N$ learners (LOKT: on 4 of the 7 datasets), so the $N{=}0$ run is repeated (fully or in part). Best in \textbf{bold}, second best \underline{underlined} (among cells with $\ge$3 seeds).}
\label{tab:p2n16}
\begin{tabular}{l cccccc}
\toprule
\textbf{Method} & 1 & 2--5 & 6--10 & 11--15 & 16--30 & 31--49 \\
\midrule
\multicolumn{7}{l}{\textit{System-One LLM}} \\
\rowcolor{focusbg} \textbf{JevKT} & \textbf{.658}{\tiny$\pm$.027} & \textbf{.719}{\tiny$\pm$.011} & \textbf{.735}{\tiny$\pm$.008} & \textbf{.721}{\tiny$\pm$.003} & \textbf{.733}{\tiny$\pm$.006} & \textbf{.737}{\tiny$\pm$.003} \\
\JevZ{}$^\circ$ & .580 & .687 & \underline{.712} & \underline{.710} & \underline{.704} & \underline{.717} \\
Laya (open-source Jev) & .564{\tiny$\pm$.012} & .605{\tiny$\pm$.006} & .642{\tiny$\pm$.005} & .580{\tiny$\pm$.009} & .544{\tiny$\pm$.004} & .522{\tiny$\pm$.004} \\
\midrule
\multicolumn{7}{l}{\textit{System-Two LLM-KT, prompted}} \\
Thinking-KT$^\circ$ & .598 & .641 & .614 & .659 & .662 & .662 \\
LOKT$^\circ$ & .602 & .659 & .684 & .682 & .668 & .677 \\
\bottomrule
\end{tabular}
\end{table*}

\begin{table*}[p]
\centering
\scriptsize
\setlength{\tabcolsep}{6pt}
\caption{P2 with $N{=}32$: AUC by the position of the target in a new learner's sequence, mean over the 7 datasets ({\tiny$\pm$} = SD over seeds), for every model with all 7 datasets. $^{(k)}$ = only $k$ seeds finished; $^\circ$Uses no data from the $N$ learners (LOKT: on 4 of the 7 datasets), so the $N{=}0$ run is repeated (fully or in part). Best in \textbf{bold}, second best \underline{underlined} (among cells with $\ge$3 seeds).}
\label{tab:p2n32}
\begin{tabular}{l cccccc}
\toprule
\textbf{Method} & 1 & 2--5 & 6--10 & 11--15 & 16--30 & 31--49 \\
\midrule
\multicolumn{7}{l}{\textit{System-One LLM}} \\
\rowcolor{focusbg} \textbf{JevKT} & \textbf{.652}{\tiny$\pm$.023} & \textbf{.726}{\tiny$\pm$.005} & \textbf{.737}{\tiny$\pm$.007} & \textbf{.735}{\tiny$\pm$.008} & \textbf{.741}{\tiny$\pm$.003} & \textbf{.746}{\tiny$\pm$.003} \\
\JevZ{}$^\circ$ & .580 & .687 & \underline{.712} & .710 & \underline{.704} & .717 \\
Laya (open-source Jev) & .555{\tiny$\pm$.014} & .610{\tiny$\pm$.006} & .645{\tiny$\pm$.005} & .575{\tiny$\pm$.004} & .536{\tiny$\pm$.005} & .520{\tiny$\pm$.002} \\
\midrule
\multicolumn{7}{l}{\textit{System-Two LLM-KT, prompted}} \\
Thinking-KT$^\circ$ & .598 & .641 & .614 & .659 & .662 & .662 \\
LOKT$^\circ$ & .598 & .661 & .685 & .682 & .670 & .675 \\
\bottomrule
\end{tabular}
\end{table*}

\begin{table*}[p]
\centering
\scriptsize
\setlength{\tabcolsep}{6pt}
\caption{P2 with $N{=}64$: AUC by the position of the target in a new learner's sequence, mean over the 7 datasets ({\tiny$\pm$} = SD over seeds), for every model with all 7 datasets. $^{(k)}$ = only $k$ seeds finished; $^\circ$Uses no data from the $N$ learners (LOKT: on 4 of the 7 datasets), so the $N{=}0$ run is repeated (fully or in part). Best in \textbf{bold}, second best \underline{underlined} (among cells with $\ge$3 seeds).}
\label{tab:p2n64}
\begin{tabular}{l cccccc}
\toprule
\textbf{Method} & 1 & 2--5 & 6--10 & 11--15 & 16--30 & 31--49 \\
\midrule
\multicolumn{7}{l}{\textit{System-One LLM}} \\
\rowcolor{focusbg} \textbf{JevKT} & \textbf{.668}{\tiny$\pm$.017} & \textbf{.731}{\tiny$\pm$.004} & \textbf{.748}{\tiny$\pm$.005} & \textbf{.738}{\tiny$\pm$.005} & \textbf{.748}{\tiny$\pm$.002} & \textbf{.756}{\tiny$\pm$.005} \\
\JevZ{}$^\circ$ & .580 & .687 & .712 & .710 & .704 & .717 \\
Laya (open-source Jev) & .577{\tiny$\pm$.011} & .610{\tiny$\pm$.007} & .641{\tiny$\pm$.004} & .580{\tiny$\pm$.008} & .535{\tiny$\pm$.004} & .514{\tiny$\pm$.004} \\
\midrule
\multicolumn{7}{l}{\textit{System-Two LLM-KT, prompted}} \\
Thinking-KT$^\circ$ & .598 & .641 & .614 & .659 & .662 & .662 \\
LOKT$^\circ$ & .601 & .655 & .681 & .678 & .669 & .676 \\
\bottomrule
\end{tabular}
\end{table*}

\begin{table*}[p]
\centering
\scriptsize
\setlength{\tabcolsep}{6pt}
\caption{P2 with $N{=}128$: AUC by the position of the target in a new learner's sequence, mean over the 7 datasets ({\tiny$\pm$} = SD over seeds), for every model with all 7 datasets. $^{(k)}$ = only $k$ seeds finished; $^\circ$Uses no data from the $N$ learners (LOKT: on 4 of the 7 datasets), so the $N{=}0$ run is repeated (fully or in part). Best in \textbf{bold}, second best \underline{underlined} (among cells with $\ge$3 seeds).}
\label{tab:p2n128}
\begin{tabular}{l cccccc}
\toprule
\textbf{Method} & 1 & 2--5 & 6--10 & 11--15 & 16--30 & 31--49 \\
\midrule
\multicolumn{7}{l}{\textit{System-One LLM}} \\
\rowcolor{focusbg} \textbf{JevKT} & \underline{.677}{\tiny$\pm$.015} & \underline{.731}{\tiny$\pm$.008} & \textbf{.759}{\tiny$\pm$.006} & \textbf{.743}{\tiny$\pm$.005} & \textbf{.753}{\tiny$\pm$.004} & \textbf{.764}{\tiny$\pm$.002} \\
\JevZ{}$^\circ$ & .580 & .687 & .712 & .710 & .704 & .717 \\
Laya (open-source Jev) & .590{\tiny$\pm$.018} & .612{\tiny$\pm$.003} & .638{\tiny$\pm$.006} & .584{\tiny$\pm$.007} & .532{\tiny$\pm$.002} & .515{\tiny$\pm$.003} \\
\midrule
\multicolumn{7}{l}{\textit{System-Two LLM-KT, prompted}} \\
Thinking-KT$^\circ$ & .598 & .641 & .614 & .659 & .662 & .662 \\
LOKT$^\circ$ & .597 & .660 & .681 & .678 & .668 & .675 \\
\bottomrule
\end{tabular}
\end{table*}

\begin{table*}[p]
\centering
\scriptsize
\setlength{\tabcolsep}{6pt}
\caption{P2 with $N{=}256$: AUC by the position of the target in a new learner's sequence, mean over the 7 datasets ({\tiny$\pm$} = SD over seeds), for every model with all 7 datasets. $^{(k)}$ = only $k$ seeds finished; $^\circ$Uses no data from the $N$ learners (LOKT: on 4 of the 7 datasets), so the $N{=}0$ run is repeated (fully or in part). Best in \textbf{bold}, second best \underline{underlined} (among cells with $\ge$3 seeds).}
\label{tab:p2n256}
\begin{tabular}{l cccccc}
\toprule
\textbf{Method} & 1 & 2--5 & 6--10 & 11--15 & 16--30 & 31--49 \\
\midrule
\multicolumn{7}{l}{\textit{System-One LLM}} \\
\rowcolor{focusbg} \textbf{JevKT} & .679{\tiny$\pm$.015} & \underline{.736}{\tiny$\pm$.008} & \underline{.763}{\tiny$\pm$.006} & \textbf{.748}{\tiny$\pm$.003} & \textbf{.760}{\tiny$\pm$.002} & \textbf{.768}{\tiny$\pm$.003} \\
\JevZ{}$^\circ$ & .580 & .687 & .712 & .710 & .704 & .717 \\
Laya (open-source Jev) & .594{\tiny$\pm$.009} & .610{\tiny$\pm$.002} & .643{\tiny$\pm$.009} & .585{\tiny$\pm$.007} & .535{\tiny$\pm$.003} & .516{\tiny$\pm$.002} \\
\midrule
\multicolumn{7}{l}{\textit{System-Two LLM-KT, prompted}} \\
Thinking-KT$^\circ$ & .598 & .641 & .614 & .659 & .662 & .662 \\
LOKT$^\circ$ & .604 & .658 & .687 & .677 & .664 & .676 \\
\bottomrule
\end{tabular}
\end{table*}

\begin{table*}[p]
\centering
\scriptsize
\setlength{\tabcolsep}{6pt}
\caption{P2 with $N{=}\mathrm{all}$: AUC by the position of the target in a new learner's sequence, mean over the 7 datasets ({\tiny$\pm$} = SD over seeds), for every model with all 7 datasets. $^{(k)}$ = only $k$ seeds finished; $^\circ$Uses no data from the $N$ learners (LOKT: on 4 of the 7 datasets), so the $N{=}0$ run is repeated (fully or in part). Best in \textbf{bold}, second best \underline{underlined} (among cells with $\ge$3 seeds).}
\label{tab:p2nall}
\begin{tabular}{l cccccc}
\toprule
\textbf{Method} & 1 & 2--5 & 6--10 & 11--15 & 16--30 & 31--49 \\
\midrule
\multicolumn{7}{l}{\textit{System-One LLM}} \\
\rowcolor{focusbg} \textbf{JevKT} & .689 & .745 & .773 & .758 & .771 & .785 \\
\JevZ{}$^\circ$ & .580 & .687 & .712 & .710 & .704 & .717 \\
Laya (open-source Jev) & .589 & .608 & .646 & .572 & .534 & .508 \\
\midrule
\multicolumn{7}{l}{\textit{System-Two LLM-KT, prompted}} \\
Thinking-KT$^\circ$ & .598 & .641 & .614 & .659 & .662 & .662 \\
LOKT$^\circ$ & .604 & .664 & .681 & .676 & .667 & .679 \\
\bottomrule
\end{tabular}
\end{table*}

\begin{table*}[p]
\centering
\scriptsize
\setlength{\tabcolsep}{3pt}
\caption{P3 (new items): AUC of every model on the held-out items, mean{\tiny$\pm$SD} over 3 item-split seeds
($^{(k)}$ = only $k$ finished). $^\S$Option weights from the training learners where options exist. Best in \textbf{bold}, second best \underline{underlined} (among cells with $\ge$3 seeds).}
\label{tab:p3all}
\begin{tabular}{l cccccccc}
\toprule
\textbf{Method} & \textbf{A09} & \textbf{NIPS34} & \textbf{XES3G5M} & \textbf{DBE} & \textbf{Alg05} & \textbf{Junyi} & \textbf{A17} & \textbf{Mean} \\
\midrule
\multicolumn{9}{l}{\textit{System-One LLM}} \\
\rowcolor{focusbg} \textbf{JevKT} & .728{\tiny$\pm$.029} & .725{\tiny$\pm$.014} & .707{\tiny$\pm$.021} & .687{\tiny$\pm$.047} & .775{\tiny$\pm$.015} & \textbf{.731}{\tiny$\pm$.039} & .639{\tiny$\pm$.012} & .713 \\
\rowcolor{focusbg} \JevZ{} & .706{\tiny$\pm$.034} & .717{\tiny$\pm$.018} & \underline{.718}{\tiny$\pm$.016} & \textbf{.705}{\tiny$\pm$.048} & .737{\tiny$\pm$.018} & .714{\tiny$\pm$.034} & .629{\tiny$\pm$.013} & .704 \\
\midrule
\multicolumn{9}{l}{\textit{System-Two LLM-KT, prompted}} \\
Thinking-KT & .670{\tiny$\pm$.026} & .667{\tiny$\pm$.029} & .651{\tiny$\pm$.007} & .669{\tiny$\pm$.034} & .689{\tiny$\pm$.009} & .662{\tiny$\pm$.037} & .593{\tiny$\pm$.012} & .657 \\
LOKT$^\S$ & .688{\tiny$\pm$.036} & .683{\tiny$\pm$.027} & .692{\tiny$\pm$.015} & .682{\tiny$\pm$.054} & .649{\tiny$\pm$.030} & .653{\tiny$\pm$.020} & .599{\tiny$\pm$.012} & .664 \\
\midrule
\multicolumn{9}{l}{\textit{Deep KT (pyKT)}} \\
\bottomrule
\end{tabular}
\end{table*}

\end{document}